\documentclass[11pt]{article}

\usepackage[preprint]{acl}
\usepackage{latexsym}
\usepackage{fontspec}
\newif\ifarxiv
\arxivtrue
\ifarxiv
\usepackage{xeCJK}
\newCJKfontfamily\japanesefont{ipam.ttf} 

\newfontfamily\arabicfont[Script=Arabic]{Amiri-Regular.ttf}
\newfontfamily\hindifont[Script=Devanagari]{NotoSerifDevanagari-Regular.ttf}
\newfontfamily\tamilfont[Script=Tamil]{NotoSerifTamil-Regular.ttf}
\newfontfamily\telugufont[Script=Telugu]{NotoSerifTelugu-Regular.ttf}
\else

\usepackage{xeCJK}
\newCJKfontfamily\japanesefont{Noto Serif CJK JP}

\newfontfamily\hindifont[Script=Devanagari]{Noto Serif Devanagari}
\newfontfamily\tamilfont[Script=Tamil]{Noto Serif Tamil}
\newfontfamily\telugufont[Script=Telugu]{Noto Serif Telugu}
\newfontfamily\arabicfont[Script=Arabic]{Amiri}

\fi

\usepackage{xurl}
\usepackage{booktabs}       
\usepackage{makecell}
\usepackage{amsfonts}       
\usepackage{nicefrac}       
\usepackage{microtype}      
\usepackage{inconsolata}  
\usepackage[inline]{enumitem}
\usepackage{graphicx}
\usepackage{subcaption}
\usepackage{wrapfig}
\usepackage{float}
\usepackage{amsmath}
\usepackage[most]{tcolorbox}
\usepackage{array}
\usepackage[section]{placeins}  
\usepackage{dblfloatfix}        
\usepackage{colortbl}
\usepackage{xspace}

\DeclareMathOperator*{\argmax}{arg\,max}
\newcommand{\benchmark}{\textsc{XLGoBench}\xspace}

\newif\ifnames
\namestrue

\title{\benchmark: Detecting cross-lingual skill gaps with algorithmic tasks}
\ifnames
\author{Purvam Jain\textsuperscript{1} \quad
  Preethi Jyothi\textsuperscript{2,1} \quad
  Vihari Piratla\textsuperscript{1}\thanks{\ \ Corresponding author: \texttt{vpiratla@google.com}} \quad
  Suvrat Raju\textsuperscript{3,1} \\
  \textsuperscript{1}Google DeepMind \\
  \textsuperscript{2}Indian Institute of Technology Bombay \\
  \textsuperscript{3}International Centre for Theoretical Sciences, Tata Institute of Fundamental Research
}
\else
\author{Anonymous ACL Submission}
\fi

\newtcolorbox{promptbox}[1][]{
  colback=gray!5,
  colframe=gray!75!black,
  fonttitle=\bfseries,
  title=#1,
  arc=0mm,
  breakable,
  enhanced,
  attach title to upper,
  after title={:\medskip\\},
  boxrule=0.5pt,
}

\newif\ifshowcomments
\showcommentstrue 
\ifshowcomments
\newcommand{\vp}[1]{\textcolor{red}{[Comment by Vihari]: #1}}
\newcommand{\pj}[1]{\textcolor{blue}{[Comment by Preethi]: #1}}
\newcommand{\pu}[1]{\textcolor{green}{[Comment by Purvam]: #1}}
\newcommand{\sr}[1]{\textcolor{magenta}{[Comment by Suvrat]: #1}}
\else
\newcommand{\vp}[1]{\textcolor{red}{}}
\newcommand{\pj}[1]{\textcolor{blue}{}}
\newcommand{\pu}[1]{\textcolor{green}{}}
\newcommand{\sr}[1]{\textcolor{magenta}{}}
\fi

\usepackage{polyglossia}
\usepackage{bidi}
\usepackage{multirow}
\begin{document}

\maketitle

\begin{abstract}
We introduce a set of synthetic algorithmic tasks to detect cross-lingual gaps in the abilities of large language models. Our benchmark is commensurate across languages, since it requires models to perform the same underlying task in different languages; scalable, since each task can be generated at varying levels of complexity allowing it to be adapted to models with different capabilities; quantifiable, since every task admits an objective notion of correctness; and transparent, since tasks are generated from simple templates that can be readily audited for translation errors. Because our benchmark focuses on algorithmic tasks, differential performance is a sufficient---but not necessary---indicator of cross-lingual gaps. Nevertheless, we show through extensive experiments that our benchmark exposes persistent cross-lingual gaps in multiple state-of-the-art models.
\end{abstract}

\section{Introduction}
\begin{figure*}[htbp]
    \centering
    \begin{subfigure}[b]{0.49\textwidth}
        \includegraphics[width=\textwidth]{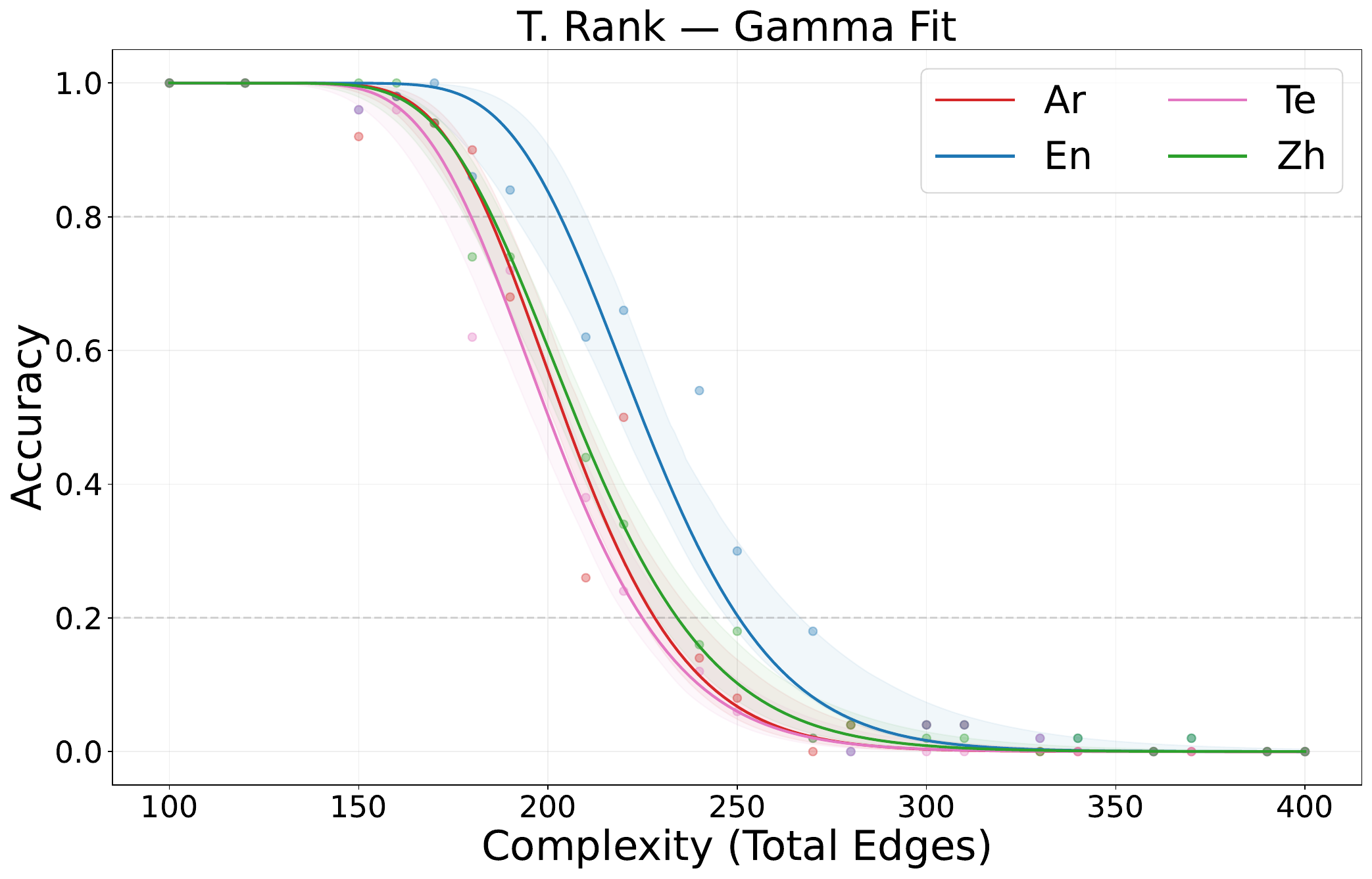}
    \end{subfigure}
    \hfill
    \begin{subfigure}[b]{0.49\textwidth}
        \includegraphics[width=\textwidth]{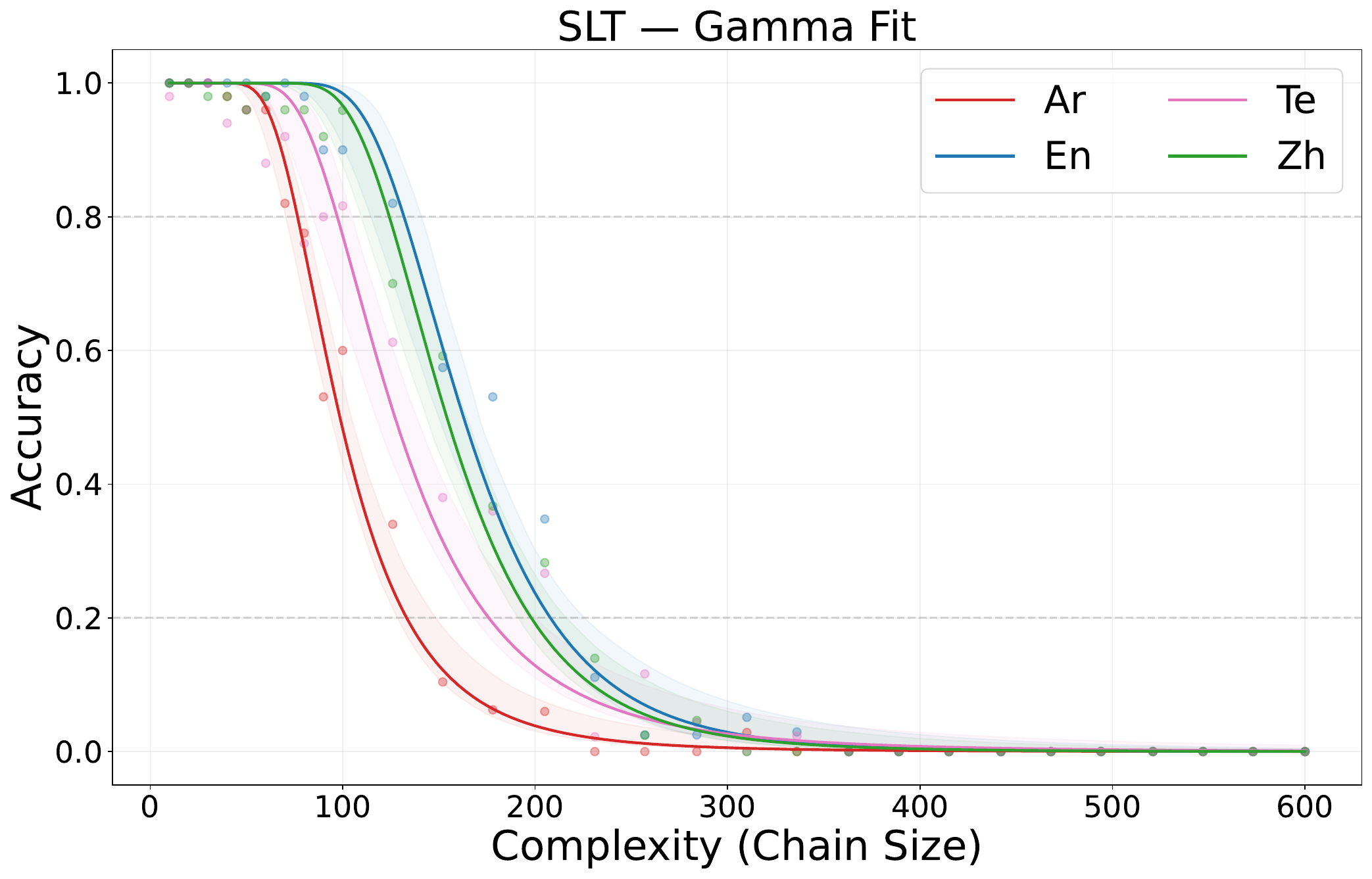}
    \end{subfigure}
    \caption{We show a graph of accuracy vs complexity for selected languages for (a) Gemini-3-Flash for the Tournament ranking task on the left and (b)  Gemini-3.1-Flash-Lite for the Sequential Linear Transforms task on the right.  Significant cross-lingual gaps are observed in the intermediate complexity region, as performance in English declines more slowly with complexity compared to other languages. The dots denote observed accuracy and the solid lines and uncertainty bands are obtained by fitting a mathematical model described in Section \ref{sec:expts} to the data.}
    \label{fig:poster}
\end{figure*}

Large language models (LLMs) have had a significant and increasing impact on society, including in fields such as education \citep{hausman2025generative}, professional writing \citep{noy2023experimental}, customer service \citep{brynjolfsson2025generative}, programming \citep{handa2025economic},  and others \citep{hausman2025generative}. Nevertheless, the widespread adoption of these models has exacerbated concerns about global inequality \citep{skare2024artificial,UNDPAIDIV,misra2025measuring}.

Cross-lingual gaps in the abilities of LLMs represent one facet of broader disparities faced by users. These gaps persist because capability evaluation benchmarks are largely in English \citep{ahuja-etal-2024-megaverse}, leaving model abilities across many non-English languages largely unmeasured \citep{wu2025bitter}. On the other hand, it is encouraging that LLMs show a measure of cross-lingual transfer, which allows them to perform well even in relatively low-resource languages \citep{nguyen2024democratizing}. Moreover, improvements in the abilities of models can make LLMs more accessible to users  \citep{microsoft2026globalai} and potentially even empower linguistically marginalized communities \citep{ding2024foundation}.

Therefore, removing cross-lingual gaps is both important and potentially achievable. The first step towards this objective is to precisely measure cross-lingual gaps. Towards this goal, we introduce a new synthetic benchmark. 

Our benchmark \benchmark comprises a set of algorithmic puzzles. This offers some advantages over alternate methods of measuring cross-lingual gaps. 
\begin{enumerate}
\item {\bf Transparency in translation:} A recurrent challenge in multilingual evaluations is to ensure translation accuracy. For example, popular multilingual benchmarks such as MMLU \citep{hendrycks2020measuring} and MGSM \citep{shi2022language} have been shown to contain translation, annotation or evaluation errors \citep{gema2025we,singh2025global,peter2025mind}. This makes it unclear whether, for example, the remaining 1.2\% gaps between English and Telugu on MGSM \citep{peter2025mind}  with Gemini-2.5-Flash arise from genuine capability differences or experimental errors. In contrast, our puzzles are formed by combining a small set of templates. The translation accuracy of these templates can be checked by hand, as we did for our empirical tests, using expert human raters. Therefore, we do not expect translation errors to affect our evaluations.
\item {\bf Scalability:} Traditional multilingual evaluations involve a multi-step process and manual collation. This has led to benchmarks that include long-context reasoning \citep{kim2025one}, expert knowledge \citep{xuan2025mmlu} or hard coding tasks \citep{huang2025benchmax}. However, as models continue to improve, such benchmarks are expected to saturate, requiring the creation of new datasets. 

In contrast, our puzzles  can be scaled to arbitrarily high complexity.  This reduces the accuracy of all models, including those that we might train in the foreseeable future, and exposes cross-lingual gaps where they exist. Similarly, the complexity can also be scaled down to test smaller models. 
\item
{\bf Quantifiability:} Since the answers to our puzzles are easily verifiable, it is possible to perform a large number of experiments. This reduces potential statistical fluctuations in our measurements and enables the precise quantification of cross-lingual variation in model performance. Moreover, we do not need to use an LLM as a judge, removing another potential source of ambiguities.
\item
{\bf Commensurate Tasks:} In a multilingual test, it is subtle to ensure the task performed in one language is commensurate with a similar task in another language. For instance, differences in nuance, sentence structure and content might make it harder for models to retrieve information from a corpus in one language compared to another. Our tasks sidestep this issue since our puzzles are simple to express and, at an abstract level, encode a language-independent task that is the same regardless of the prompt language.%
\footnote{Here, commensurate refers to the underlying abstract puzzle rather than its surface realization in tokens. The potential implications of inefficient tokenization on cross-lingual gaps are discussed in Section~\ref{sec:discussion}.} 
\end{enumerate}

The use of synthetic algorithm-based tasks is well-recognized for training and evaluating reasoning abilities of LLMs~\citep{liu2025saturn,zhang2025trainingevaluatinglanguagemodels,tang2025grapharenaevaluatingexploringlarge,wei2025satbenchbenchmarkingllmslogical,openai_2026,he2026resynautonomouslyscalingsynthetic}.
Nevertheless, it was not a priori guaranteed that cross-lingual gaps in the abilities of models on real-world tasks would be reflected in the simple synthetic tasks that we evaluate. A surprising empirical result of our study is that frontier models display differences in performance even on such synthetic puzzles. 

We emphasize that it is possible that a model that performs equally in two different languages on our simple algorithmic tasks, might show uneven performance when tested on tasks like creative writing, or awareness of nuance.  So, our perspective is that gaps in performance in algorithmic tasks are {\em sufficient} but not necessary to demonstrate the existence of cross-lingual gaps. As a corollary, it is important to supplement the benchmark presented in this paper with measures of more subtle linguistic capabilities, and we leave this to future work. 

\begin{figure*}[htbp]
    \centering
        \includegraphics[width=0.95\textwidth]{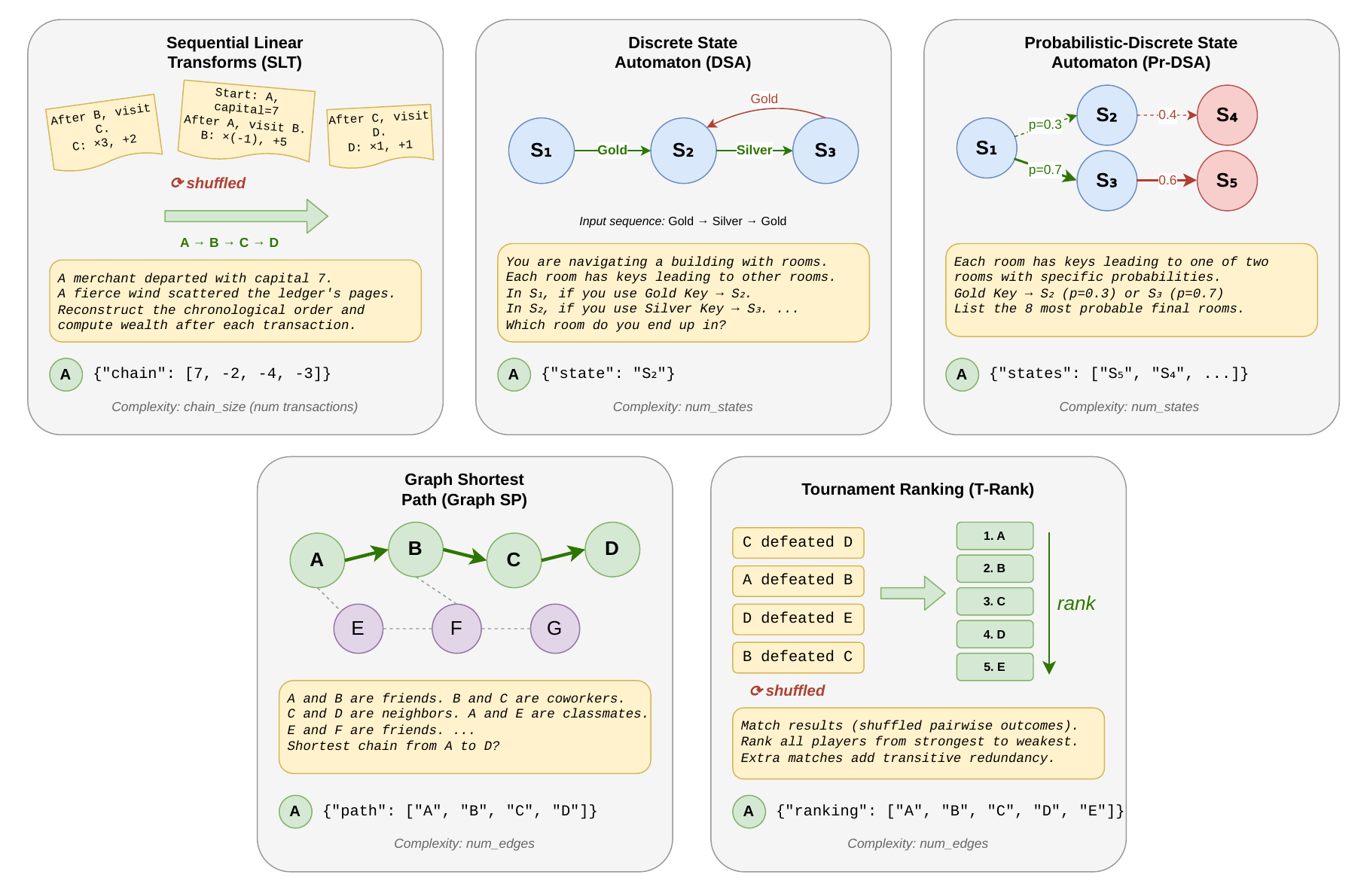}
    \caption{Five puzzle types in \benchmark. We visually depict the basic schema of all puzzle types in our benchmark, along with their respective complexity measures. More detailed puzzle descriptions are in Section~\ref{sec:method}.}
    \label{fig:task_desc}
\end{figure*}

\paragraph{\bf Summary of results.}
We test four frontier models on a suite of five puzzles (described in more detail in Section \ref{sec:method}) across  seven different languages. The puzzles are based on templates, that are vetted by expert human raters as described in Section \ref{secdatavalidation}, and are scalable in complexity. We track the performance of the models from low complexity --- where all models achieve perfect accuracy in all languages --- to high complexity, where the accuracy collapses as described in Section \ref{sec:expts}. A statistically significant variation in the performance of models at a given complexity reflects a cross-lingual gap. We present a summary metric that measures the maximum possible cross-lingual gap utilizing a model of the variation of the accuracy with complexity.  We show in Section \ref{sec:main_results} that this suite of puzzles, taken collectively,  is sufficient  to expose unambiguous cross-lingual gaps in all models.  In Section \ref{sec:mit_expts}, we show that these gaps cannot be attributed to translation errors. We also note that these gaps arise at intermediate complexity levels. We comment on possible underlying causes and future work in Section \ref{sec:discussion}. The Appendices provide extended empirical results, including plots of accuracy vs complexity, and details on prompts and the construction of puzzles.  The full set of results, code to reproduce our plots, and templates for the puzzles are included in the supplementary material for this paper.\ifnames This supplementary material can be downloaded from \href{https://zenodo.org/records/20423110}{zenodo.org/records/20423110}
\fi

\subsection{Related Work}
Benchmarks dedicated to measuring cross-lingual gaps have been of great interest since the era of encoder-only language models~\citep{hu2020xtreme}, and their scope has since expanded from classification tasks to more complex reasoning tasks for evaluating modern LLMs~\citep{ghosh-etal-2025-survey}. Most closely related to our work, \citet{dobler2026multilingual} previously introduced cross-lingual evaluations using synthetic tasks by translating templates from an English reasoning gym \citep{stojanovski2025reasoning}. \citet{bruun2025multizebralogic} previously studied the variation in the accuracy of models on configurations of the Zebra puzzle across several European languages.  \citet{kim2025one} also described synthetic tasks to test the ability of models to retrieve and aggregate data in different languages.

In contrast to prior work, our puzzles were selected explicitly for cross-lingual evaluation and are not translations of any existing reasoning benchmark. They also involve considerably more reasoning than the typical tasks presented in \citet{dobler2026multilingual} and \citet{kim2025one}, which allows our benchmark to expose cross-lingual gaps in some of the largest frontier LLMs. 

The structure of our puzzles, which can be solved by repetitively applying a simple algorithm,  allows us to make contact with parallel quantitative studies of errors in LLMs \citep{raju2026model} leading to interesting novel observations about the variation of the cross-lingual gap with complexity. 

Additional benchmarks are surveyed in \citet{uppadhyay2026the}. Our benchmark is immune to concerns about data leakage described there.   Each puzzle instance is randomly generated from a template, allowing for a very large number of unique realizations, making memorization unlikely.

\section{The \benchmark Benchmark}
\label{sec:method}
Our benchmark comprises a suite of algorithmic tasks. Each of these tasks can be solved by a repeated application of a small number of atomic operations. 
We propose to compare model correctness across the same problem translated in various languages. We leave the language of the thinking and response space to model default. 
In this paper, we consider six languages apart from English:  Arabic (AR), Hindi (HI), Japanese (JA), Chinese (ZH), Tamil (TA), Telugu (TE). These languages represent large speaker populations and span diverse language families and scripts. The benchmark can be easily extended to other languages.

Now that several flagship language models offer context windows of one million or longer, reasoning over information presented in context is an important skill. This is a potential source of cross-lingual gaps, owing to the volume of non-English tokens. We picked three graph walk and two graph search tasks to align with the practical challenge of navigating information graphs (in context). We illustrate the structure of the five puzzles in our evaluation suite in Figure~\ref{fig:task_desc}.
Unique instances of the puzzle are generated using a base template that can be vetted for translation errors, and then scaled to any desired complexity. 

While the actual content of the puzzles is always in the respective script, the identifiers used for people's names or states are always in alphanumeric form, such as A1234. 
This avoids potential ambiguities --- for instance, in some cultures, ``Banerjee" and ``Bandopadhyaya" are understood to be equivalent whereas in others ``Robert" and ``Bob" are used interchangeably. The use of alphanumeric identifiers also allows us to scale the number of unique entities to any desired value. We provide example prompts for each task in Appendix~\ref{appendix:dataset_examples}, both in English and other languages. 

While the puzzle content is in the respective scripts, all formatting details, algorithmic constraints, and output constraints are provided in English. This is a design choice to ensure that the gaps we observe do not arise due to known deficiencies in multilingual instruction-following \citep{li2026xifbench, li2024cif}. Moreover, while code-switching in the input could be a cause for concern as a potential trigger of errors, frontier LLMs are generally robust to such inputs. In fact, in-context code-switching has been shown to improve performance across multiple LLMs and tasks~\citep{yoo2025code}.

\subsection{Graph walk problems}
\paragraph{Sequential Linear Transforms (SLT).} A traveling merchant makes a sequence of transactions with $T$ people starting with an initial wealth of $w_0$. Each $t^{\text{th}}$ transaction is uniquely identified by the customer name, along with multiplicative ($a_t$) and additive ($b_t$) adjustment factors to the merchant's wealth, and mentions the $(t+1)^{\text{th}}$ customer. The transactions are chronologically jumbled to make the task harder. The problem requires computing the final wealth $w_T$ through a sequence of transformations $w_{t+1}=a_tw_t+b_t$. The numbers are chosen to satisfy  $a_t, b_t, w_t \in [-9, 9]$.

The total number of transactions $T$ determines the complexity of the problem. Please see Appendix~\ref{sec:slt_example} for a full prompt. 

\paragraph{Discrete State Automaton (DSA).} 
Each problem specifies a discrete state automaton, i.e., states transition depending on the action. We verbalize the problem by defining states as different rooms and actions as keys to various other connecting rooms. We then prompt for the final state given the starting room (initial state) and a sequence of keys (actions). The state transitions are all deterministic. We verify the answer through exact match on the correct final state.

We use an equal number of states and actions, which defines the complexity. Please see an example prompt in Appendix~\ref{sec:det_fsm_example}.

\paragraph{Probabilistic-Discrete State Automata (Pr-DSA).}
The setup is similar to DSA except that the state transitions are probabilistic. For a given (state, action) pair, we specify the probabilities of two possible next states for a given action. Given the initial state and a sequence of actions, we prompt the model for the eight most likely final states, and validate the response based on whether the top-8 set (order-invariant) is correctly identified. We fixed the number of actions to a constant value of 15 and varied the number of states.
Puzzle complexity, thereby, is defined by the number of states. Please refer to Appendix~\ref{sec:prob_fsm_example} for an example prompt.

\subsection{Graph search problems}
\paragraph{Graph shortest path (Graph SP).}
Given a social acquaintance graph with connections described as {\it \{Person A\} and \{Person B\} are friends}, the problem requires finding the shortest path between any two people, which is ensured to be unique by construction. Please find graph construction details in Appendix~\ref{sec:graph_sp_generation}. The number of edges in the graph defines complexity. Please see Appendix~\ref{sec:graph_sp_example} for an example prompt.

\paragraph{Tournament Ranking (T-Rank).}
Given $n$ players with unique strengths, the problem requires ordering players according to their strength given a directed graph of game outcomes. During construction, we ensure the problem is well-specified and the solution is unique through acyclic and connected graphs, as we first construct the backbone  of a Hamiltonian path and then add extra valid edges. 
We maintain a ratio of $1:3$ between number of nodes (players) and number of edges (matches) in the graph such that the complexity is defined by a single variable: the number of edges. Please see Appendix~\ref{sec:tsort_example} for an example prompt.

\subsection{Validation  of questions\label{secdatavalidation}}
Traditional approaches demand considerable annotation effort to ensure questions remain valid in all languages.
Our setup requires considerably low annotation effort by construction. We only need to ensure that the few templates that constitute each problem are translated correctly to all languages. The average number of templates that make up any prompt across tasks is only $5$. 

We confirmed the validity of translations in English and six non-English languages with native speakers who performed the following checks. (a) Template translations in various languages are semantically equivalent to that in English, (b) spot-checking with five full prompts if they make a meaningful problem of intended meaning in each language. For English, we only need step (b) since there are no translations in English. We provide further details in Appendix~\ref{sec:annot_pipe}.

\subsection{Validation of responses}
We instructed models to produce JSON-formatted output that was parsed using an automated script. For Pr-DSA, and the two graph search problems where the solution is a list of values, the answer was marked as correct only when the entire list matched the ground-truth solution. For Pr-DSA, the order of the states was ignored for response matching. 

This procedure sidesteps errors that could have been introduced by using an LLM judge. 

\subsection{Cross-lingual gaps}
The templates described above can be used to generate unique puzzles with varying complexity. The accuracy of models is expected to drop from nearly 100\% at very low complexity to nearly 0\% at very high complexity ~\citep{shojaee2026the,raju2026model}. However, the precise rate of decline of the accuracy with complexity might vary across languages. Our benchmark detects cross-lingual gaps by identifying differences in the accuracy vs complexity curve across languages.

\section{Experiments}
\label{sec:expts}

\paragraph{Models.} Our empirical validation spans the following models: Gemini-3-Flash~\citep{gemini-3}, Gemini 3.1 Flash-Lite~\citep{gemini-3.1}, Gemma-4-31B~\citep{gemma-4}, GLM-5.1~\citep{glm-5-1}. This selection was made to ensure that (a) we cover both open and proprietary models and (b) that we cover models developed in different countries, since this might influence the linguistic distribution of the training data and evaluation process. 

\paragraph{Complexity.}
Our setup naturally enables studying cross-lingual gaps at various complexity levels. 
For each problem type, we generate problems of varying complexities such that we see perfect accuracy at the lowest complexity level and zero accuracy at the highest complexity level for English language (as seen in Figure~\ref{fig:poster}). Table~\ref{tab:complexity_param} shows the default complexity ranges we used for different tasks and also describes the parameter used to measure the complexity. The complexity range for Gemini 3.1 Flash-Lite was chosen separately as described in Appendix \ref{app:litecomplexityselection}.\footnote{Complexity ranges for each puzzle type were identified using one of our models (Gemini-3-Flash). $c_{\min}$ and $c_{\max}$ denote configurations with consistent near-$100\%$ and near-$0\%$ accuracy across multiple runs, enabling a full gamma fit.}

\paragraph{Sample collection.}
We uniformly sample 20 complexity levels from the ranges specified in Table~\ref{tab:complexity_param}. For each complexity level, we randomly generate 50 questions per language (seeds are reused across languages). This protocol is followed across all puzzle types defined in Section~\ref{sec:method}. Therefore, in total, our experiments comprise about $4\times 20\times50\times7\times5 = 140,000$ prompts. (Total prompts~=~models $\times$ configurations $\times$ questions per configuration $\times$ languages $\times$ tasks.) 

\begin{table}[htbp]
\centering
\footnotesize
\begin{tabular}{l@{\hspace{10pt}}ccccc}
\toprule
\makecell[l]{\textbf{Puzzle}} & \textbf{Complexity Measure} & \textbf{$c_{\min}$} & \textbf{$c_{\max}$} \\
\midrule
\makecell[l]{SLT} & Chain length & $100$ & $600$ \\
\midrule
\makecell[l]{DSA} & Number of states & $32$ & $2090$  \\ 
\midrule
\makecell[l]{Pr\ DSA} & Number of states & $9$ & $61$  \\
\midrule
\makecell[l]{Graph SP} &  Number of edges & $70$ & $720$ \\
\midrule
\makecell[l]{T.\ Rank} & Number of edges & $100$ & $400$  \\
\bottomrule
\end{tabular}
\caption{Complexity measure and default range of complexity values for each puzzle.}
\label{tab:complexity_param}
\end{table}

\paragraph{Analyzing cross-lingual gaps.}
The procedure above yields a picture of the decline of accuracy with complexity in each language. At a fine-grained level, any variation in the accuracy of the model at a given complexity indicates a cross-lingual gap.  However, although the cross-lingual gap is, strictly speaking, measured by a function worth of information, we find it convenient to compress this into a single numerical metric using the procedure below.

We utilize the quantitative model of the variation of the accuracy of LLMs with the complexity of simple algorithmic tasks described in \citet{raju2026model}. In this model, the accuracy $f_{\ell}$  in language $\ell$ is related to the complexity via
\begin{equation}
\label{gammafit}
    f_\ell(c) = \frac{1}{\Gamma(q_\ell/2)}\gamma\!\left(\frac{q_\ell}{2},\; \frac{q_\ell}{2\, r_\ell\, c^2}\right)
\end{equation}
where $\gamma(x, y)$ denotes the lower incomplete gamma function,
 $c$ is the task complexity and $r_\ell, q_\ell > 0$ are language-specific parameters that can be interpreted in terms of an elementary error rate, and the number of plausible erroneous tokens at each step. 
This functional form naturally captures the sigmoidal decay from near-perfect accuracy at low complexity to near-zero accuracy at high complexity,  and it provides a good fit to our empirical data. We computed $R^2$ values for the fit and found $R^2 > 0.97$ across all models and tasks.

We fit the mean accuracy as measured above, with the complexity parameter as defined in Table~\ref{tab:complexity_param}, to define the {\em signed max divergence (SMD)} for a language $\ell$.
\begin{equation}
 \operatorname{SMD}(\ell) = f_{\text{En}}(c^*) - f_\ell(c^*),
\end{equation}
where
\begin{equation}
\label{cstardef}
    c^* = \argmax\lvert f_{\text{En}}(c) - f_\ell(c) \rvert.
\end{equation}
The SMD captures the worst possible cross-lingual gap for the respective model, task and language combination when compared to English. 

The SMD can potentially be subtle to interpret in cases where a language performs better than English for some complexity range and worse in another complexity range. Therefore, in Appendix \ref{appfullresults}, we define a second measure called the reciprocal deviation (RD) that can be used to check for for such a feature in the data.  We do not find any statistically significant value for the RD in cases where we report a significant SMD.

\paragraph{Confidence intervals.} 
We generate 300 samples of the accuracy at each value of the complexity using the posterior beta distribution described in Appendix C of \cite{raju2026model} to get multiple estimates for $q_\ell, r_\ell$. The distribution of gamma fit parameters defines the standard deviation for the SMD, which is reported below.

\section{Results}
\label{sec:main_results}

Our main results are in Tables~\ref{tab:maxdiv_gemini_3_flash}, \ref{tab:maxdiv_gemma_4_31b_it}, \ref{tab:maxdiv_flash_lite}, \ref{tab:maxdiv_glm-5-1}.
We consider problems with insignificant cross-lingual gaps equally valuable and report observations on every task irrespective of their cross-lingual gaps. We report SMD values along with standard deviations per task for each model.

On average, Gemma-4-31B shows the smallest cross-lingual gaps, which is consistent with the stated multilingual emphasis of the model~\citep{gemma-4}. It is notable that GLM-5.1's performance on Chinese exceeds that of English on the Graph SP task and is on par with English on other tasks. This is consistent with the stated emphasis on training and evaluating the model on Chinese \citep{glm-5-1}.  On the whole, Tamil and Telugu saw the largest gaps which might arise due to their relatively low representation in the training data among the six non-English languages.

\begin{table*}[htbp]
\centering
\normalsize

\begin{subtable}{1\linewidth}
\centering
\begin{tabular}{l@{\hspace{10pt}}ccccccc}
\toprule
\textbf{Task} & \textbf{AR} & \textbf{HI} & \textbf{JA} & \textbf{TA} & \textbf{TE} & \textbf{ZH} \\
\midrule
SLT        &  \cellcolor{red!20}$9.49{\scriptstyle\pm4.12}$ & $3.51{\scriptstyle\pm4.69}$ & $8.56{\scriptstyle\pm4.92}$ & \cellcolor{red!20}$10.77{\scriptstyle\pm4.26}$ & \cellcolor{red!20}$18.56{\scriptstyle\pm4.36}$ & $-1.79{\scriptstyle\pm5.60}$   \\
\midrule
DSA         & \cellcolor{red!20}$29.72{\scriptstyle\pm5.75}$ & \cellcolor{red!20}$53.48{\scriptstyle\pm5.31}$ & \cellcolor{red!20}$47.73{\scriptstyle\pm5.12}$ & \cellcolor{red!20}$42.71{\scriptstyle\pm5.28}$ & \cellcolor{red!20}$45.11{\scriptstyle\pm5.11}$ & \cellcolor{red!20}$28.40{\scriptstyle\pm6.14}$   \\
\midrule
Pr\ DSA     & $6.05{\scriptstyle\pm4.34}$ & $5.39{\scriptstyle\pm5.23}$ & $-4.34{\scriptstyle\pm4.73}$ & $4.00{\scriptstyle\pm4.57}$ & $-1.50{\scriptstyle\pm5.33}$ & $-8.49{\scriptstyle\pm4.52}$   \\
\midrule
Graph SP    & $6.54{\scriptstyle\pm4.63}$ & \cellcolor{red!20}$16.83{\scriptstyle\pm3.97}$ & $3.88{\scriptstyle\pm4.43}$ & $8.29{\scriptstyle\pm5.10}$ & \cellcolor{red!20}$20.48{\scriptstyle\pm3.66}$ & \cellcolor{red!20}$12.47{\scriptstyle\pm4.93}$   \\
\midrule
T.\ Rank   & \cellcolor{red!20}$29.85{\scriptstyle\pm6.35}$ & \cellcolor{red!20}$27.49{\scriptstyle\pm6.39}$ & \cellcolor{red!20}$19.43{\scriptstyle\pm6.20}$ & \cellcolor{red!20}$24.84{\scriptstyle\pm6.30}$ & \cellcolor{red!20}$35.25{\scriptstyle\pm6.32}$ & \cellcolor{red!20}$25.01{\scriptstyle\pm6.30}$   \\
\midrule
\bottomrule
\end{tabular}
\caption{\textbf{Gemini 3 Flash}}
\label{tab:maxdiv_gemini_3_flash}
\end{subtable}

\vspace{0.5em}
\begin{subtable}{1\linewidth}
\centering
\begin{tabular}{l@{\hspace{10pt}}ccccccc}
\toprule
\textbf{Task} & \textbf{AR} & \textbf{HI} & \textbf{JA} & \textbf{TA} & \textbf{TE} & \textbf{ZH} \\
\midrule
SLT        & \cellcolor{red!20}$20.98{\scriptstyle\pm5.35}$ & $4.68{\scriptstyle\pm9.13}$ & $5.35{\scriptstyle\pm8.08}$ & \cellcolor{red!20}$14.21{\scriptstyle\pm6.72}$ & $4.81{\scriptstyle\pm7.06}$ & $-5.37{\scriptstyle\pm7.31}$   \\
\midrule
DSA        & \cellcolor{red!20}$13.48{\scriptstyle\pm4.99}$ & \cellcolor{red!20}$16.40{\scriptstyle\pm4.97}$ & \cellcolor{red!20}$12.94{\scriptstyle\pm5.41}$ & $4.31{\scriptstyle\pm6.65}$ & \cellcolor{red!20}$12.65{\scriptstyle\pm5.13}$ & $1.06{\scriptstyle\pm6.14}$   \\
\midrule
Pr\ DSA     & $-6.13{\scriptstyle\pm5.98}$ & $9.15{\scriptstyle\pm6.03}$ & $4.53{\scriptstyle\pm6.00}$ & \cellcolor{red!20}$12.50{\scriptstyle\pm5.80}$ & $-3.33{\scriptstyle\pm5.48}$ & $7.62{\scriptstyle\pm5.31}$   \\
\midrule
Graph SP    & $-1.88{\scriptstyle\pm5.06}$ & $-3.60{\scriptstyle\pm5.58}$ & $-4.47{\scriptstyle\pm5.16}$ & $6.97{\scriptstyle\pm4.30}$ & $7.99{\scriptstyle\pm4.00}$ & $5.33{\scriptstyle\pm4.64}$   \\
\midrule
T.\ Rank    & \cellcolor{red!20}$14.06{\scriptstyle\pm5.23}$ & $9.83{\scriptstyle\pm5.27}$ & $-4.58{\scriptstyle\pm7.84}$ & \cellcolor{red!20}$13.90{\scriptstyle\pm6.58}$ & \cellcolor{red!20}$13.36{\scriptstyle\pm5.76}$ & $6.33{\scriptstyle\pm5.32}$   \\
\midrule
\bottomrule
\end{tabular}
\caption{\textbf{Gemma 4 31B IT}}
\label{tab:maxdiv_gemma_4_31b_it}
\end{subtable}

\vspace{0.5em}

\begin{subtable}{1\linewidth}
\centering
\begin{tabular}{l@{\hspace{10pt}}ccccccc}
\toprule
\textbf{Task} & \textbf{AR} & \textbf{HI} & \textbf{JA} & \textbf{TA} & \textbf{TE} & \textbf{ZH} \\
\midrule
SLT       & \cellcolor{red!20}$61.15{\scriptstyle\pm6.53}$ & \cellcolor{red!20}$43.60{\scriptstyle\pm6.65}$ & \cellcolor{red!20}$27.72{\scriptstyle\pm7.68}$ & \cellcolor{red!20}$30.60{\scriptstyle\pm7.60}$ & \cellcolor{red!20}$34.21{\scriptstyle\pm8.16}$ & $8.37{\scriptstyle\pm8.92}$   \\
\midrule
DSA         & \cellcolor{red!20}$14.33{\scriptstyle\pm5.13}$ & $2.48{\scriptstyle\pm6.74}$ & $4.77{\scriptstyle\pm6.18}$ & \cellcolor{red!20}$21.95{\scriptstyle\pm5.72}$ & \cellcolor{red!20}$31.46{\scriptstyle\pm5.09}$ & \cellcolor{red!20}$19.99{\scriptstyle\pm5.40}$   \\
\midrule
Pr\ DSA     & $4.46{\scriptstyle\pm5.53}$ & \cellcolor{red!20}$14.10{\scriptstyle\pm4.16}$ & \cellcolor{red!20}$12.20{\scriptstyle\pm4.56}$ & $-3.20{\scriptstyle\pm7.59}$ & $7.37{\scriptstyle\pm4.62}$ & \cellcolor{red!20}$10.95{\scriptstyle\pm4.70}$   \\
\midrule
Graph SP   & $-3.68{\scriptstyle\pm8.58}$ & $-6.12{\scriptstyle\pm7.37}$ & $-4.46{\scriptstyle\pm7.56}$ & $-6.17{\scriptstyle\pm7.28}$ & $-10.50{\scriptstyle\pm7.10}$ & $-5.00{\scriptstyle\pm8.06}$   \\
\midrule
T.\ Rank   & \cellcolor{red!20}$27.45{\scriptstyle\pm3.53}$ & \cellcolor{red!20}$24.44{\scriptstyle\pm3.80}$ & \cellcolor{red!20}$25.13{\scriptstyle\pm3.76}$ & \cellcolor{red!20}$16.50{\scriptstyle\pm3.80}$ & \cellcolor{red!20}$20.70{\scriptstyle\pm4.34}$ & \cellcolor{red!20}$16.53{\scriptstyle\pm4.71}$   \\
\midrule
\bottomrule
\end{tabular}
\caption{\textbf{Gemini 3.1 Flash Lite}}
\label{tab:maxdiv_flash_lite}
\end{subtable}

\vspace{0.5em}

\begin{subtable}{1\linewidth}
\centering
\begin{tabular}{l@{\hspace{10pt}}ccccccc}
\toprule
\textbf{Task} & \textbf{AR} & \textbf{HI} & \textbf{JA} & \textbf{TA} & \textbf{TE} & \textbf{ZH} \\
\midrule
SLT         & $9.10{\scriptstyle\pm5.36}$ & $8.86{\scriptstyle\pm6.21}$ & $6.99{\scriptstyle\pm7.33}$ & \cellcolor{red!20}$18.05{\scriptstyle\pm5.37}$ & \cellcolor{red!20}$12.56{\scriptstyle\pm5.66}$ & $9.22{\scriptstyle\pm5.83}$   \\
\midrule
DSA         & \cellcolor{red!20}$12.28{\scriptstyle\pm5.19}$ & \cellcolor{red!20}$22.64{\scriptstyle\pm5.83}$ & $-6.79{\scriptstyle\pm6.30}$ & \cellcolor{red!20}$43.89{\scriptstyle\pm6.09}$ & \cellcolor{red!20}$42.59{\scriptstyle\pm5.45}$ & $6.15{\scriptstyle\pm6.46}$   \\
\midrule
Pr\ DSA     & $6.71{\scriptstyle\pm3.50}$ & \cellcolor{red!20}$9.80{\scriptstyle\pm3.90}$ & $6.59{\scriptstyle\pm3.96}$ & \cellcolor{red!20}$15.87{\scriptstyle\pm3.10}$ & \cellcolor{red!20}$11.48{\scriptstyle\pm3.00}$ & $3.44{\scriptstyle\pm5.50}$   \\
\midrule
Graph SP    & $-6.83{\scriptstyle\pm6.29}$ & $-8.02{\scriptstyle\pm7.58}$ & $-5.95{\scriptstyle\pm6.97}$ & $-6.82{\scriptstyle\pm5.52}$ & $-9.53{\scriptstyle\pm6.20}$ & \cellcolor{red!20}$-25.48{\scriptstyle\pm5.28}$   \\
\midrule
T.\ Rank    & $2.70{\scriptstyle\pm4.15}$ & $4.09{\scriptstyle\pm5.57}$ & \cellcolor{red!20}$14.66{\scriptstyle\pm3.75}$ & \cellcolor{red!20}$8.83{\scriptstyle\pm3.28}$ & $4.33{\scriptstyle\pm4.33}$ & $1.91{\scriptstyle\pm4.47}$   \\
\midrule
\bottomrule
\end{tabular}
\caption{\textbf{GLM-5-1}}
\label{tab:maxdiv_glm-5-1}
\end{subtable}

\vspace{0.5em}

\caption{Signed Max Divergence (w. En in \%) $\pm\sigma$ per task and language across all evaluated models, shown along with its standard deviation. Positive values indicate superior performance by EN. We highlight cells with a significant gap (outside the reported $1.96*\sigma$) in red.  Please see Section~\ref{sec:main_results} for observations and Appendix ~\ref{appfullresults} for extended empirical results including additional measures of accuracy and cross-lingual gaps. 
}
\label{tab:merged_maxdiv}
\end{table*}

\subsection{Observed cross-lingual gaps are not due to translation or other extraneous errors}
\label{sec:mit_expts}
As already mentioned, our puzzles are generated through templates that are human validated. However, beyond this, the following observations reinforce our confidence in the cause of gaps. 

\paragraph{No gaps at low complexity.} For lower complexity values, we observe that all languages achieve near 100\% performance and gaps arise only for intermediate levels of complexity. This implies the correctness of the puzzle template and  suggests that the reasoning gaps cannot be attributed to prompt ambiguities. This can  also be seen in Figure~\ref{fig:poster} and in the figures of Appendix~\ref{sec:gamma_profiles}.

\paragraph{Back-translation tests.} We performed additional checks to rule out the possibility that gaps arise due to mild phrasing or syntactic errors between languages. For cases where a significant SMD is observed, we back-translated templates from the worst-performing non-English languages to English for Gemini 3 Flash and Gemini 3.1 Flash Lite, and compared their performance with the original English prompt. Details of these experiments are provided in Appendix \ref{sec:bt_expts}. We did not find statistically significant deviations, which reinforces our confidence in the correctness of the templates. 

\subsection{Gaps arise at intermediate complexity} \label{subsec:complexity_gap}
As mentioned, all models achieve near perfect accuracy at low complexity. Similarly, the accuracy of all models collapses for sufficiently high complexity. Cross-lingual reasoning gaps are apparent only in the range of intermediate complexity. This highlights a critical drawback with static benchmarks for fast evolving models.  In our benchmark, we can easily vary the complexity range for each puzzle type and identify the ranges where cross-lingual gaps, if any, emerge.

\subsection{Not all problems show gaps} 
Even though all our puzzles require extensive reasoning, we do not observe uniform cross-lingual gaps across tasks. Moreover, the tasks on which we observe cross-lingual gaps also vary across models. One notable exception is the T-Rank task, that exposed cross-lingual gaps for almost all models and languages.

\section{Discussion}
\label{sec:discussion}
In this paper, we have introduced a method of detecting cross-lingual reasoning gaps. Unlike traditional benchmarks, generating problems with language templates enables us to scale problem complexity in any language while still preserving the accuracy of translation. Our extensive empirical tests demonstrate the utility of our benchmark, which serves to detect cross-lingual gaps in several state-of-the-art models.

One might naively have suspected that gaps can be ameliorated through longer thinking or traced back to the model expending less thinking effort in non-English languages.

However, our preliminary investigations do not indicate a clear correlation between thinking length and cross-lingual gaps in the results. This is discussed quantitatively in Appendix~\ref{sec:thought_corr}. 
This suggests that cross-lingual gaps might  reflect a fundamental issue that cannot be resolved merely by tweaking thinking in non-English languages. It might also be the case that the thinking process for multilingual prompts is sub-optimal.

A second observation is that most models (including GLM-5.1) tend to think in English even when the problem is given in another language. In our examination of thinking traces, we also observed the ``quote and think'' phenomenon described in \citet{yong2025crosslingual}, where the model quotes from the original prompt but then proceeds to think in English.  So, it is possible that performance in non-English languages degrades faster than English due to cross-attention between thinking tokens and the prompt tokens.

This suggests that the models could improve their accuracy by translating the question entirely into English and then ignoring the original prompt. However, we did not observe any model exploiting this procedure consistently.

Another possibility is that cross-lingual gaps might arise from inefficient tokenization in non-English languages. We have focused on the decline of accuracy with complexity, which is what matters from a user's point of view since tokenization is transparent to the user. But prompts at the same level of complexity might consume significantly different token budgets in different languages, depending on the tokenizer used by the model. Our preliminary observations do not suggest that models with a wide variation in token budget across languages, such as GLM-5.1, necessarily show larger cross-lingual gaps. However, this remains an important issue for further investigation.

This paper focuses on the detection of cross-lingual gaps. It is important to follow this up by developing methods to ameliorate these gaps. We leave this to future work.

\ifnames
\paragraph{\bf Acknowledgments.} We are grateful to Kinshuk Vasisht for collaboration on related work, which was a precursor to this study. We would like to thank Jason Hickey and Katja Filippova for comments on a draft of this manuscript. We would like to thank Partha Talukdar, Markus Freitag for engaging in discussions regarding the benchmark.
\fi

\section{Limitations.}
Our benchmark has two important limitations.
\paragraph{Real-world tasks.} We expect that a model that displays differential performance on the synthetic tasks here will also show differential performance in real world tasks. However, the converse is not true. Our puzzles do not directly correspond to realistic applications and, for simplicity, we even removed names from the prompts.

Eventually, we would like models to be sensitive to subtleties and nuance in various languages, and also generate text as proficiently as they do in English (e.g., as evaluated in \citet{vayani2025all}). The removal of obvious cross-lingual gaps that can be detected by our benchmark is an important step in this process. Our empirical results suggest that state-of-the-art models have not yet cleanly crossed this threshold.

\paragraph{Expensive evaluation.}
Since gaps emerge only on very long reasoning problems, the token count for our puzzles --- both in terms of input and output --- is high. For instance, we estimate incurring costs of approximately USD 2100 for Gemini 3 Flash and USD 6900 for GLM-5.1. These costs are expected to be higher for larger models. 

It might be possible to reduce these costs by sampling thinly at extreme values of the complexity and focusing on the range of complexity where the SMD appears.

\FloatBarrier  
\bibliography{main}
\clearpage
\appendix
\section{Extended Empirical Results \label{appfullresults}}
In this section, we present the full extended results of our empirical experiments. This section is divided into several subsections. In section \ref{app:litecomplexityselection}, we describe the complexity selection for Gemini-3.1-Flash-Lite. In section \ref{app:addaccmeasures}, we describe some additional measures of accuracy and cross-lingual gaps and present detailed tables of results. In section \ref{app:backtranslation}, we present results for the back-translation experiments described in section \ref{sec:mit_expts}. In section \ref{sec:thought_corr}, we present results showing the lack of clear correlations between thinking length and accuracy, as discussed in section \ref{sec:thought_corr}.

\subsection{Complexity selection for Gemini-3.1-Flash-Lite \label{app:litecomplexityselection}}
The complexity measures for Gemini-3-Flash, Gemma-4-31B, and GLM-5.1 were selected as described in section \ref{sec:expts}. But, since Gemini-3.1-Flash-Lite model is a smaller scale model than the other models in our evaluation set we chose to include some easier complexity levels for its evaluation to get the full accuracy curve starting from $100\%$. We present the complexity bounds in the Table~\ref{tab:lite-complexity}.  Note that DSA and Pr DSA had the same complexity levels as the default.  Separate complexity levels were selected for SLT, Graph SP and T. Rank. For SLT we chose to include easier levels starting from chain length of 10; for Graph SP, we decided to sample more questions from the complexity range of 70-200(total edges); and for T. Rank we decided to sample questions only from the range of 25-150(total edges).
\begin{table}[htbp]
\centering
\footnotesize
\begin{tabular}{l@{\hspace{10pt}}ccccc}
\toprule
\makecell[l]{\textbf{Puzzle}} & \textbf{Complexity Measure} & \textbf{$c_{\min}$} & \textbf{$c_{\max}$} \\
\midrule
\makecell[l]{SLT*} & Chain length & $10$ & $600$ \\
\midrule
\makecell[l]{DSA} & Number of states & $32$ & $2090$  \\ 
\midrule
\makecell[l]{Pr\ DSA} & Number of states & $9$ & $61$  \\
\midrule
\makecell[l]{Graph SP*} &  Number of edges & $70$ & $200$ \\
\midrule
\makecell[l]{T.\ Rank*} & Number of edges & $25$ & $150$  \\
\bottomrule
\end{tabular}
\caption{Complexity measure and range of complexity values for each puzzle for Gemini-3.1-Flash-Lite.}
\label{tab:lite-complexity}
\end{table}

\subsection{Additional Accuracy Measures \label{app:addaccmeasures}}
The SMD reported in the main text is a sufficient indicator of the presence of a cross-lingual gap. However, one might be concerned that it obscures cases where some language displays superior performance to English at one complexity and inferior performance at another. 

Therefore, it is convenient to define two supplementary measures. We define the average accuracy by
\begin{equation}
\text{Avg Acc}(\ell) = \frac{1}{c_{\max} - c_{\min}} \int_{c_{\min}}^{c_{\max}} f_\ell(c)\, dc.
\end{equation}
 We report the Average Accuracy in Table~\ref{tab:accuracy}.
 
We also define the {\em Reciprocal Divergence} (RD) by 
\begin{equation}
 \operatorname{RD}(\ell) = \min_c \text{sign}(\operatorname{SMD}(\ell)) \left( f_{\text{En}}(c) - f_\ell(c) \right).
 \end{equation}
 If the SMD is positive, which means that at the point of the largest cross-lingual gap English displays superior performance, the RD tells us the best relative-performance of the non-English language. If the SMD is negative (as in the case of GLM-5.1), the RD tells us the best relative-performance of English.

The RD is reported in Table \ref{tab:RD}. This Table shows the SMD is a good representative of the holistic performance, since the RD is insignificant in all cases. Therefore the peak difference in the accuracy is also an indicator of the difference in accuracy at other complexity values.

Finally, Tables \ref{tab:cpeak_gemini}, \ref{tab:cpeak_gemma}, \ref{tab:cpeak_flashlite} and \ref{tab:cpeak_glm51} expand the SMD results displayed in Table \ref{tab:merged_maxdiv} by presenting an estimate of the accuracy of English and the non-English languages at the point of maximum cross-lingual gap.  The values in this table are generated as follows. For all the metrics reported, we sample $N=300$ times from the observed distribution of accuracies using the posterior distribution of \cite{raju2026model} to generate the accuracy curve, both for English and the other language. Each sample, $i$, yields a value $c^{*}_i$ as defined in \eqref{cstardef}. Across these values of $c^*_i$, we now compute
\begin{equation}
\begin{split}
&\langle  f_{\text{En}} \rangle = {1 \over N} \sum_{i} f_{\text{En}}(c^*_i);\\
&\langle  f_{\ell} \rangle = {1 \over N} \sum_{i} f_{\ell}(c^*_i);\\
&\langle  c^{*} \rangle = {1 \over N} \sum_{i} c^*_i;\\
\end{split}
\end{equation}
together with their corresponding standard deviations.

\begin{table*}[htbp]
\centering
\normalsize

\begin{subtable}{1\linewidth}
\centering
\begin{tabular}{l@{\hspace{1pt}}ccccccc}
\toprule
\textbf{Task} & \textbf{EN} & \textbf{AR} & \textbf{HI} & \textbf{JA} & \textbf{TA} & \textbf{TE} & \textbf{ZH} \\
\midrule
SLT        & $45.18{\scriptstyle\pm1.13}$ & $41.18{\scriptstyle\pm1.17}$ & $43.70{\scriptstyle\pm1.13}$ & $43.58{\scriptstyle\pm1.31}$ & $42.45{\scriptstyle\pm1.30}$ & $40.30{\scriptstyle\pm1.33}$ & $45.65{\scriptstyle\pm1.26}$   \\
\midrule
DSA        & $52.88{\scriptstyle\pm1.74}$ & $40.52{\scriptstyle\pm1.95}$ & $29.61{\scriptstyle\pm1.48}$ & $33.08{\scriptstyle\pm1.54}$ & $34.81{\scriptstyle\pm1.90}$ & $33.32{\scriptstyle\pm2.13}$ & $41.75{\scriptstyle\pm2.03}$   \\
\midrule
Pr\ DSA    & $31.74{\scriptstyle\pm1.16}$ & $29.97{\scriptstyle\pm1.05}$ & $31.02{\scriptstyle\pm1.21}$ & $32.93{\scriptstyle\pm1.25}$ & $30.28{\scriptstyle\pm0.98}$ & $32.27{\scriptstyle\pm1.09}$ & $33.64{\scriptstyle\pm1.00}$   \\
\midrule
Graph SP   & $26.01{\scriptstyle\pm1.40}$ & $22.40{\scriptstyle\pm1.13}$ & $17.74{\scriptstyle\pm1.27}$ & $23.93{\scriptstyle\pm1.34}$ & $21.40{\scriptstyle\pm1.26}$ & $16.32{\scriptstyle\pm0.92}$ & $19.46{\scriptstyle\pm1.36}$   \\
\midrule
T.\ Rank   & $42.69{\scriptstyle\pm1.15}$ & $35.83{\scriptstyle\pm0.81}$ & $36.39{\scriptstyle\pm0.84}$ & $38.33{\scriptstyle\pm0.83}$ & $37.01{\scriptstyle\pm0.86}$ & $34.51{\scriptstyle\pm0.75}$ & $37.03{\scriptstyle\pm0.81}$   \\
\midrule
\bottomrule
\end{tabular}
\caption{\textbf{Gemini 3 Flash}}
\label{tab:auc_gemini_3_flash}
\end{subtable}

\vspace{1.5em}

\begin{subtable}{1\linewidth}
\centering
\begin{tabular}{l@{\hspace{1pt}}ccccccc}
\toprule
\textbf{Task} & \textbf{EN} & \textbf{AR} & \textbf{HI} & \textbf{JA} & \textbf{TA} & \textbf{TE} & \textbf{ZH} \\
\midrule
SLT        & $28.88{\scriptstyle\pm0.87}$ & $24.32{\scriptstyle\pm0.75}$ & $29.09{\scriptstyle\pm1.07}$ & $27.97{\scriptstyle\pm1.03}$ & $27.41{\scriptstyle\pm0.95}$ & $28.88{\scriptstyle\pm0.97}$ & $30.53{\scriptstyle\pm0.95}$   \\
\midrule
DSA        & $24.81{\scriptstyle\pm1.20}$ & $21.59{\scriptstyle\pm1.13}$ & $20.33{\scriptstyle\pm0.91}$ & $22.53{\scriptstyle\pm1.21}$ & $24.32{\scriptstyle\pm1.37}$ & $21.77{\scriptstyle\pm1.17}$ & $24.99{\scriptstyle\pm1.34}$   \\
\midrule
Pr\ DSA    & $30.14{\scriptstyle\pm1.21}$ & $32.84{\scriptstyle\pm1.31}$ & $29.51{\scriptstyle\pm1.21}$ & $30.30{\scriptstyle\pm1.05}$ & $29.45{\scriptstyle\pm1.12}$ & $31.55{\scriptstyle\pm1.15}$ & $28.27{\scriptstyle\pm1.25}$   \\
\midrule
Graph SP   & $17.03{\scriptstyle\pm0.92}$ & $17.92{\scriptstyle\pm1.07}$ & $18.81{\scriptstyle\pm1.04}$ & $19.29{\scriptstyle\pm1.07}$ & $16.11{\scriptstyle\pm1.07}$ & $14.99{\scriptstyle\pm1.16}$ & $16.00{\scriptstyle\pm1.06}$   \\
\midrule
T.\ Rank   & $23.39{\scriptstyle\pm1.11}$ & $19.09{\scriptstyle\pm0.94}$ & $20.31{\scriptstyle\pm0.89}$ & $23.19{\scriptstyle\pm0.89}$ & $19.13{\scriptstyle\pm0.96}$ & $19.29{\scriptstyle\pm1.06}$ & $21.34{\scriptstyle\pm0.92}$   \\
\midrule
\bottomrule
\end{tabular}
\caption{\textbf{Gemma 4 31B IT}}
\label{tab:auc_gemma_4_31b_it}
\end{subtable}

\vspace{1.5em}

\begin{subtable}{1\linewidth}
\centering
\begin{tabular}{l@{\hspace{1pt}}ccccccc}
\toprule
\textbf{Task} & \textbf{EN} & \textbf{AR} & \textbf{HI} & \textbf{JA} & \textbf{TA} & \textbf{TE} & \textbf{ZH} \\
\midrule
SLT*       & $27.84{\scriptstyle\pm0.95}$ & $16.84{\scriptstyle\pm0.60}$ & $20.77{\scriptstyle\pm0.71}$ & $24.48{\scriptstyle\pm0.85}$ & $23.13{\scriptstyle\pm0.76}$ & $22.53{\scriptstyle\pm0.96}$ & $26.35{\scriptstyle\pm0.89}$   \\
\midrule
DSA        & $18.83{\scriptstyle\pm0.93}$ & $16.10{\scriptstyle\pm0.93}$ & $17.96{\scriptstyle\pm1.02}$ & $17.36{\scriptstyle\pm1.07}$ & $13.91{\scriptstyle\pm0.83}$ & $13.33{\scriptstyle\pm0.79}$ & $16.34{\scriptstyle\pm0.84}$   \\
\midrule
Pr\ DSA    & $21.04{\scriptstyle\pm0.91}$ & $19.56{\scriptstyle\pm0.83}$ & $17.12{\scriptstyle\pm0.84}$ & $17.65{\scriptstyle\pm0.86}$ & $20.32{\scriptstyle\pm0.86}$ & $18.67{\scriptstyle\pm0.77}$ & $18.48{\scriptstyle\pm0.91}$   \\
\midrule
Graph SP*  & $14.71{\scriptstyle\pm1.47}$ & $15.59{\scriptstyle\pm1.15}$ & $16.25{\scriptstyle\pm1.00}$ & $15.82{\scriptstyle\pm0.89}$ & $16.26{\scriptstyle\pm0.90}$ & $17.50{\scriptstyle\pm1.08}$ & $15.96{\scriptstyle\pm1.20}$   \\
\midrule
T.\ Rank*  & $52.93{\scriptstyle\pm0.90}$ & $45.21{\scriptstyle\pm0.70}$ & $45.73{\scriptstyle\pm0.76}$ & $45.45{\scriptstyle\pm0.80}$ & $47.93{\scriptstyle\pm0.81}$ & $47.09{\scriptstyle\pm0.92}$ & $47.91{\scriptstyle\pm1.02}$   \\
\midrule
\bottomrule
\end{tabular}
\caption{\textbf{Gemini 3.1 Flash Lite}}
\label{tab:auc_flash_lite}
\end{subtable}

\vspace{1.5em}

\begin{subtable}{1\linewidth}
\centering
\begin{tabular}{l@{\hspace{1pt}}ccccccc}
\toprule
\textbf{Task} & \textbf{EN} & \textbf{AR} & \textbf{HI} & \textbf{JA} & \textbf{TA} & \textbf{TE} & \textbf{ZH} \\
\midrule
SLT        & $32.28{\scriptstyle\pm0.93}$ & $30.06{\scriptstyle\pm0.87}$ & $30.13{\scriptstyle\pm0.94}$ & $31.44{\scriptstyle\pm1.07}$ & $27.73{\scriptstyle\pm0.86}$ & $28.99{\scriptstyle\pm0.76}$ & $30.20{\scriptstyle\pm0.90}$   \\
\midrule
DSA        & $24.95{\scriptstyle\pm1.07}$ & $21.68{\scriptstyle\pm0.99}$ & $19.12{\scriptstyle\pm0.91}$ & $25.89{\scriptstyle\pm0.97}$ & $15.05{\scriptstyle\pm0.79}$ & $15.37{\scriptstyle\pm0.76}$ & $24.48{\scriptstyle\pm1.14}$   \\
\midrule
Pr\ DSA    & $44.15{\scriptstyle\pm1.61}$ & $39.36{\scriptstyle\pm1.49}$ & $37.25{\scriptstyle\pm1.58}$ & $39.52{\scriptstyle\pm1.51}$ & $33.30{\scriptstyle\pm1.25}$ & $36.13{\scriptstyle\pm1.23}$ & $42.26{\scriptstyle\pm1.57}$   \\
\midrule
Graph SP   & $32.88{\scriptstyle\pm1.14}$ & $34.91{\scriptstyle\pm1.58}$ & $37.40{\scriptstyle\pm1.49}$ & $32.99{\scriptstyle\pm1.52}$ & $35.62{\scriptstyle\pm1.47}$ & $36.70{\scriptstyle\pm1.59}$ & $45.27{\scriptstyle\pm1.52}$   \\
\midrule
T.\ Rank   & $40.75{\scriptstyle\pm1.06}$ & $40.07{\scriptstyle\pm1.00}$ & $40.68{\scriptstyle\pm1.32}$ & $34.81{\scriptstyle\pm1.16}$ & $36.78{\scriptstyle\pm0.94}$ & $38.74{\scriptstyle\pm1.02}$ & $39.93{\scriptstyle\pm1.19}$   \\
\midrule
\bottomrule
\end{tabular}
\caption{\textbf{GLM-5-1}}
\label{tab:auc_glm-5-1}
\end{subtable}

\vspace{1.5em}
\caption{Average Accuracy (\%) as calculated from the gamma fits. Higher is better. * denotes custom complexity levels used as described in Appendix \ref{app:litecomplexityselection}. }
\label{tab:accuracy}
\end{table*}

\begin{table*}
    \centering
    \normalsize

\begin{subtable}{1\linewidth}
\centering
\begin{tabular}{l@{\hspace{10pt}}cccccc}
\toprule
\textbf{Task} & \textbf{AR} & \textbf{HI} & \textbf{JA} & \textbf{TA} & \textbf{TE} & \textbf{ZH} \\
\midrule
SLT        & $-0.01{\scriptstyle\pm1.42}$ & $0.00{\scriptstyle\pm1.22}$ & $-1.49{\scriptstyle\pm1.46}$ & $-0.83{\scriptstyle\pm1.30}$ & $-1.44{\scriptstyle\pm1.42}$ & $-1.66{\scriptstyle\pm1.39}$   \\
\midrule
DSA        & $0.00{\scriptstyle\pm2.25}$ & $0.00{\scriptstyle\pm0.00}$ & $0.00{\scriptstyle\pm0.43}$ & $0.00{\scriptstyle\pm1.64}$ & $0.00{\scriptstyle\pm0.59}$ & $0.00{\scriptstyle\pm2.27}$   \\
\midrule
Pr\ DSA    & $0.00{\scriptstyle\pm1.06}$ & $-0.87{\scriptstyle\pm1.45}$ & $-0.04{\scriptstyle\pm1.24}$ & $0.00{\scriptstyle\pm1.08}$ & $-1.22{\scriptstyle\pm1.36}$ & $-0.35{\scriptstyle\pm1.27}$   \\
\midrule
Graph SP   & $0.04{\scriptstyle\pm1.11}$ & $0.19{\scriptstyle\pm0.71}$ & $0.08{\scriptstyle\pm0.92}$ & $-0.62{\scriptstyle\pm1.48}$ & $0.23{\scriptstyle\pm0.73}$ & $-0.55{\scriptstyle\pm1.66}$   \\
\midrule
T.\ Rank   & $0.00{\scriptstyle\pm0.13}$ & $0.00{\scriptstyle\pm0.20}$ & $0.00{\scriptstyle\pm0.37}$ & $0.00{\scriptstyle\pm0.20}$ & $0.00{\scriptstyle\pm0.03}$ & $-0.00{\scriptstyle\pm0.26}$   \\
\midrule
\bottomrule
\end{tabular}
\caption{\textbf{Gemini 3 Flash}}
\label{tab:negdiv_gemini_3_flash}
\end{subtable}

\vspace{1.5em}

\begin{subtable}{1\linewidth}
\centering
\begin{tabular}{l@{\hspace{10pt}}cccccc}
\toprule
\textbf{Task} & \textbf{AR} & \textbf{HI} & \textbf{JA} & \textbf{TA} & \textbf{TE} & \textbf{ZH} \\
\midrule
SLT        & $0.00{\scriptstyle\pm0.93}$ & $-2.14{\scriptstyle\pm2.32}$ & $-0.02{\scriptstyle\pm1.68}$ & $-1.37{\scriptstyle\pm1.74}$ & $-1.62{\scriptstyle\pm2.12}$ & $-0.40{\scriptstyle\pm2.38}$   \\
\midrule
DSA        & $0.00{\scriptstyle\pm0.97}$ & $0.00{\scriptstyle\pm0.75}$ & $-0.10{\scriptstyle\pm1.36}$ & $-0.30{\scriptstyle\pm1.80}$ & $0.00{\scriptstyle\pm0.78}$ & $-0.69{\scriptstyle\pm1.63}$   \\
\midrule
Pr\ DSA    & $-0.01{\scriptstyle\pm1.41}$ & $-2.35{\scriptstyle\pm1.52}$ & $-2.02{\scriptstyle\pm1.22}$ & $-3.50{\scriptstyle\pm1.57}$ & $-0.00{\scriptstyle\pm1.12}$ & $-0.11{\scriptstyle\pm0.96}$   \\
\midrule
Graph SP   & $-0.12{\scriptstyle\pm0.86}$ & $-1.71{\scriptstyle\pm1.40}$ & $-1.02{\scriptstyle\pm1.52}$ & $-0.37{\scriptstyle\pm0.86}$ & $0.12{\scriptstyle\pm0.67}$ & $-0.05{\scriptstyle\pm0.66}$   \\
\midrule
T.\ Rank   & $0.05{\scriptstyle\pm0.89}$ & $0.05{\scriptstyle\pm1.79}$ & $-2.85{\scriptstyle\pm2.34}$ & $0.05{\scriptstyle\pm3.12}$ & $0.05{\scriptstyle\pm1.96}$ & $0.04{\scriptstyle\pm1.45}$   \\
\midrule
\bottomrule
\end{tabular}
\caption{\textbf{Gemma 4 31B IT}}
\label{tab:negdiv_gemma_4_31b_it}
\end{subtable}

\vspace{1.5em}

\begin{subtable}{1\linewidth}
\centering
\begin{tabular}{l@{\hspace{10pt}}cccccc}
\toprule
\textbf{Task} & \textbf{AR} & \textbf{HI} & \textbf{JA} & \textbf{TA} & \textbf{TE} & \textbf{ZH} \\
\midrule
SLT*       & $0.00{\scriptstyle\pm0.15}$ & $-0.29{\scriptstyle\pm0.88}$ & $-1.58{\scriptstyle\pm1.66}$ & $-0.28{\scriptstyle\pm0.94}$ & $-0.31{\scriptstyle\pm1.09}$ & $0.00{\scriptstyle\pm1.50}$   \\
\midrule
DSA        & $0.00{\scriptstyle\pm0.81}$ & $-0.35{\scriptstyle\pm1.57}$ & $-0.00{\scriptstyle\pm1.53}$ & $0.00{\scriptstyle\pm0.49}$ & $0.00{\scriptstyle\pm0.50}$ & $-0.43{\scriptstyle\pm1.11}$   \\
\midrule
Pr\ DSA    & $-0.00{\scriptstyle\pm1.51}$ & $0.11{\scriptstyle\pm0.47}$ & $0.09{\scriptstyle\pm1.01}$ & $-3.16{\scriptstyle\pm2.02}$ & $-0.00{\scriptstyle\pm1.53}$ & $-0.00{\scriptstyle\pm0.59}$   \\
\midrule
Graph SP*  & $0.00{\scriptstyle\pm2.64}$ & $-0.17{\scriptstyle\pm2.13}$ & $0.00{\scriptstyle\pm2.40}$ & $0.00{\scriptstyle\pm2.05}$ & $-0.48{\scriptstyle\pm2.05}$ & $0.00{\scriptstyle\pm2.25}$   \\
\midrule
T.\ Rank*  & $-0.00{\scriptstyle\pm0.44}$ & $0.00{\scriptstyle\pm0.07}$ & $0.00{\scriptstyle\pm0.15}$ & $0.00{\scriptstyle\pm0.27}$ & $-0.04{\scriptstyle\pm0.83}$ & $0.00{\scriptstyle\pm0.62}$   \\
\midrule
\bottomrule
\end{tabular}
\caption{\textbf{Gemini 3.1 Flash Lite}}
\label{tab:negdiv_flash_lite}
\end{subtable}

\vspace{1.5em}

\begin{subtable}{1\linewidth}
\centering
\begin{tabular}{l@{\hspace{10pt}}cccccc}
\toprule
\textbf{Task} & \textbf{AR} & \textbf{HI} & \textbf{JA} & \textbf{TA} & \textbf{TE} & \textbf{ZH} \\
\midrule
SLT        & $0.00{\scriptstyle\pm0.64}$ & $0.00{\scriptstyle\pm1.30}$ & $-0.64{\scriptstyle\pm0.87}$ & $0.00{\scriptstyle\pm0.28}$ & $-0.84{\scriptstyle\pm2.50}$ & $0.00{\scriptstyle\pm0.64}$   \\
\midrule
DSA        & $0.00{\scriptstyle\pm0.77}$ & $0.00{\scriptstyle\pm0.68}$ & $-0.28{\scriptstyle\pm0.96}$ & $0.00{\scriptstyle\pm0.38}$ & $0.00{\scriptstyle\pm0.00}$ & $-0.77{\scriptstyle\pm1.53}$   \\
\midrule
Pr\ DSA    & $0.01{\scriptstyle\pm1.22}$ & $0.02{\scriptstyle\pm1.45}$ & $0.06{\scriptstyle\pm1.18}$ & $0.02{\scriptstyle\pm0.67}$ & $0.05{\scriptstyle\pm0.94}$ & $-1.48{\scriptstyle\pm1.85}$   \\
\midrule
Graph SP   & $-0.06{\scriptstyle\pm1.05}$ & $-2.97{\scriptstyle\pm2.80}$ & $-1.92{\scriptstyle\pm1.33}$ & $0.00{\scriptstyle\pm1.05}$ & $0.00{\scriptstyle\pm1.03}$ & $0.00{\scriptstyle\pm0.05}$   \\
\midrule
T.\ Rank   & $-0.26{\scriptstyle\pm1.18}$ & $-1.99{\scriptstyle\pm1.71}$ & $0.03{\scriptstyle\pm0.49}$ & $0.00{\scriptstyle\pm0.63}$ & $-0.00{\scriptstyle\pm0.94}$ & $0.00{\scriptstyle\pm1.10}$   \\
\midrule
\bottomrule
\end{tabular}
\caption{\textbf{GLM-5-1}}
\label{tab:negdiv_glm-5-1}
\end{subtable}

\vspace{1.5em}
\caption{Reciprocal Divergence(\%). }
\label{tab:RD}
\end{table*}

\begin{table*}
    \centering
    \normalsize
\begin{tabular}{ll@{\hspace{6pt}}cccccc}
\toprule
\textbf{Task} & \textbf{Metric} & \textbf{AR} & \textbf{HI} & \textbf{JA} & \textbf{TA} & \textbf{TE} & \textbf{ZH} \\
\midrule
\multirow{3}{*}{SLT}
  & $c^*$ & $326{\scriptstyle\pm35}$ & $303{\scriptstyle\pm52}$ & $254{\scriptstyle\pm36}$ & $258{\scriptstyle\pm25}$ & $251{\scriptstyle\pm12}$ & $410{\scriptstyle\pm54}$ \\
  & $f_{\mathrm{EN}}$ & $41.4{\scriptstyle\pm15.7}$ & $52.6{\scriptstyle\pm22.0}$ & $78.9{\scriptstyle\pm16.3}$ & $76.9{\scriptstyle\pm12.0}$ & $80.4{\scriptstyle\pm5.9}$ & $15.6{\scriptstyle\pm23.6}$ \\
  & $f_{\ell}$ & $31.9{\scriptstyle\pm15.4}$ & $49.1{\scriptstyle\pm23.0}$ & $70.3{\scriptstyle\pm14.3}$ & $66.1{\scriptstyle\pm10.7}$ & $61.8{\scriptstyle\pm5.2}$ & $17.4{\scriptstyle\pm21.7}$ \\
\midrule
\multirow{3}{*}{DSA}
  & $c^*$ & $721{\scriptstyle\pm73}$ & $667{\scriptstyle\pm32}$ & $651{\scriptstyle\pm30}$ & $680{\scriptstyle\pm45}$ & $692{\scriptstyle\pm47}$ & $704{\scriptstyle\pm80}$ \\
  & $f_{\mathrm{EN}}$ & $79.8{\scriptstyle\pm7.5}$ & $85.6{\scriptstyle\pm4.8}$ & $87.3{\scriptstyle\pm4.0}$ & $84.3{\scriptstyle\pm4.7}$ & $83.0{\scriptstyle\pm5.9}$ & $81.6{\scriptstyle\pm8.2}$ \\
  & $f_{\ell}$ & $50.1{\scriptstyle\pm5.1}$ & $32.1{\scriptstyle\pm3.9}$ & $39.5{\scriptstyle\pm3.3}$ & $41.6{\scriptstyle\pm4.2}$ & $37.9{\scriptstyle\pm5.3}$ & $53.2{\scriptstyle\pm5.9}$ \\
\midrule
\multirow{3}{*}{Pr\ DSA}
  & $c^*$ & $21{\scriptstyle\pm4}$ & $19{\scriptstyle\pm5}$ & $21{\scriptstyle\pm5}$ & $22{\scriptstyle\pm4}$ & $33{\scriptstyle\pm5}$ & $21{\scriptstyle\pm3}$ \\
  & $f_{\mathrm{EN}}$ & $67.8{\scriptstyle\pm18.5}$ & $77.5{\scriptstyle\pm21.9}$ & $64.5{\scriptstyle\pm21.1}$ & $57.4{\scriptstyle\pm20.9}$ & $16.6{\scriptstyle\pm25.6}$ & $65.8{\scriptstyle\pm15.7}$ \\
  & $f_{\ell}$ & $61.7{\scriptstyle\pm16.8}$ & $72.1{\scriptstyle\pm19.2}$ & $68.9{\scriptstyle\pm22.4}$ & $53.5{\scriptstyle\pm19.7}$ & $18.1{\scriptstyle\pm22.6}$ & $74.3{\scriptstyle\pm17.1}$ \\
\midrule
\multirow{3}{*}{Graph SP}
  & $c^*$ & $211{\scriptstyle\pm53}$ & $192{\scriptstyle\pm22}$ & $195{\scriptstyle\pm60}$ & $245{\scriptstyle\pm51}$ & $189{\scriptstyle\pm18}$ & $230{\scriptstyle\pm36}$ \\
  & $f_{\mathrm{EN}}$ & $42.5{\scriptstyle\pm17.4}$ & $49.2{\scriptstyle\pm9.2}$ & $48.2{\scriptstyle\pm19.9}$ & $32.8{\scriptstyle\pm16.2}$ & $50.7{\scriptstyle\pm8.2}$ & $36.5{\scriptstyle\pm11.4}$ \\
  & $f_{\ell}$ & $36.0{\scriptstyle\pm19.2}$ & $32.3{\scriptstyle\pm8.2}$ & $44.3{\scriptstyle\pm20.0}$ & $24.5{\scriptstyle\pm17.8}$ & $30.3{\scriptstyle\pm6.4}$ & $24.1{\scriptstyle\pm12.7}$ \\
\midrule
\multirow{3}{*}{T.\ Rank}
  & $c^*$ & $212{\scriptstyle\pm6}$ & $210{\scriptstyle\pm6}$ & $212{\scriptstyle\pm8}$ & $211{\scriptstyle\pm7}$ & $208{\scriptstyle\pm4}$ & $209{\scriptstyle\pm6}$ \\
  & $f_{\mathrm{EN}}$ & $68.9{\scriptstyle\pm9.1}$ & $71.4{\scriptstyle\pm9.2}$ & $68.9{\scriptstyle\pm11.6}$ & $69.7{\scriptstyle\pm10.1}$ & $73.8{\scriptstyle\pm7.6}$ & $72.2{\scriptstyle\pm9.3}$ \\
  & $f_{\ell}$ & $39.0{\scriptstyle\pm7.0}$ & $43.9{\scriptstyle\pm7.2}$ & $49.4{\scriptstyle\pm10.2}$ & $44.9{\scriptstyle\pm8.5}$ & $38.6{\scriptstyle\pm5.6}$ & $47.2{\scriptstyle\pm7.9}$ \\
\bottomrule
\end{tabular}
\caption{\textbf{Gemini 3 Flash.} Here we present the $\langle c^*\rangle$  at which SMD values were reported along with $\langle f_{EN}\rangle$ and $\langle f_\ell\rangle$ at the respective complexities over 300 bootstraps.}
\label{tab:cpeak_gemini}
\end{table*}

\begin{table*}
    \centering
    \normalsize
\begin{tabular}{ll@{\hspace{6pt}}cccccc}
\toprule
\textbf{Task} & \textbf{Metric} & \textbf{AR} & \textbf{HI} & \textbf{JA} & \textbf{TA} & \textbf{TE} & \textbf{ZH} \\
\midrule
\multirow{3}{*}{SLT}
  & $c^*$ & $155{\scriptstyle\pm15}$ & $131{\scriptstyle\pm35}$ & $146{\scriptstyle\pm30}$ & $133{\scriptstyle\pm23}$ & $132{\scriptstyle\pm36}$ & $204{\scriptstyle\pm40}$ \\
  & $f_{\mathrm{EN}}$ & $59.6{\scriptstyle\pm13.1}$ & $81.8{\scriptstyle\pm24.0}$ & $68.2{\scriptstyle\pm20.3}$ & $79.8{\scriptstyle\pm15.6}$ & $80.8{\scriptstyle\pm24.0}$ & $25.8{\scriptstyle\pm25.5}$ \\
  & $f_{\ell}$ & $38.6{\scriptstyle\pm10.2}$ & $77.1{\scriptstyle\pm18.4}$ & $62.8{\scriptstyle\pm20.0}$ & $65.6{\scriptstyle\pm12.1}$ & $76.0{\scriptstyle\pm20.1}$ & $31.1{\scriptstyle\pm22.2}$ \\
\midrule
\multirow{3}{*}{DSA}
  & $c^*$ & $309{\scriptstyle\pm57}$ & $346{\scriptstyle\pm47}$ & $285{\scriptstyle\pm53}$ & $285{\scriptstyle\pm102}$ & $312{\scriptstyle\pm54}$ & $260{\scriptstyle\pm103}$ \\
  & $f_{\mathrm{EN}}$ & $65.5{\scriptstyle\pm15.3}$ & $53.6{\scriptstyle\pm13.4}$ & $74.0{\scriptstyle\pm14.2}$ & $74.0{\scriptstyle\pm23.9}$ & $64.5{\scriptstyle\pm14.8}$ & $82.3{\scriptstyle\pm23.2}$ \\
  & $f_{\ell}$ & $52.1{\scriptstyle\pm14.0}$ & $37.2{\scriptstyle\pm12.0}$ & $61.1{\scriptstyle\pm12.4}$ & $69.7{\scriptstyle\pm21.7}$ & $51.8{\scriptstyle\pm13.8}$ & $81.2{\scriptstyle\pm23.2}$ \\
\midrule
\multirow{3}{*}{Pr\ DSA}
  & $c^*$ & $26{\scriptstyle\pm5}$ & $18{\scriptstyle\pm4}$ & $18{\scriptstyle\pm4}$ & $18{\scriptstyle\pm3}$ & $25{\scriptstyle\pm4}$ & $20{\scriptstyle\pm4}$ \\
  & $f_{\mathrm{EN}}$ & $33.2{\scriptstyle\pm21.2}$ & $80.9{\scriptstyle\pm18.1}$ & $82.1{\scriptstyle\pm20.6}$ & $83.4{\scriptstyle\pm14.4}$ & $36.0{\scriptstyle\pm21.2}$ & $71.5{\scriptstyle\pm18.4}$ \\
  & $f_{\ell}$ & $39.3{\scriptstyle\pm19.6}$ & $71.7{\scriptstyle\pm14.7}$ & $77.6{\scriptstyle\pm18.3}$ & $70.8{\scriptstyle\pm11.5}$ & $39.3{\scriptstyle\pm19.7}$ & $63.9{\scriptstyle\pm16.9}$ \\
\midrule
\multirow{3}{*}{Graph SP}
  & $c^*$ & $199{\scriptstyle\pm47}$ & $234{\scriptstyle\pm60}$ & $220{\scriptstyle\pm58}$ & $114{\scriptstyle\pm27}$ & $125{\scriptstyle\pm24}$ & $119{\scriptstyle\pm31}$ \\
  & $f_{\mathrm{EN}}$ & $27.8{\scriptstyle\pm22.3}$ & $18.9{\scriptstyle\pm27.2}$ & $22.0{\scriptstyle\pm25.6}$ & $74.7{\scriptstyle\pm13.8}$ & $66.9{\scriptstyle\pm13.3}$ & $70.8{\scriptstyle\pm16.0}$ \\
  & $f_{\ell}$ & $29.7{\scriptstyle\pm20.5}$ & $22.4{\scriptstyle\pm23.0}$ & $26.5{\scriptstyle\pm22.2}$ & $67.7{\scriptstyle\pm12.3}$ & $58.9{\scriptstyle\pm12.5}$ & $65.4{\scriptstyle\pm15.0}$ \\
\midrule
\multirow{3}{*}{T.\ Rank}
  & $c^*$ & $160{\scriptstyle\pm13}$ & $163{\scriptstyle\pm17}$ & $138{\scriptstyle\pm25}$ & $161{\scriptstyle\pm15}$ & $161{\scriptstyle\pm14}$ & $169{\scriptstyle\pm21}$ \\
  & $f_{\mathrm{EN}}$ & $53.5{\scriptstyle\pm14.4}$ & $50.0{\scriptstyle\pm17.4}$ & $79.0{\scriptstyle\pm22.5}$ & $52.8{\scriptstyle\pm15.3}$ & $52.8{\scriptstyle\pm15.2}$ & $44.1{\scriptstyle\pm20.2}$ \\
  & $f_{\ell}$ & $39.4{\scriptstyle\pm12.2}$ & $40.2{\scriptstyle\pm17.4}$ & $83.6{\scriptstyle\pm27.6}$ & $38.9{\scriptstyle\pm16.1}$ & $39.4{\scriptstyle\pm14.6}$ & $37.7{\scriptstyle\pm20.3}$ \\
\bottomrule
\end{tabular}
\caption{\textbf{Gemma 4 31B IT.} Here we present the $\langle c^*\rangle$  at which SMD values were reported along with $\langle f_{EN}\rangle$ and $\langle f_\ell\rangle$ at the respective complexities over 300 bootstraps.}
\label{tab:cpeak_gemma}
\end{table*}

\begin{table*}
\centering
\normalsize
\begin{tabular}{ll@{\hspace{6pt}}cccccc}
\toprule
\textbf{Task} & \textbf{Metric} & \textbf{AR} & \textbf{HI} & \textbf{JA} & \textbf{TA} & \textbf{TE} & \textbf{ZH} \\
\midrule
\multirow{3}{*}{SLT}
  & $c^*$ & $122{\scriptstyle\pm4}$ & $126{\scriptstyle\pm5}$ & $128{\scriptstyle\pm5}$ & $131{\scriptstyle\pm6}$ & $129{\scriptstyle\pm6}$ & $147{\scriptstyle\pm26}$ \\
  & $f_{\mathrm{EN}}$ & $88.0{\scriptstyle\pm3.7}$ & $85.2{\scriptstyle\pm3.8}$ & $83.2{\scriptstyle\pm4.4}$ & $81.2{\scriptstyle\pm6.5}$ & $82.2{\scriptstyle\pm6.3}$ & $65.1{\scriptstyle\pm20.1}$ \\
  & $f_{\ell}$ & $26.8{\scriptstyle\pm3.9}$ & $41.6{\scriptstyle\pm3.9}$ & $55.5{\scriptstyle\pm5.1}$ & $50.6{\scriptstyle\pm4.9}$ & $48.0{\scriptstyle\pm5.2}$ & $56.8{\scriptstyle\pm19.4}$ \\
\midrule
\multirow{3}{*}{DSA}
  & $c^*$ & $271{\scriptstyle\pm55}$ & $461{\scriptstyle\pm104}$ & $370{\scriptstyle\pm105}$ & $288{\scriptstyle\pm40}$ & $242{\scriptstyle\pm11}$ & $242{\scriptstyle\pm24}$ \\
  & $f_{\mathrm{EN}}$ & $70.0{\scriptstyle\pm14.0}$ & $27.3{\scriptstyle\pm21.1}$ & $42.8{\scriptstyle\pm22.0}$ & $64.8{\scriptstyle\pm11.6}$ & $79.4{\scriptstyle\pm4.2}$ & $79.4{\scriptstyle\pm6.7}$ \\
  & $f_{\ell}$ & $55.7{\scriptstyle\pm11.6}$ & $24.8{\scriptstyle\pm22.6}$ & $38.0{\scriptstyle\pm22.5}$ & $42.8{\scriptstyle\pm9.0}$ & $47.9{\scriptstyle\pm3.8}$ & $59.4{\scriptstyle\pm5.5}$ \\
\midrule
\multirow{3}{*}{Pr\ DSA}
  & $c^*$ & $20{\scriptstyle\pm4}$ & $17{\scriptstyle\pm2}$ & $17{\scriptstyle\pm2}$ & $14{\scriptstyle\pm4}$ & $20{\scriptstyle\pm3}$ & $16{\scriptstyle\pm2}$ \\
  & $f_{\mathrm{EN}}$ & $39.4{\scriptstyle\pm20.4}$ & $61.1{\scriptstyle\pm12.2}$ & $61.1{\scriptstyle\pm15.4}$ & $82.5{\scriptstyle\pm22.7}$ & $41.3{\scriptstyle\pm16.8}$ & $69.5{\scriptstyle\pm14.8}$ \\
  & $f_{\ell}$ & $34.9{\scriptstyle\pm22.4}$ & $47.0{\scriptstyle\pm10.2}$ & $48.9{\scriptstyle\pm13.3}$ & $85.7{\scriptstyle\pm28.0}$ & $34.0{\scriptstyle\pm17.9}$ & $58.6{\scriptstyle\pm13.0}$ \\
\midrule
\multirow{3}{*}{Graph SP}
  & $c^*$ & $89{\scriptstyle\pm9}$ & $95{\scriptstyle\pm9}$ & $91{\scriptstyle\pm10}$ & $92{\scriptstyle\pm10}$ & $96{\scriptstyle\pm9}$ & $94{\scriptstyle\pm9}$ \\
  & $f_{\mathrm{EN}}$ & $45.9{\scriptstyle\pm23.5}$ & $27.7{\scriptstyle\pm24.1}$ & $38.5{\scriptstyle\pm24.4}$ & $36.9{\scriptstyle\pm24.1}$ & $25.2{\scriptstyle\pm22.6}$ & $30.4{\scriptstyle\pm23.7}$ \\
  & $f_{\ell}$ & $49.6{\scriptstyle\pm25.3}$ & $33.9{\scriptstyle\pm22.4}$ & $42.9{\scriptstyle\pm25.1}$ & $43.1{\scriptstyle\pm24.4}$ & $35.7{\scriptstyle\pm20.4}$ & $35.4{\scriptstyle\pm24.3}$ \\
\midrule
\multirow{3}{*}{T.\ Rank}
  & $c^*$ & $87{\scriptstyle\pm2}$ & $86{\scriptstyle\pm3}$ & $84{\scriptstyle\pm3}$ & $85{\scriptstyle\pm4}$ & $90{\scriptstyle\pm3}$ & $87{\scriptstyle\pm4}$ \\
  & $f_{\mathrm{EN}}$ & $53.9{\scriptstyle\pm6.8}$ & $57.3{\scriptstyle\pm8.0}$ & $62.1{\scriptstyle\pm7.7}$ & $59.4{\scriptstyle\pm10.2}$ & $47.3{\scriptstyle\pm9.2}$ & $53.9{\scriptstyle\pm10.6}$ \\
  & $f_{\ell}$ & $26.5{\scriptstyle\pm6.4}$ & $32.9{\scriptstyle\pm7.0}$ & $37.0{\scriptstyle\pm7.5}$ & $42.9{\scriptstyle\pm9.2}$ & $26.6{\scriptstyle\pm9.0}$ & $37.4{\scriptstyle\pm10.8}$ \\
\bottomrule
\end{tabular}
\caption{\textbf{Gemini 3.1 Flash Lite.} Here we present the $\langle c^*\rangle$  at which SMD values were reported along with $\langle f_{EN}\rangle$ and $\langle f_\ell\rangle$ at the respective complexities over 300 bootstraps.}
\label{tab:cpeak_flashlite}
\end{table*}

\begin{table*}
\centering
\normalsize
\begin{tabular}{ll@{\hspace{6pt}}cccccc}
\toprule
\textbf{Task} & \textbf{Metric} & \textbf{AR} & \textbf{HI} & \textbf{JA} & \textbf{TA} & \textbf{TE} & \textbf{ZH} \\
\midrule
\multirow{3}{*}{SLT}
  & $c^*$ & $167{\scriptstyle\pm24}$ & $167{\scriptstyle\pm28}$ & $151{\scriptstyle\pm25}$ & $170{\scriptstyle\pm15}$ & $207{\scriptstyle\pm21}$ & $163{\scriptstyle\pm27}$ \\
  & $f_{\mathrm{EN}}$ & $62.7{\scriptstyle\pm15.5}$ & $62.7{\scriptstyle\pm16.8}$ & $75.9{\scriptstyle\pm15.8}$ & $60.9{\scriptstyle\pm11.1}$ & $35.2{\scriptstyle\pm12.5}$ & $66.5{\scriptstyle\pm16.5}$ \\
  & $f_{\ell}$ & $53.6{\scriptstyle\pm13.8}$ & $53.9{\scriptstyle\pm16.2}$ & $68.9{\scriptstyle\pm13.7}$ & $42.8{\scriptstyle\pm8.1}$ & $22.6{\scriptstyle\pm13.8}$ & $57.3{\scriptstyle\pm14.8}$ \\
\midrule
\multirow{3}{*}{DSA}
  & $c^*$ & $302{\scriptstyle\pm51}$ & $296{\scriptstyle\pm33}$ & $282{\scriptstyle\pm72}$ & $259{\scriptstyle\pm24}$ & $256{\scriptstyle\pm10}$ & $256{\scriptstyle\pm83}$ \\
  & $f_{\mathrm{EN}}$ & $60.4{\scriptstyle\pm14.9}$ & $62.4{\scriptstyle\pm11.5}$ & $67.5{\scriptstyle\pm17.0}$ & $75.8{\scriptstyle\pm9.0}$ & $76.8{\scriptstyle\pm5.3}$ & $76.8{\scriptstyle\pm20.7}$ \\
  & $f_{\ell}$ & $48.1{\scriptstyle\pm13.7}$ & $39.7{\scriptstyle\pm9.1}$ & $74.3{\scriptstyle\pm18.9}$ & $31.9{\scriptstyle\pm7.1}$ & $34.2{\scriptstyle\pm4.5}$ & $70.7{\scriptstyle\pm17.5}$ \\
\midrule
\multirow{3}{*}{Pr\ DSA}
  & $c^*$ & $30{\scriptstyle\pm8}$ & $30{\scriptstyle\pm7}$ & $27{\scriptstyle\pm8}$ & $29{\scriptstyle\pm4}$ & $28{\scriptstyle\pm5}$ & $43{\scriptstyle\pm11}$ \\
  & $f_{\mathrm{EN}}$ & $42.5{\scriptstyle\pm18.3}$ & $44.7{\scriptstyle\pm16.1}$ & $50.2{\scriptstyle\pm19.0}$ & $46.7{\scriptstyle\pm10.4}$ & $47.2{\scriptstyle\pm12.4}$ & $22.9{\scriptstyle\pm23.6}$ \\
  & $f_{\ell}$ & $35.8{\scriptstyle\pm18.1}$ & $34.9{\scriptstyle\pm16.0}$ & $43.6{\scriptstyle\pm18.5}$ & $30.8{\scriptstyle\pm9.1}$ & $35.7{\scriptstyle\pm11.3}$ & $19.5{\scriptstyle\pm27.1}$ \\
\midrule
\multirow{3}{*}{Graph SP}
  & $c^*$ & $212{\scriptstyle\pm54}$ & $377{\scriptstyle\pm90}$ & $190{\scriptstyle\pm65}$ & $226{\scriptstyle\pm63}$ & $229{\scriptstyle\pm54}$ & $259{\scriptstyle\pm20}$ \\
  & $f_{\mathrm{EN}}$ & $64.7{\scriptstyle\pm17.4}$ & $18.0{\scriptstyle\pm27.5}$ & $75.2{\scriptstyle\pm19.9}$ & $58.1{\scriptstyle\pm18.9}$ & $57.0{\scriptstyle\pm16.0}$ & $45.1{\scriptstyle\pm5.4}$ \\
  & $f_{\ell}$ & $71.5{\scriptstyle\pm18.4}$ & $26.0{\scriptstyle\pm21.1}$ & $81.2{\scriptstyle\pm23.1}$ & $64.9{\scriptstyle\pm19.2}$ & $66.5{\scriptstyle\pm17.7}$ & $70.6{\scriptstyle\pm8.8}$ \\
\midrule
\multirow{3}{*}{T.\ Rank}
  & $c^*$ & $183{\scriptstyle\pm34}$ & $172{\scriptstyle\pm41}$ & $192{\scriptstyle\pm12}$ & $211{\scriptstyle\pm21}$ & $227{\scriptstyle\pm30}$ & $204{\scriptstyle\pm33}$ \\
  & $f_{\mathrm{EN}}$ & $73.8{\scriptstyle\pm23.4}$ & $82.2{\scriptstyle\pm27.4}$ & $66.3{\scriptstyle\pm9.8}$ & $50.4{\scriptstyle\pm14.9}$ & $39.0{\scriptstyle\pm20.7}$ & $56.2{\scriptstyle\pm22.3}$ \\
  & $f_{\ell}$ & $71.1{\scriptstyle\pm21.6}$ & $78.1{\scriptstyle\pm23.2}$ & $51.6{\scriptstyle\pm8.9}$ & $41.6{\scriptstyle\pm14.0}$ & $34.6{\scriptstyle\pm21.1}$ & $54.3{\scriptstyle\pm22.6}$ \\
\bottomrule
\end{tabular}
\caption{\textbf{GLM-5-1.} Here we present the $\langle c^*\rangle$  at which SMD values were reported along with $\langle f_{EN}\rangle$ and $\langle f_\ell\rangle$ at the respective complexities over 300 bootstraps.}
\label{tab:cpeak_glm51}
\end{table*}


\subsection{Back-Translation Experiments \label{app:backtranslation}}
\label{sec:bt_expts}

As described in section \ref{sec:mit_expts}, we performed some back-translation experiments to rule out translation errors. For Gemini 3 Flash and Gemini 3.1 Flash Lite, in cases where a significant SMD was observed, we picked the worst-performing language and back-translated its prompt template to English. 

In all cases,  the back-translated prompts were semantically identical to the original prompt and, in addition had $\geq95\%$ word-level overlap with the original prompt in English. 

Nevertheless, to rule out the possibility that small variations in the prompt lead to variations in accuracy, we evaluated the accuracy of the model using this back-translated prompt on the two original complexity values $c_1, c_2$ that were closest to the value of $c^*$ defined in equation \eqref{cstardef} with $c_1 < c^* < c_2$. This  yielded accuracy values $f_{\text{En}}^{b}(c_1)$ and $f_{\text{En}}^{b}(c_2)$.  This allows us to compute
\begin{equation}
\delta_i = f^b_{\text{En}}(c_i) - f_{\text{En}}(c_i),
\end{equation}
where $f_{\text{En}}(c_i)$ are the original accuracies and $c_i$ runs over $c_1$ and $c_2$. With the original standard deviations, $\sigma_1$ and $\sigma_2$, we can then finally compute
\begin{equation}
\delta = {1 \over (c_2 - c_1)} \left( |c_2 - c^*| {\delta_1^2 \over \sigma_1^2} + |c_1 - c^*| {\delta_2^2 \over \sigma_2^2} \right)
\end{equation}

This measure $\delta$ yields a weighted $\chi^2$ value, and values of $\delta \gg 1$ would indicate that back-translation has induced a significant difference in accuracy. 

We present the observed value of $\delta$ in Table~\ref{tab:bt_expts}.  These show that there are no significant gaps between the original prompt performance and back-translated prompt performance. 

We note that for Gemini-3.1-Flash-Lite, the tournament ranking prompt was an exact match to the original prompt. The Pr DSA task with Gemini 3 Flash and the Graph SP tasks  with Gemini 3.1 Flash Lite did not display any significant SMD values at all.  For these reasons, the corresponding entries in Table \ref{tab:bt_expts} have been left blank. 

\begin{table}[htbp]
\centering
\footnotesize

\begin{tabular}{l@{\hspace{10pt}}cc}
\toprule
\textbf{Task} & \textbf{Gemini 3 Flash} & \textbf{Gemini 3.1 Flash Lite} \\
\midrule
SLT  & $0.064$ (TE) & $0.368$ (AR) \\
\midrule
DSA  & $0.683$ (HI) & $1.195$ (TE) \\
\midrule
Pr DSA & --- & $0.761$ (HI) \\
\midrule
Graph SP & $2.903$ (TE)& --- \\
\midrule 
T. Rank & $2.003$ (TE) & --- (AR) \\
\bottomrule
\end{tabular}
\vspace{1.5em}

\caption{The variance $\delta$ between the back-translated prompt and the original prompt.}
\label{tab:bt_expts}
\end{table}
\paragraph{}

\subsection{Correlation between Thinking Budget and Cross-Lingual gaps} \label{sec:thought_corr}
To verify if the thought length is correlated with the cross-lingual gaps, we did the following analysis. We ranked the language based on SMD and then computed the Spearman's rank correlation with the rankings observed on the average thought usage. This average was computed over all the datapoints for that specific model x language x task triplet (approximately 1000 points).  The number of thought tokens was accessible for Gemini 3 Flash, whereas for other models we estimated the average thought token length using the thought character lengths.\footnote{This is an imperfect measure since character-based measurement does not account for script density differences. For instance, a single CJK character encodes more information than an ASCII character. This introduces an estimated 10--20\% measurement error in cross-script comparisons.} We present the results of this analysis in Table~\ref{tab:thinking_rank_correlation}. These results indicate that thought length is not strongly correlated with language performance.

\begin{table*}[htbp]
\centering
\normalsize
\begin{tabular}{l@{\hspace{10pt}}ccccc}
\toprule
\textbf{Model} & \textbf{SLT} & \textbf{DSA} & \textbf{Pr DSA} & \textbf{Graph SP} & \textbf{T.\ Rank} \\
\midrule
Gemini 3 Flash & $-0.371$ & $-0.600$ & $-0.086$ & $-0.429$ & $-0.657$ \\
\midrule
Gemma 4 31B IT & $+0.429$ & $-0.314$ & $-0.257$ & $+0.257$ & $+0.029$ \\
\midrule
Gemini 3.1 Flash Lite & $-0.257$ & $+0.200$ & $-0.314$ & $-0.143$ & $+0.143$ \\
\midrule
GLM-5-1 & \cellcolor{green!20}$\mathbf{-0.886^*}$ & $-0.657$ & \cellcolor{green!20}$\mathbf{-0.886^*}$ & $+0.029$ & $-0.257$ \\
\bottomrule
\end{tabular}
\caption{Spearman rank correlation ($\rho$) between language ranking by thinking effort and language ranking by Signed Max Divergence (SMD) across tasks and models.
Thinking effort is measured by \texttt{avg\_thinking\_tokens} for Gemini 3 Flash (script-agnostic) and by \texttt{avg\_thought\_len} (character count) for all other models.
$^*$Significant at $p < 0.05$ (highlighted in bold and green).
A negative $\rho$ would support the hypothesis that languages with lower SMD (better cross-lingual parity) achieve this by eliciting greater thinking effort from the model.
Only 2/20 combinations show a significant negative correlation (both GLM-5-1), while 18/20 are statistically indistinguishable from zero, confirming that \textbf{cross-lingual gaps cannot be explained merely by differential thinking effort}.}
\label{tab:thinking_rank_correlation}
\end{table*}

\section{Gamma Fits}
In this section we present graphs for the incomplete gamma function fit described in section \ref{sec:expts} for all the model, task pairs, and for all the languages. These are generated using a Bayesian bootstrap approach: for each bootstrap iteration, we sample the accuracy at each complexity level from the posterior Beta distribution Beta(k+1, n−k+1), where k is the number of correct responses and n is the total, and refit the gamma CDF to obtain a new curve. We repeat this 300 times and shade the region between the 2.5th and 97.5th percentiles of the predicted curves, yielding a 95\% credible interval for the fitted curve at each complexity.

\label{sec:gamma_profiles}
\begin{figure*}[htbp]
    \centering
    \begin{subfigure}[b]{0.48\textwidth}
        \includegraphics[width=\textwidth]{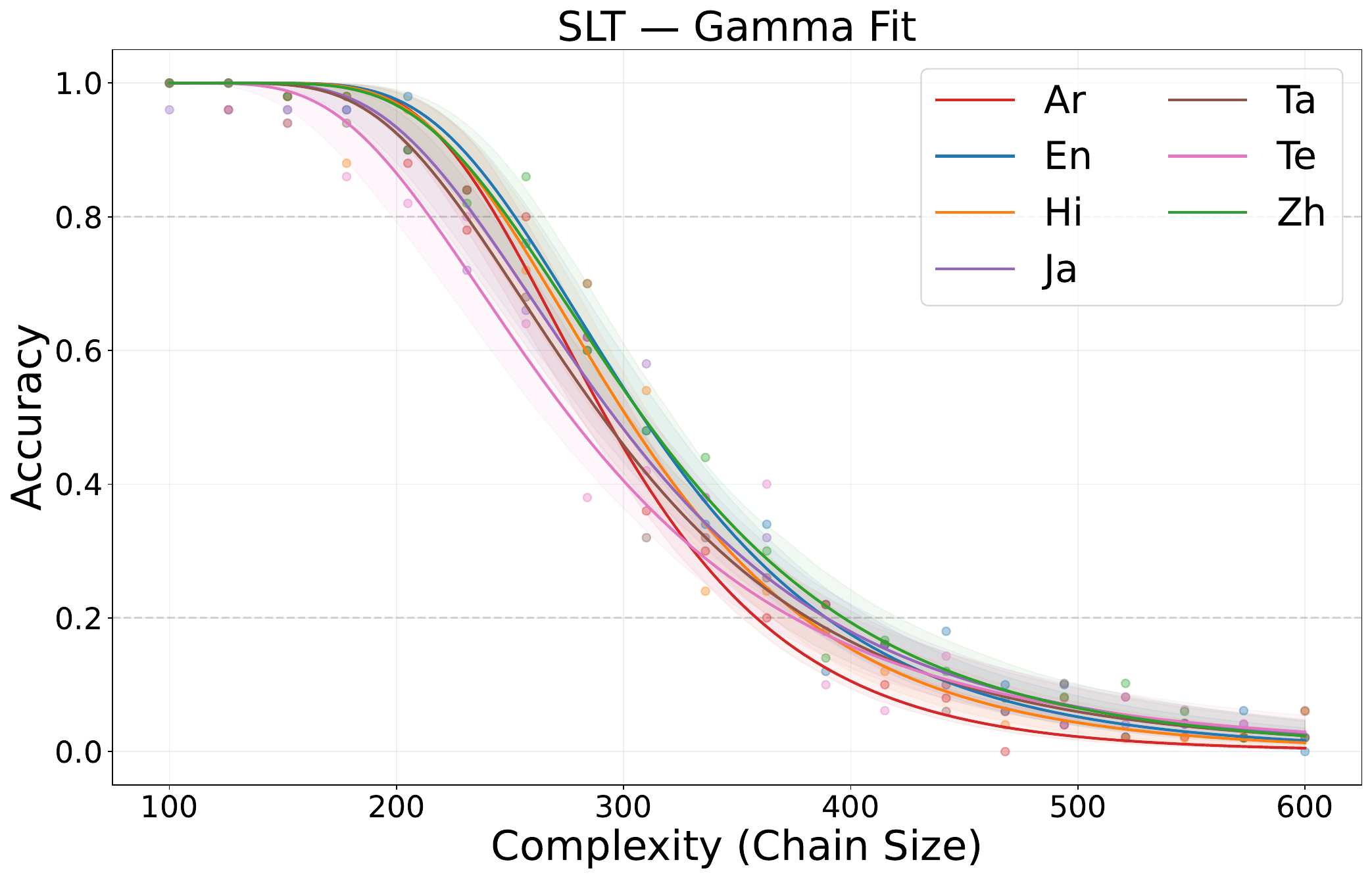}
    \end{subfigure}
    \hfill
    \begin{subfigure}[b]{0.48\textwidth}
        \includegraphics[width=\textwidth]{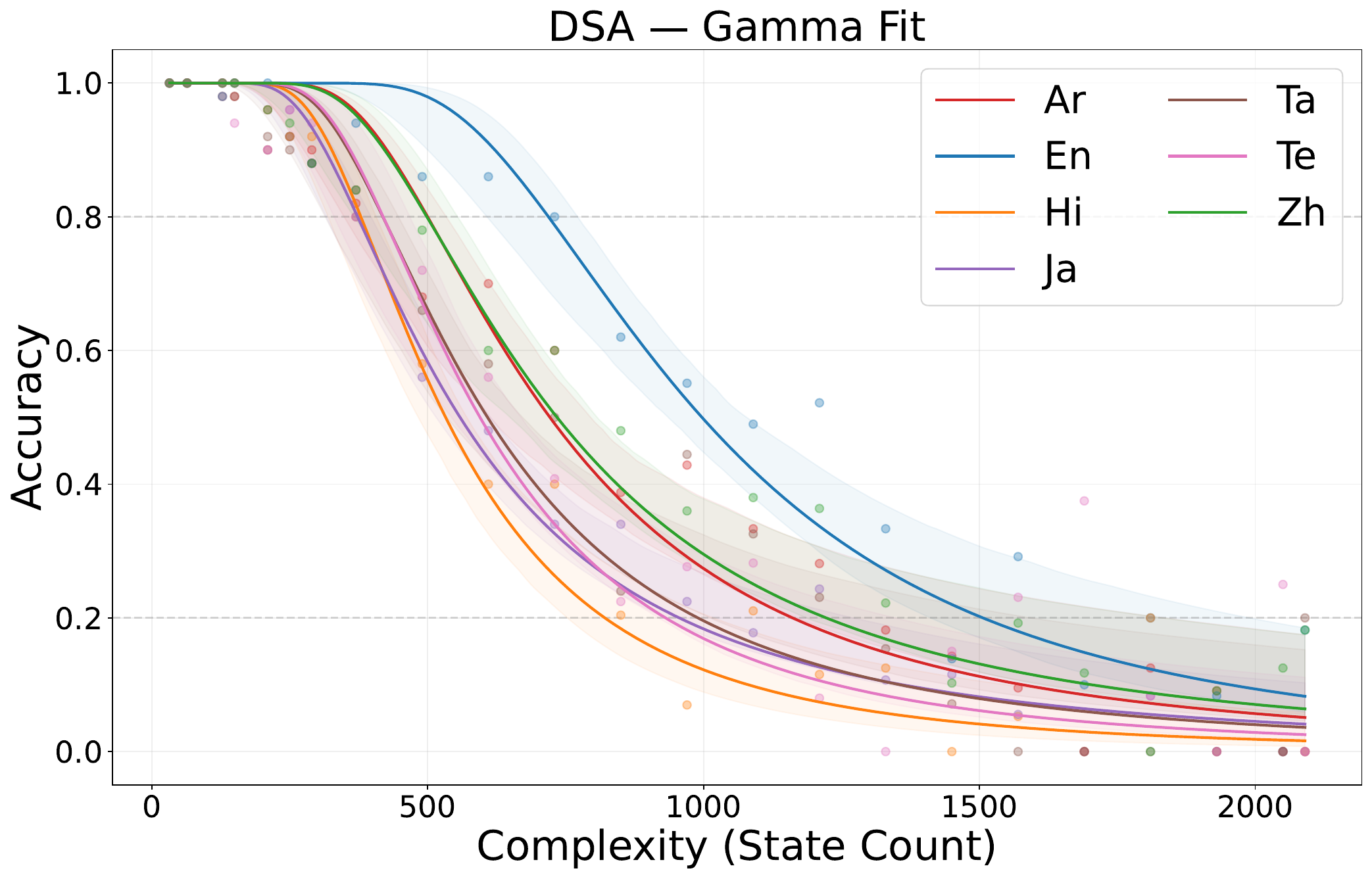}
    \end{subfigure}
    \begin{subfigure}[b]{0.48\textwidth}
        \includegraphics[width=\textwidth]{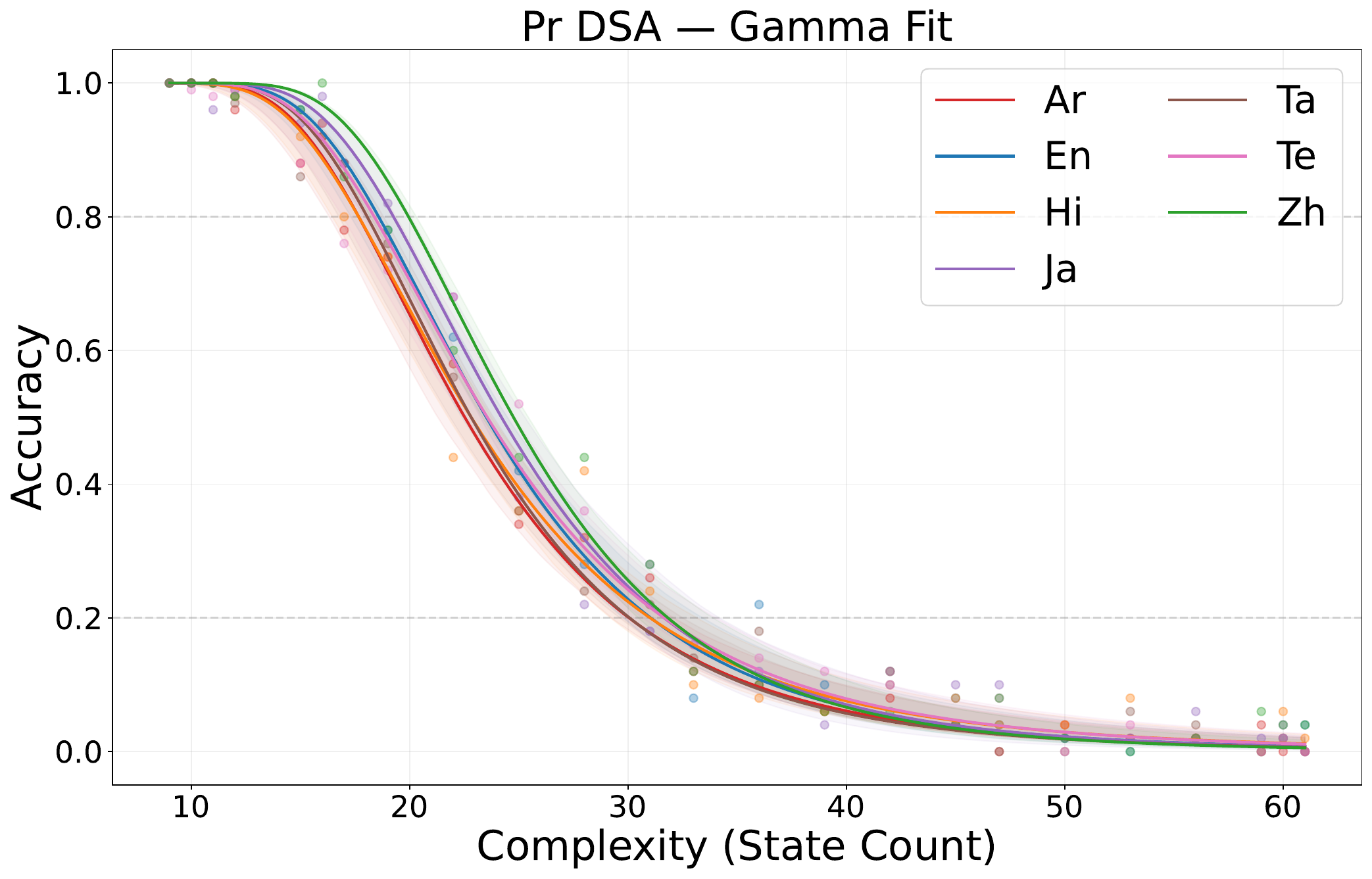}
    \end{subfigure}
    \hfill
    \begin{subfigure}[b]{0.48\textwidth}
        \includegraphics[width=\textwidth]{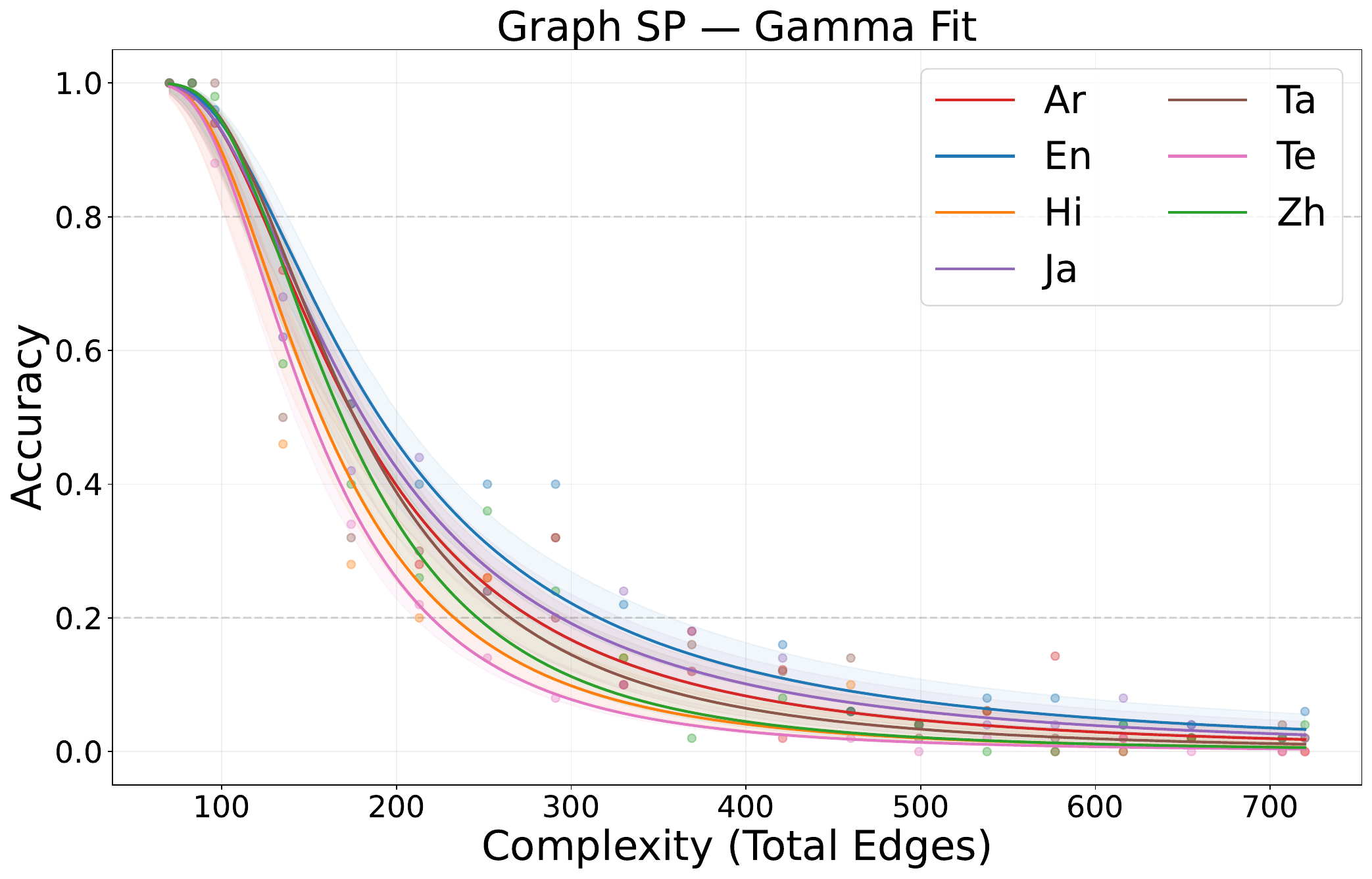}
    \end{subfigure}
    \hfill
    \begin{subfigure}[b]{0.48\textwidth}
        \includegraphics[width=\textwidth]{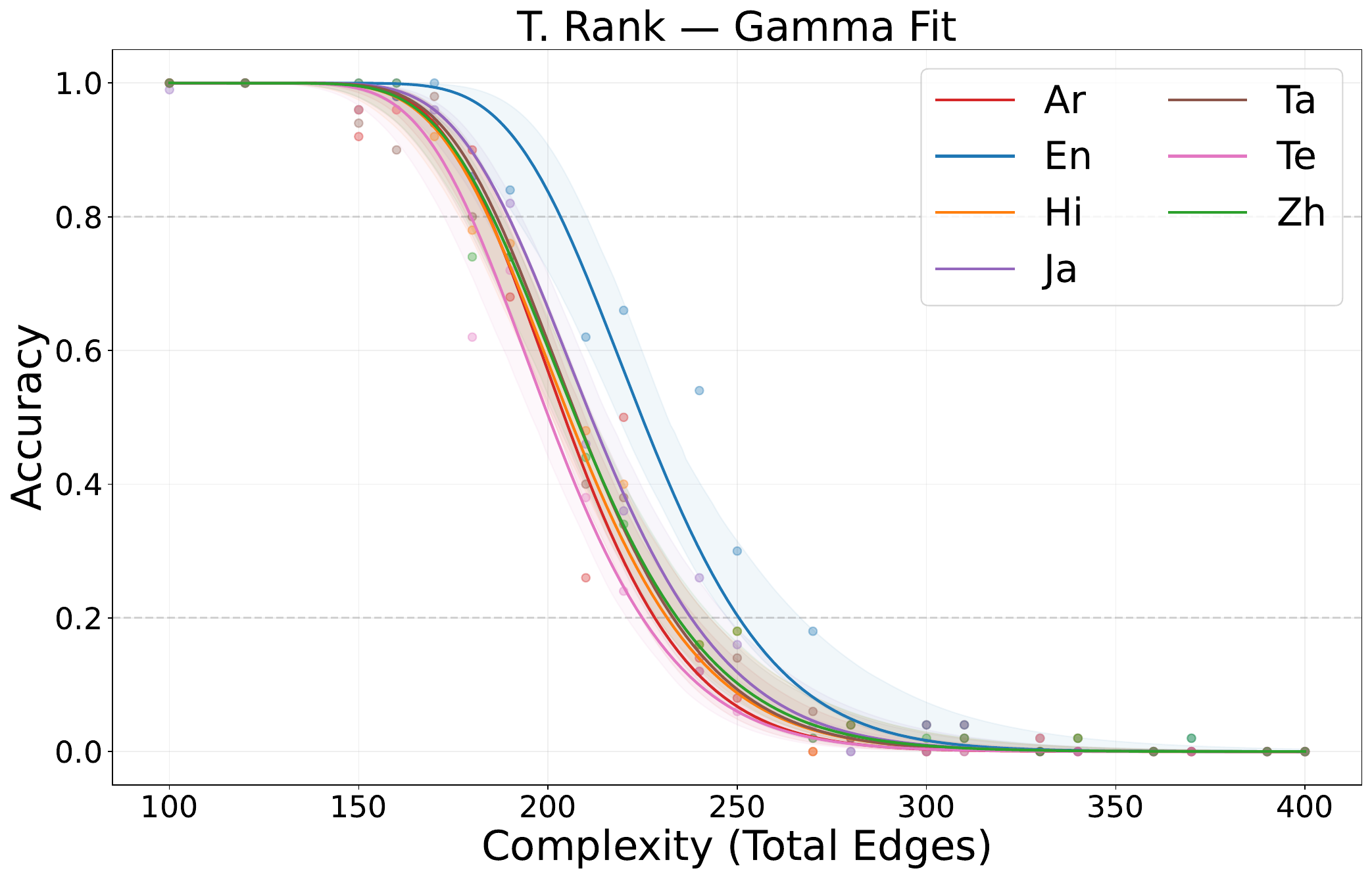}
    \end{subfigure}
    \caption{Accuracy vs Complexity plots for different tasks. In each sub-plot, we compare between English and non-English with Gemini 3 Flash model.}
    \label{fig:flash_acc_vs_complexity}
\end{figure*}

\begin{figure*}[htbp]
    \centering
    \begin{subfigure}[b]{0.48\textwidth}
        \includegraphics[width=\textwidth]{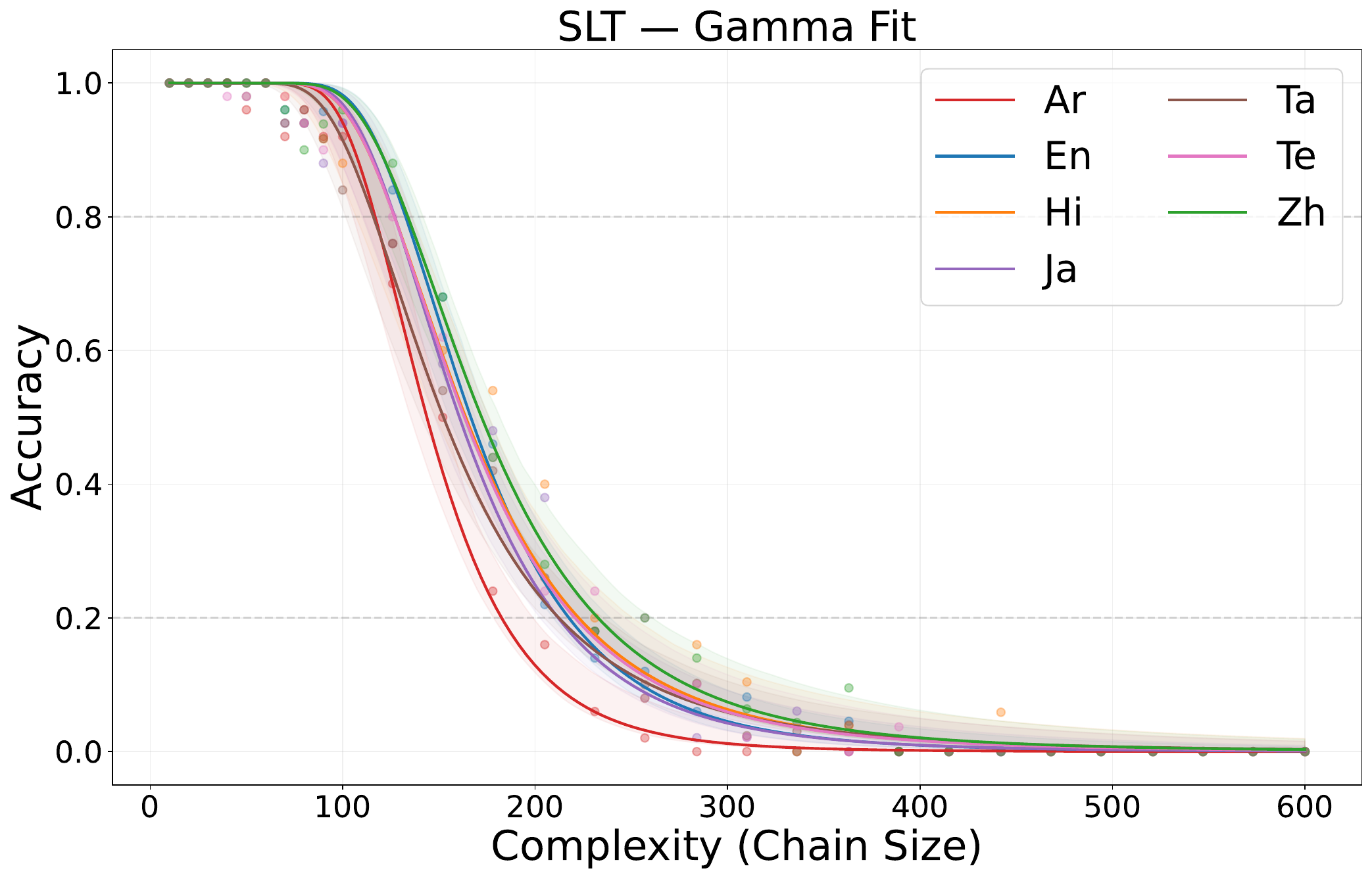}
    \end{subfigure}
    \hfill
    \begin{subfigure}[b]{0.48\textwidth}
        \includegraphics[width=\textwidth]{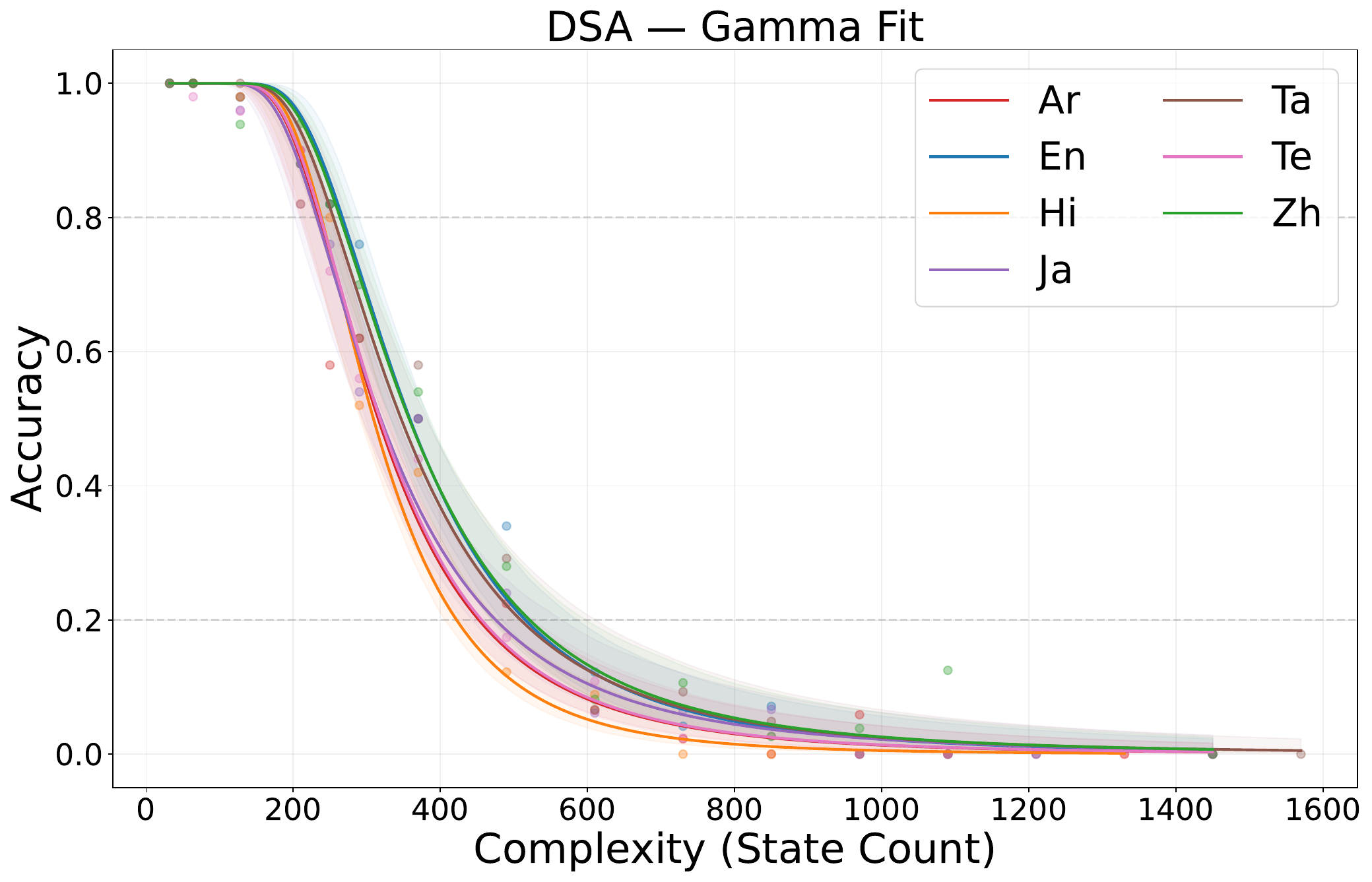}
    \end{subfigure}
    \begin{subfigure}[b]{0.48\textwidth}
        \includegraphics[width=\textwidth]{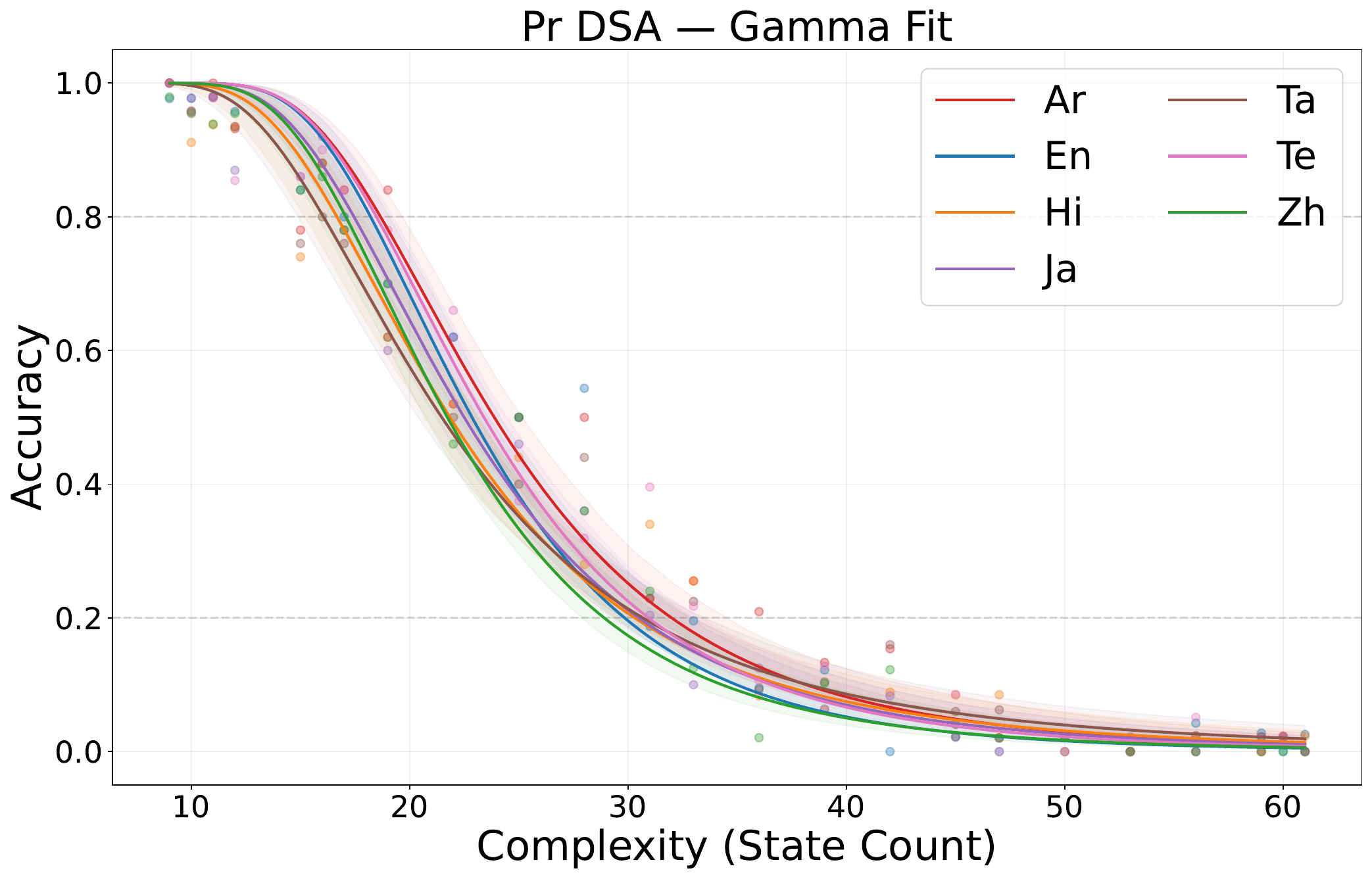}
    \end{subfigure}
    \hfill
    \begin{subfigure}[b]{0.48\textwidth}
        \includegraphics[width=\textwidth]{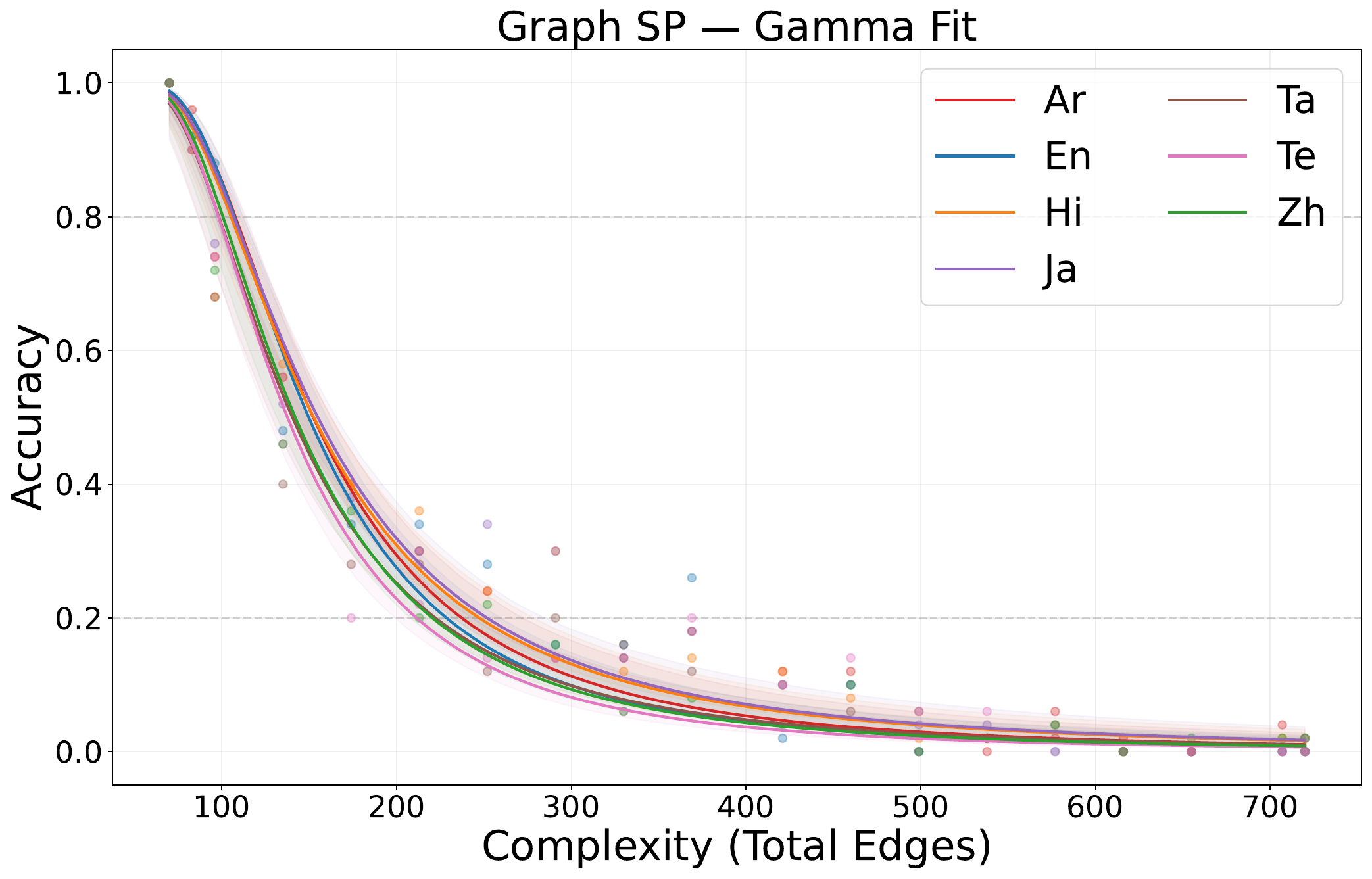}
    \end{subfigure}
    \hfill
    \begin{subfigure}[b]{0.48\textwidth}
        \includegraphics[width=\textwidth]{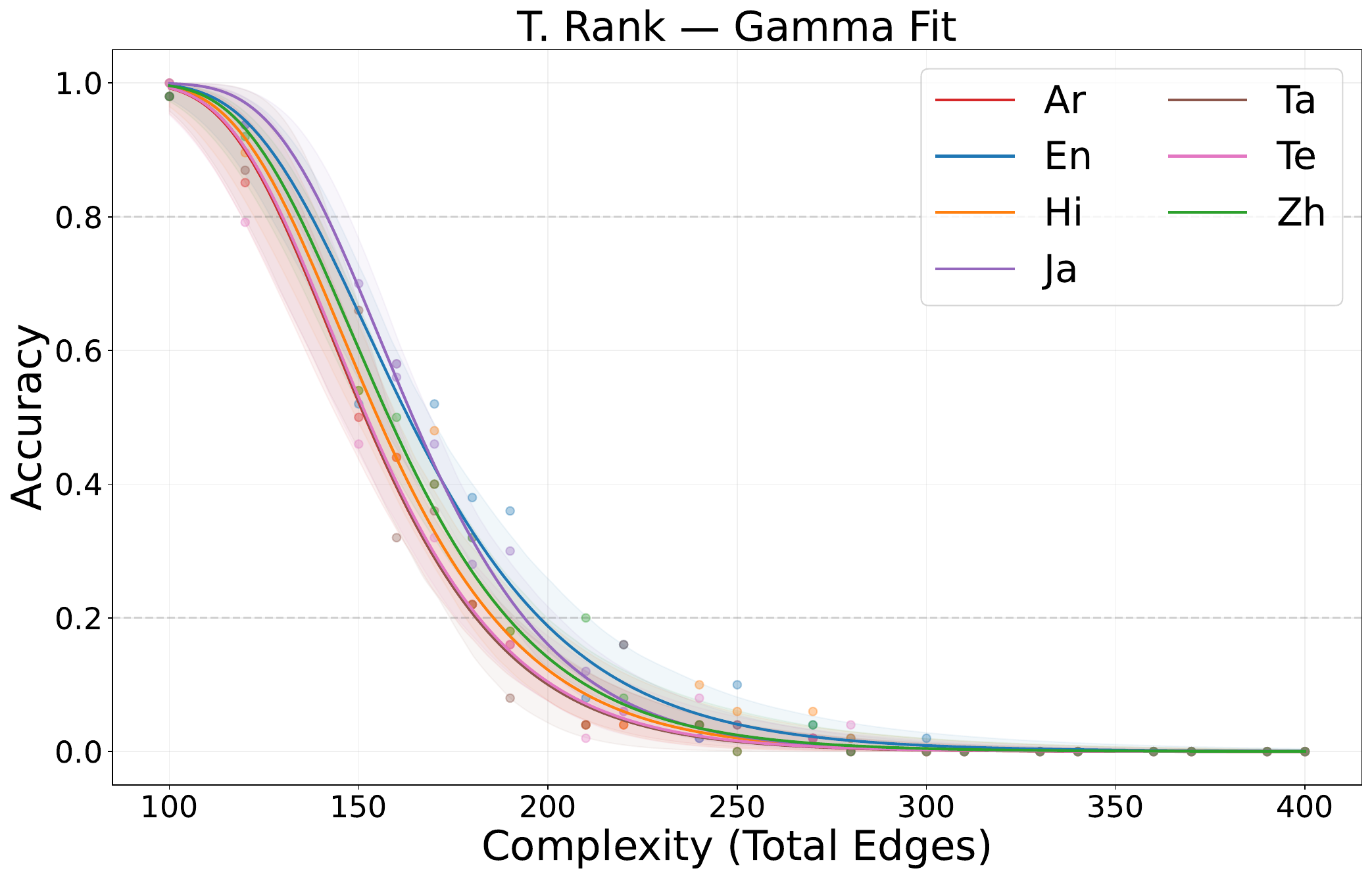}
    \end{subfigure}
    \caption{Accuracy vs Complexity plots for different tasks. In each sub-plot, we compare between English and non-English with Gemma-4-31b-IT model.}
    \label{fig:gemma_acc_vs_complexity}
\end{figure*}

\begin{figure*}[htbp]
    \centering
    \begin{subfigure}[b]{0.48\textwidth}
        \includegraphics[width=\textwidth]{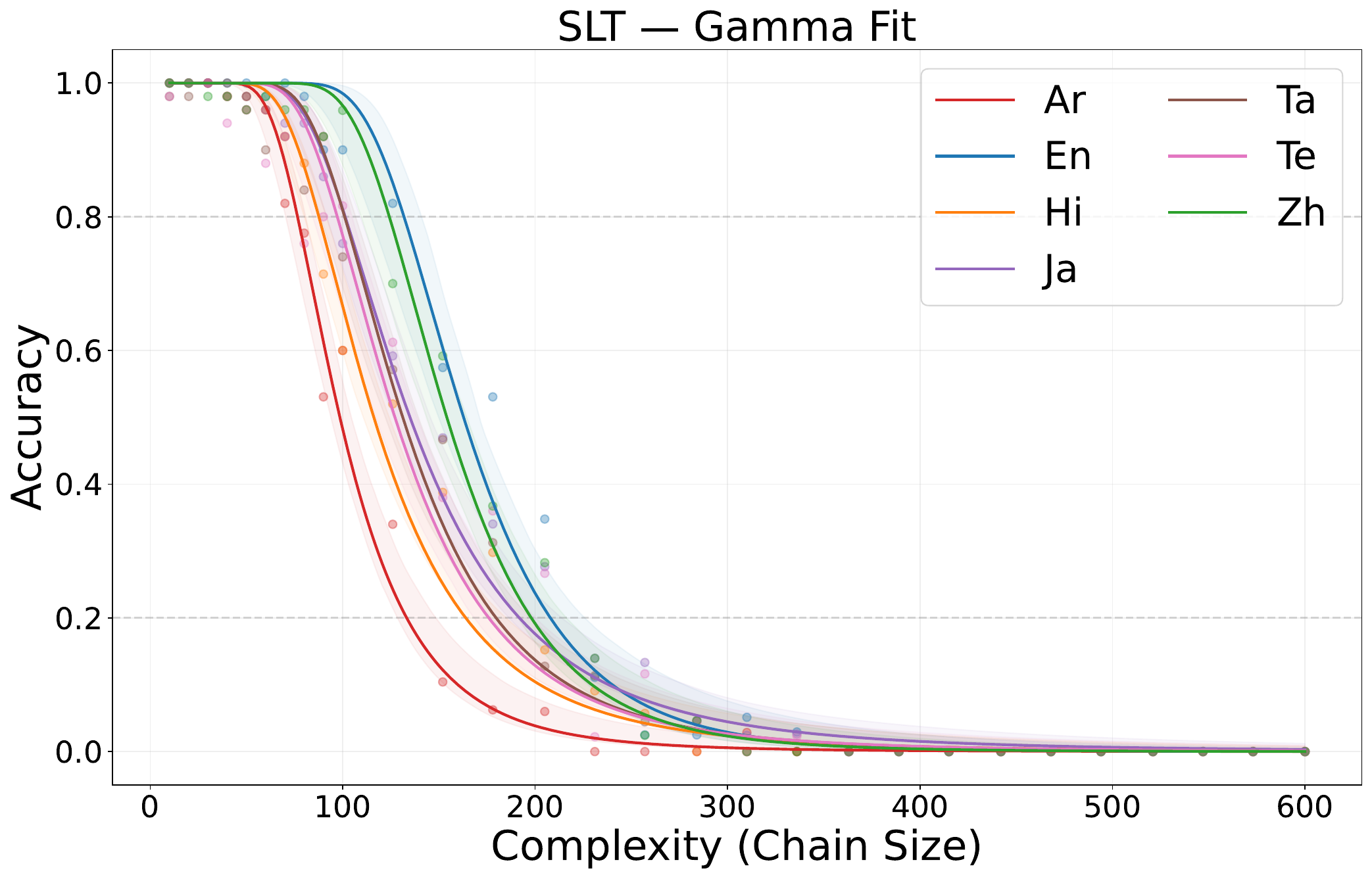}
    \end{subfigure}
    \hfill
    \begin{subfigure}[b]{0.48\textwidth}
        \includegraphics[width=\textwidth]{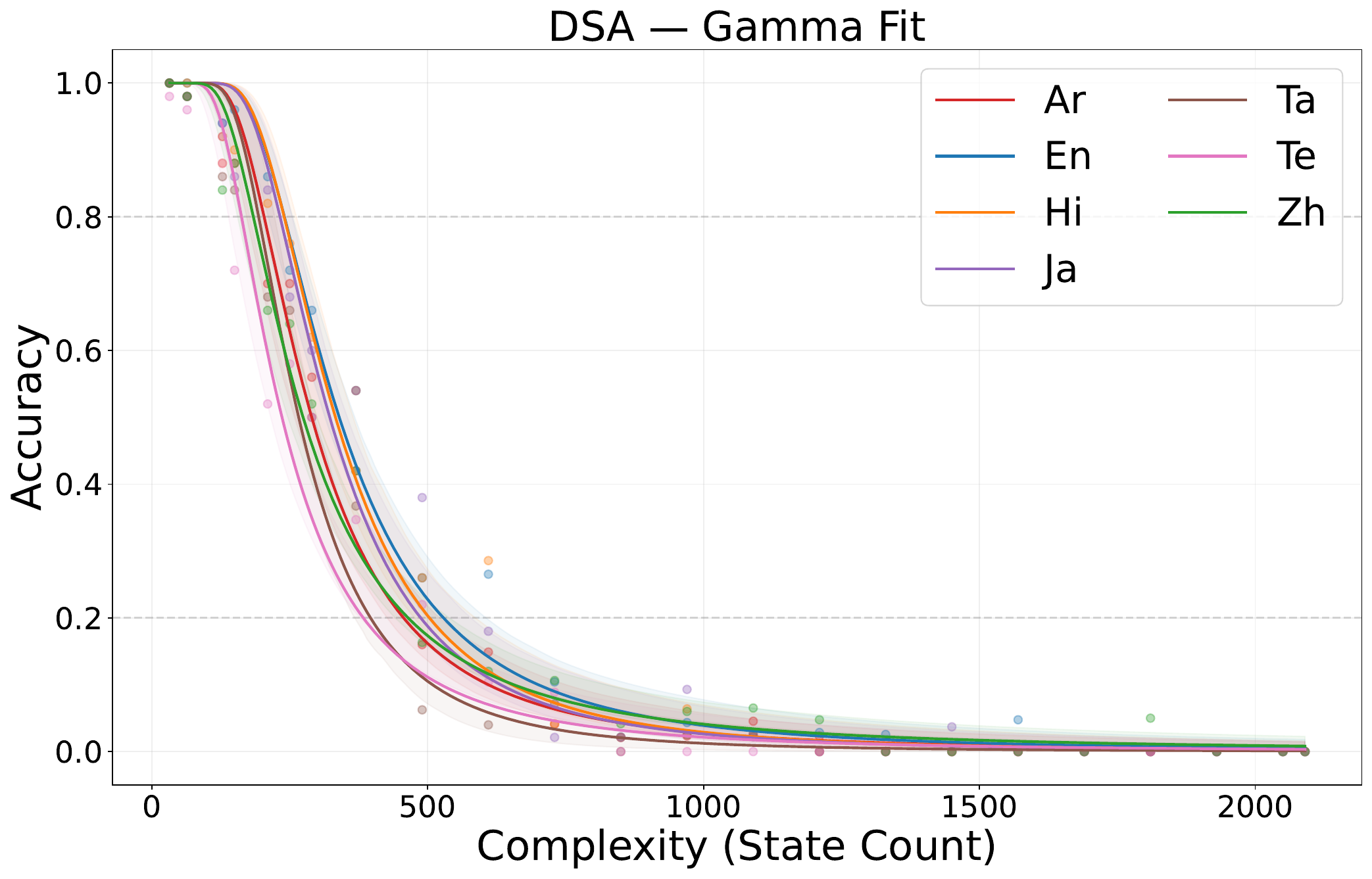}
    \end{subfigure}
    \begin{subfigure}[b]{0.48\textwidth}
        \includegraphics[width=\textwidth]{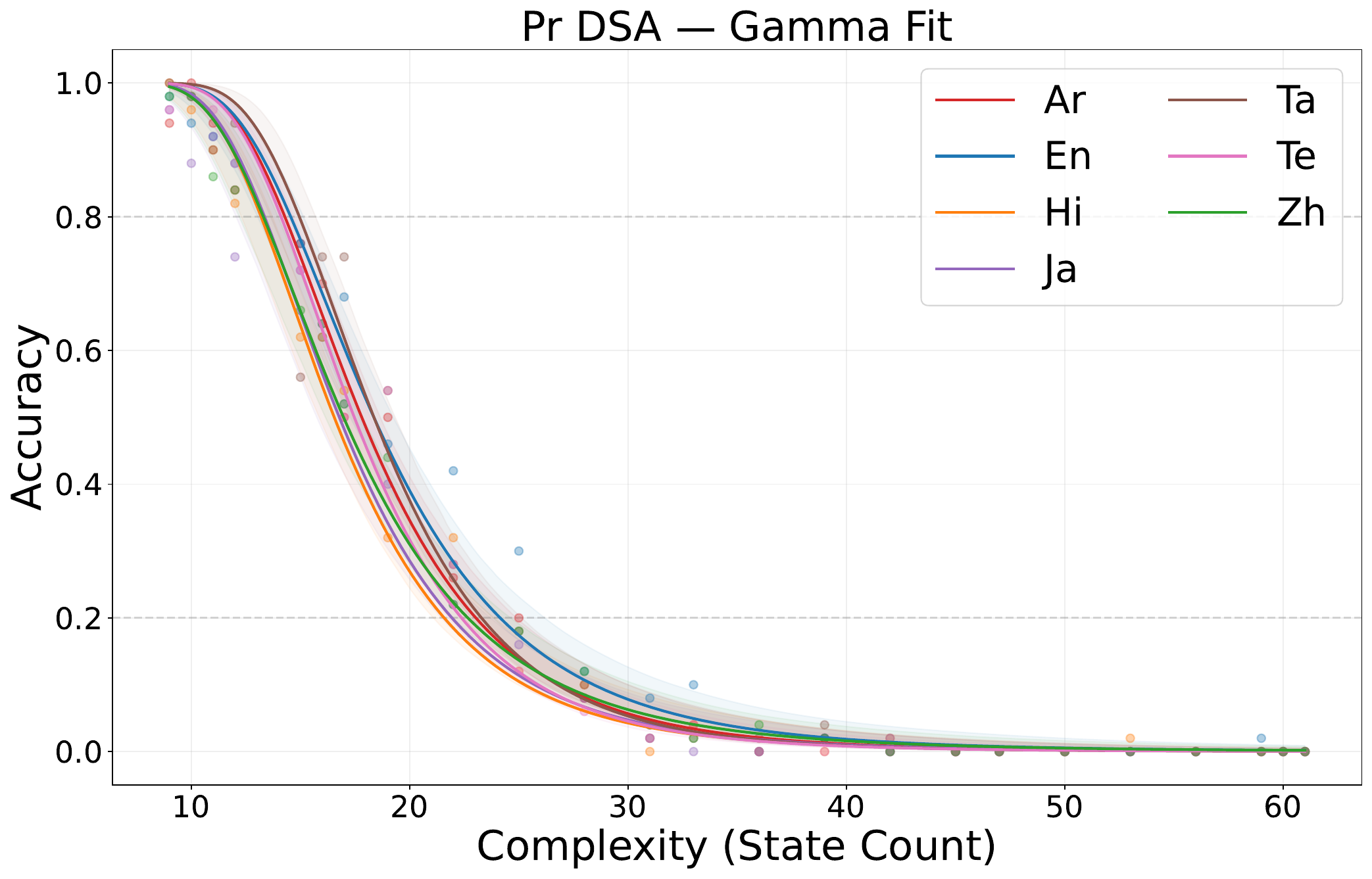}
    \end{subfigure}
    \hfill
    \begin{subfigure}[b]{0.48\textwidth}
        \includegraphics[width=\textwidth]{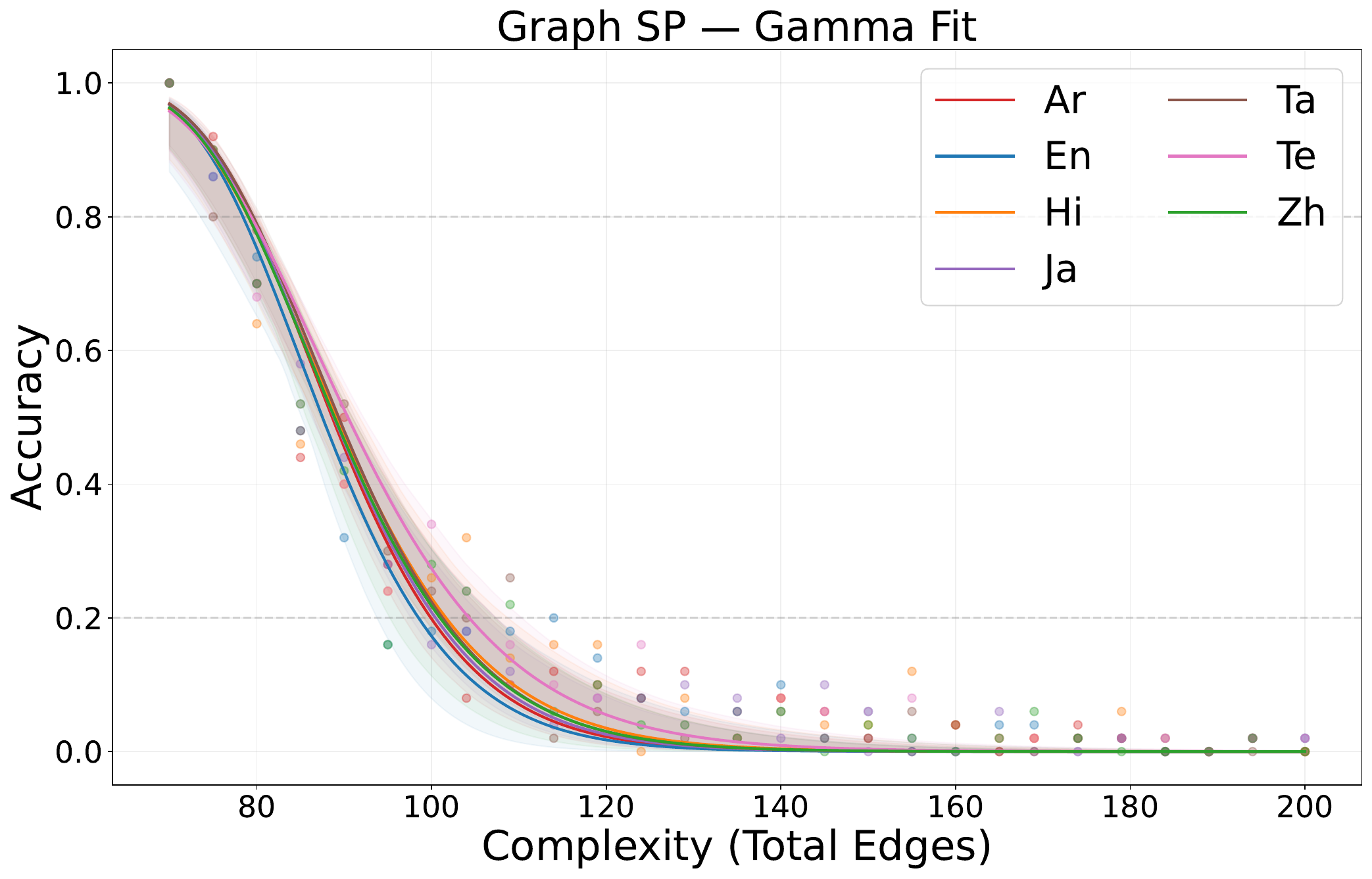}
    \end{subfigure}
    \hfill
    \begin{subfigure}[b]{0.48\textwidth}
        \includegraphics[width=\textwidth]{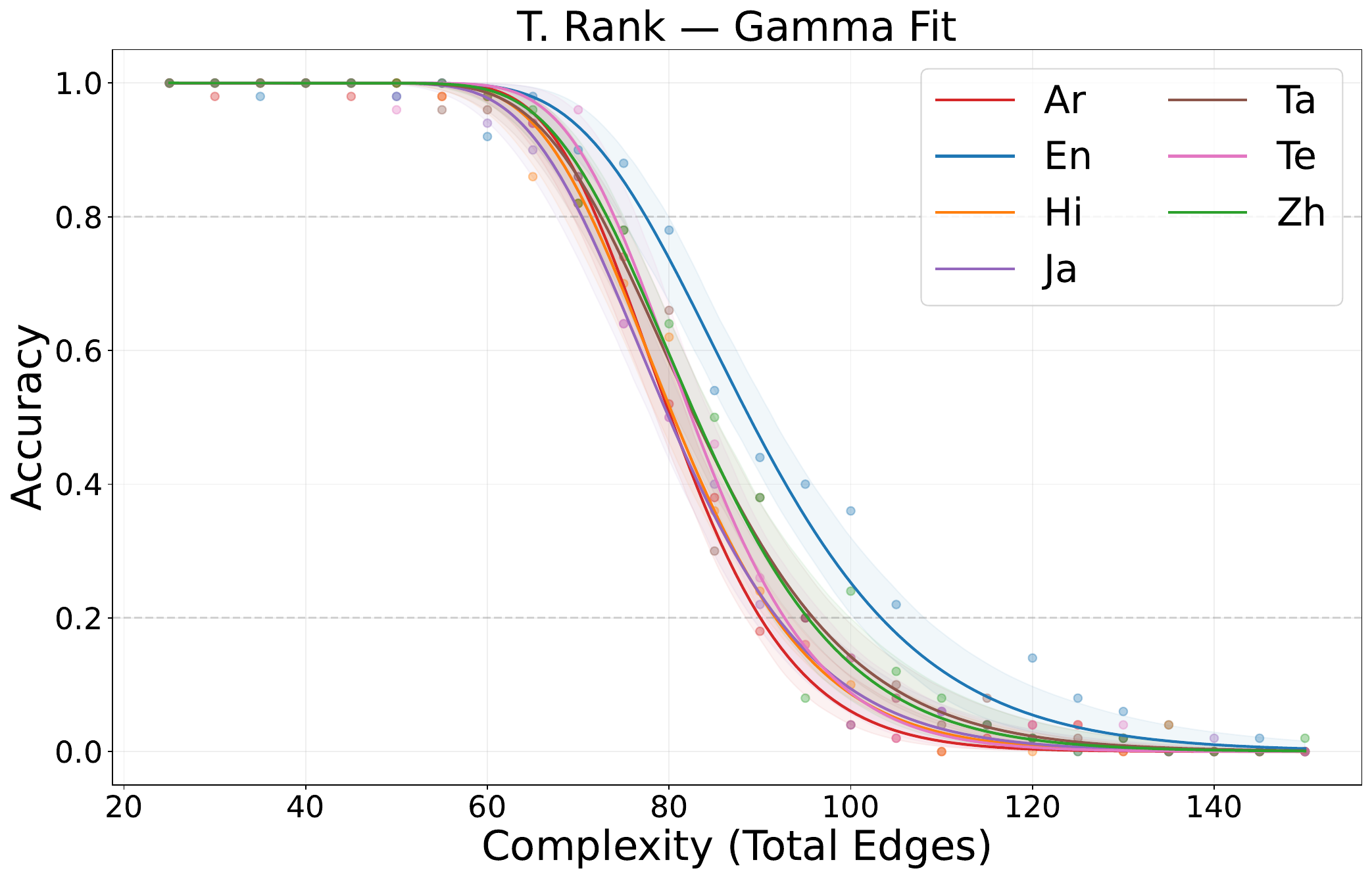}
    \end{subfigure}
    \caption{Accuracy vs Complexity plots for different tasks. In each sub-plot, we compare between English and non-English with Gemini-3.1-Flash-Lite model.}
    \label{fig:lite_acc_vs_complexity}
\end{figure*}

\begin{figure*}[htbp]
    \centering
    \begin{subfigure}[b]{0.48\textwidth}
        \includegraphics[width=\textwidth]{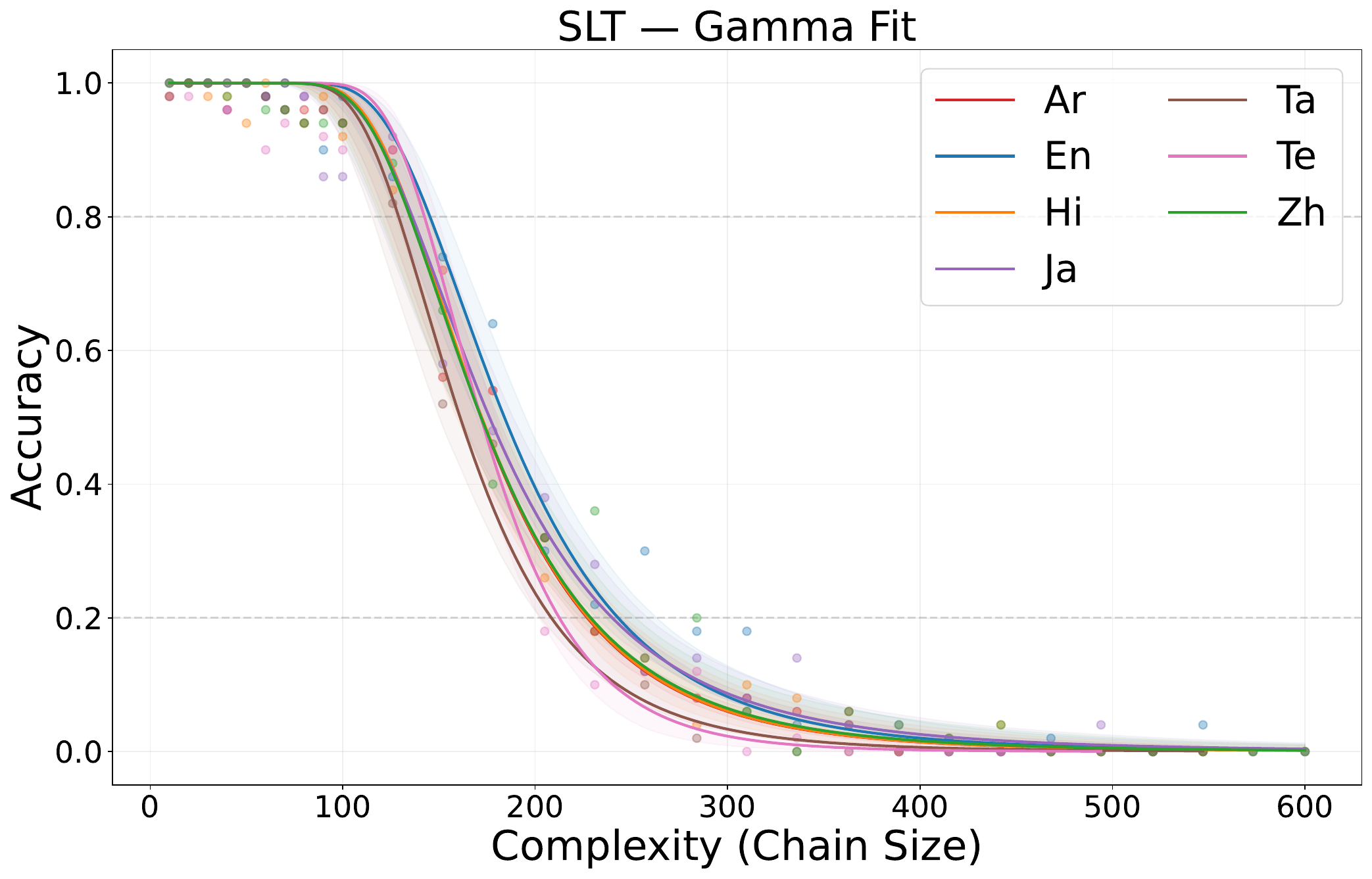}
    \end{subfigure}
    \hfill
    \begin{subfigure}[b]{0.48\textwidth}
        \includegraphics[width=\textwidth]{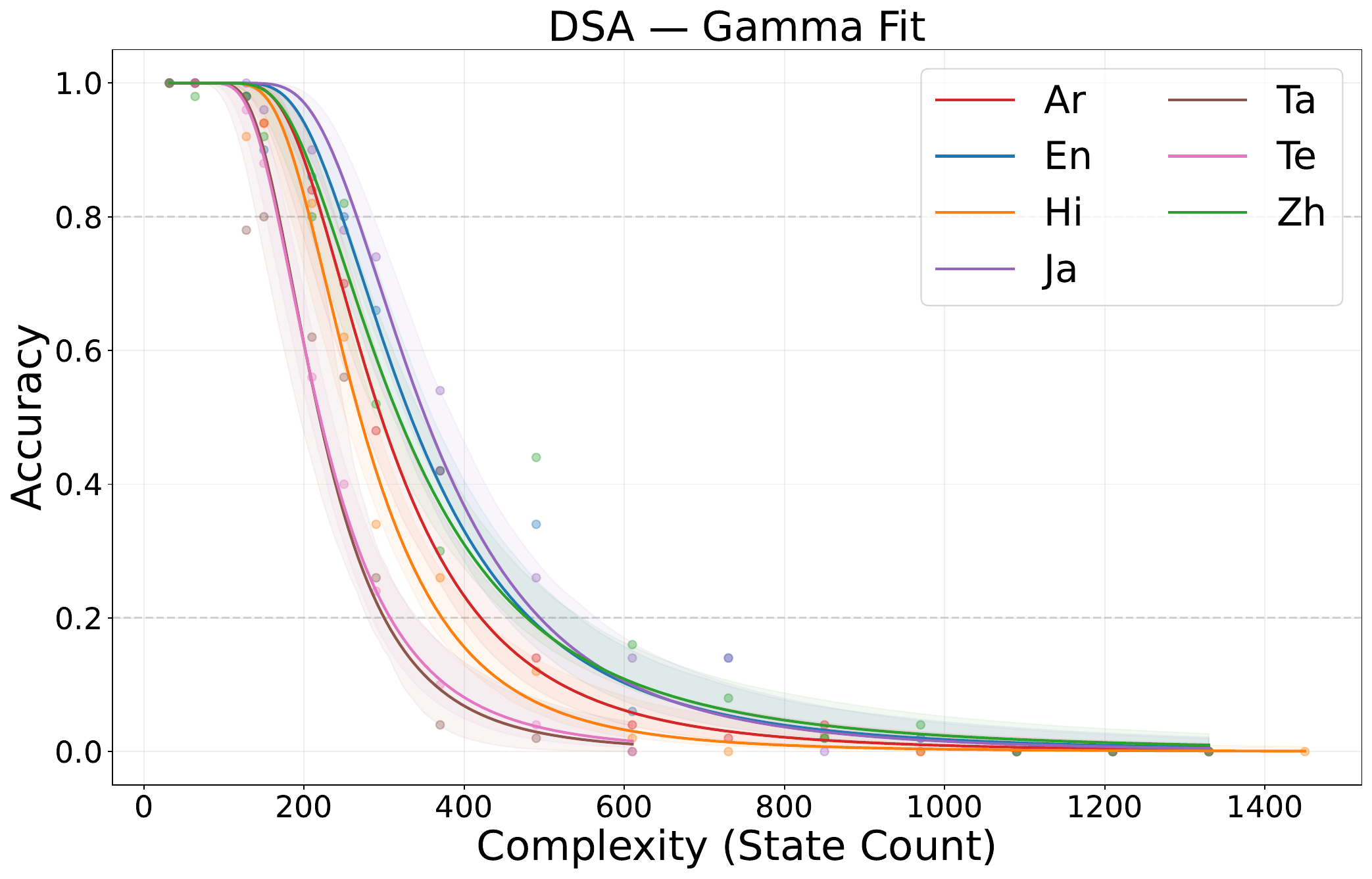}
    \end{subfigure}
    \begin{subfigure}[b]{0.48\textwidth}
        \includegraphics[width=\textwidth]{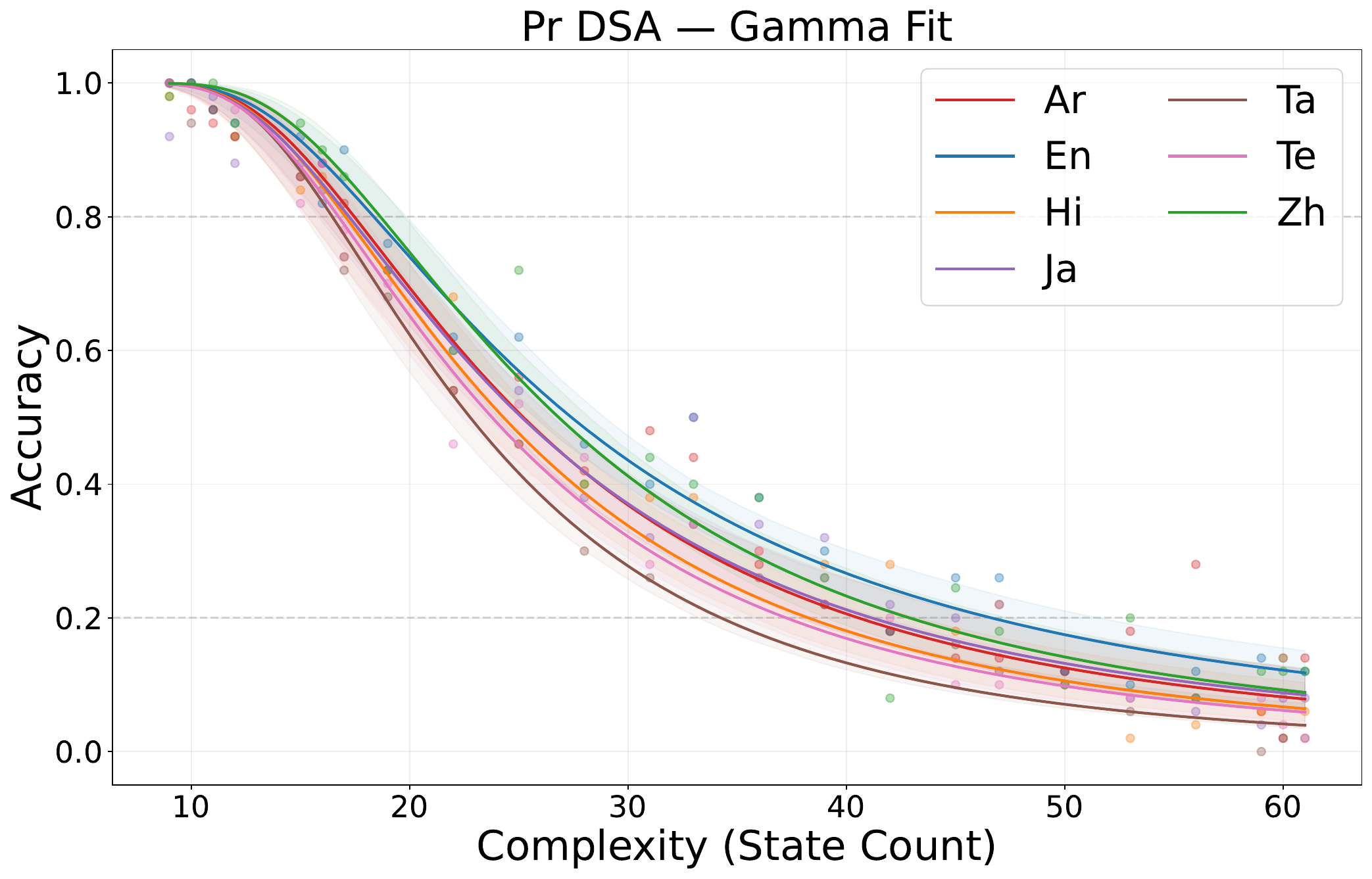}
    \end{subfigure}
    \hfill
    \begin{subfigure}[b]{0.48\textwidth}
        \includegraphics[width=\textwidth]{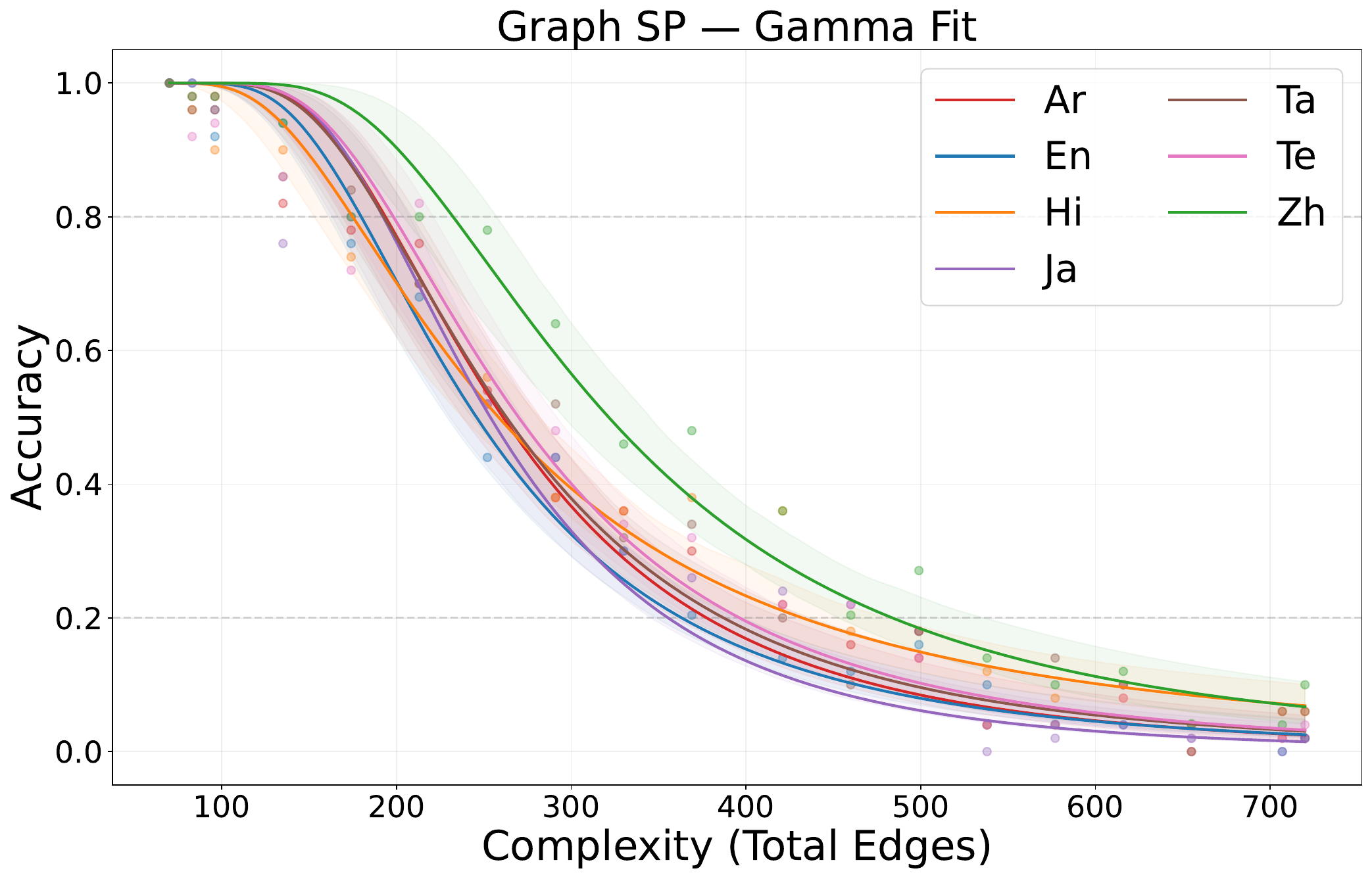}
    \end{subfigure}
    
    \begin{subfigure}[b]{0.48\textwidth}
        \includegraphics[width=\textwidth]{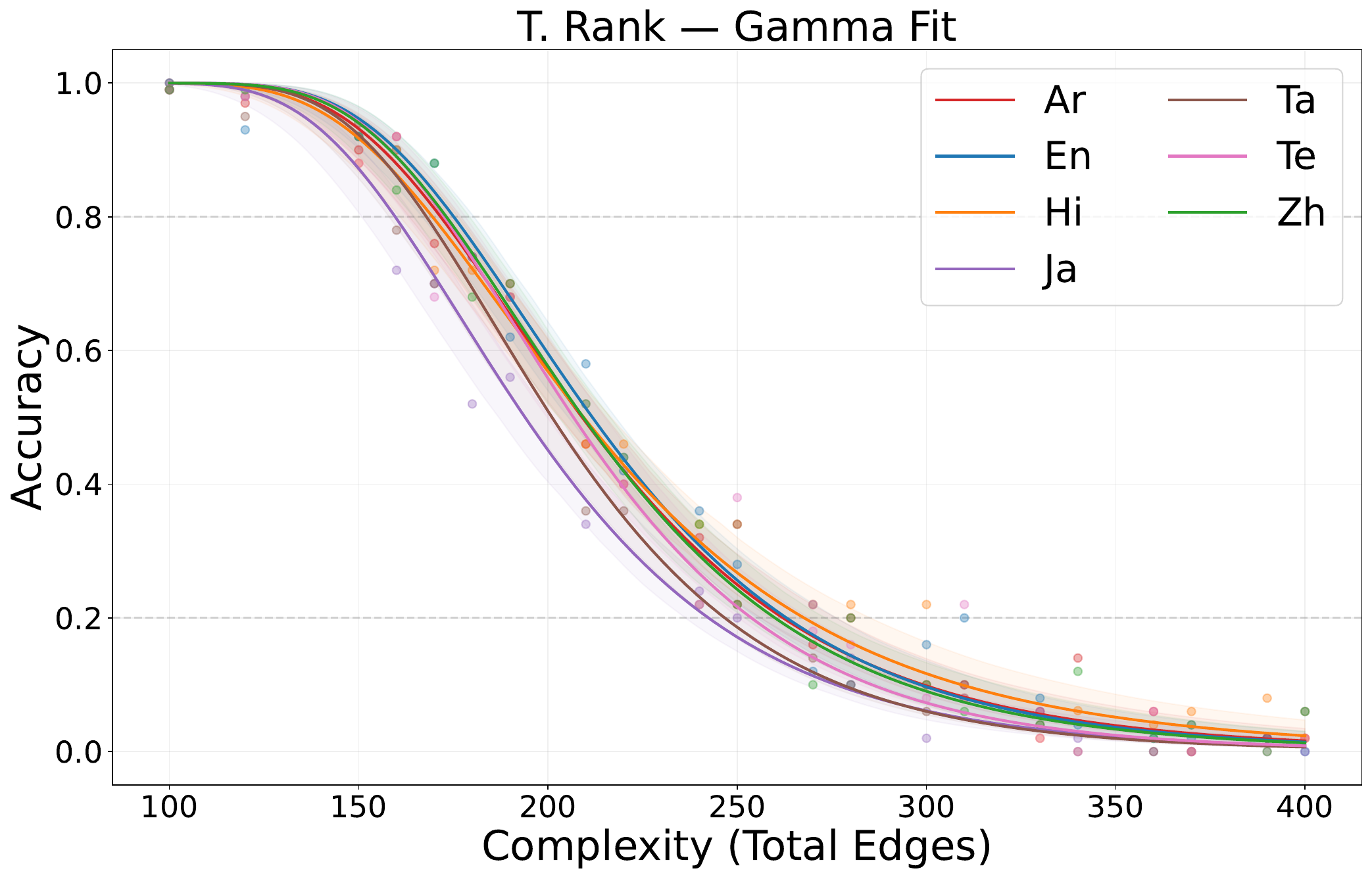}
    \end{subfigure}
    \caption{Accuracy vs Complexity plots for different tasks. In each sub-plot, we compare between English and non-English with GLM-5.1 model.}
    \label{fig:glm_acc_vs_complexity}
\end{figure*}

\section{Construction of Prompts and Puzzles}\label{appendix:dataset_examples}
In this section, we explain the construction of prompts and puzzles and provide sample prompts for each puzzle. We start by explaining the construction of the Graph Shortest Path Puzzle. The other puzzles in our suite can be straightforwardly constructed given their description and the prompts given in section \ref{subsecprompts}. We explain the procedure used to translate the puzzles in section \ref{sec:annot_pipe}.

\subsection{Graph Shortest Path Construction}
\label{sec:graph_sp_generation}
The Graph Shortest Path puzzle presents a social network as an undirected graph and asks the model to find the shortest path between a source and a target node. The puzzle is constructed so that a \emph{unique} shortest path of length $d$ exists, embedded within a densely cross-wired graph of bypass chains. The generation procedure is parameterized by:
\begin{itemize}[noitemsep,topsep=2pt]
    \item $d$: the shortest path length (backbone length),
    \item $n_c$: the number of parallel bypass chains,
    \item $c_e$: the chain excess (each bypass chain has length $d + c_e$, making it strictly longer than the backbone),
    \item $\rho \in [0,1]$: noise density controlling the fraction of safe cross-edges added.
\end{itemize}
\noindent The total number of nodes and edges are given by:
$$N = (d + 1) + n_c \cdot (d + c_e - 1)$$ 
$$E_{\text{structural}} = d + n_c \cdot (d + c_e) $$
$$E_{\text{estimated}} = E_{\text{structural}} + \rho \cdot |C|$$
where $|C|$ is the number of candidate noise edges.

\begin{promptbox}
\textbf{Algorithm: Graph Shortest Path Puzzle Generation}

\medskip
\textbf{Input:} Parameters $(d, n_c, c_e, \rho)$

\medskip
\textbf{Step 1: Backbone Construction.}
\begin{enumerate}[noitemsep,topsep=2pt]
    \item Compute total nodes $N = (d+1) + n_c \cdot (d + c_e - 1)$.
    \item Generate $N$ unique alphanumeric node IDs (e.g.\ \texttt{A1234}).
    \item Assign the first $d+1$ nodes as the \emph{backbone}: $b_0 \to b_1 \to \cdots \to b_d$.
    \item Set $\textit{source} = b_0$, $\textit{target} = b_d$.
    \item Add edges $(b_i, b_{i+1})$ for $i = 0, \ldots, d-1$. This creates the unique shortest path of length $d$.
\end{enumerate}

\medskip
\textbf{Step 2: Parallel Bypass Chains.}
\begin{enumerate}[noitemsep,topsep=2pt]
    \item For each chain $j = 1, \ldots, n_c$:
    \begin{itemize}[noitemsep,topsep=2pt]
        \item Allocate $d + c_e - 1$ fresh nodes: $c^j_1, c^j_2, \ldots, c^j_{d+c_e-1}$.
        \item Add edges: $b_0 \to c^j_1 \to c^j_2 \to \cdots \to c^j_{d+c_e-1} \to b_d$.
    \end{itemize}
    \item Each bypass chain has total length $d + c_e > d$, so they cannot create a shorter path than the backbone.
\end{enumerate}

\medskip
\textbf{Step 3: Cross-Wiring Noise Edges.}
\begin{enumerate}[noitemsep,topsep=2pt]
    \item \textbf{Enumerate candidate edges.} For each chain node $u$, consider two types of partners:
    \begin{itemize}[noitemsep,topsep=2pt]
        \item \textbf{Chain--chain pairs:} For each other chain node $v$:
        \begin{itemize}[noitemsep,topsep=2pt]
            \item Same chain: allow if depth gap $> 1$ and $\leq c_e$.
            \item Different chains: allow if $\textit{depth}(v) - \textit{depth}(u) \leq \textit{excess}(v)$ and $\textit{depth}(u) - \textit{depth}(v) \leq \textit{excess}(u)$.
        \end{itemize}
        \item \textbf{Chain--backbone pairs:} For each backbone node $b_i$: allow only if $b_i$ is at the same depth or behind $u$ (i.e.\ $i \leq \textit{depth}(u)$ and $\textit{depth}(u) - i \leq \textit{excess}(u)$). This prevents shortcut edges.
    \end{itemize}
    No backbone--backbone pairs are ever considered.
    \item \textbf{Add edges with distance safety check.} Shuffle candidates, then for each $(u, v)$ sampled with probability $\rho$:
    \begin{itemize}[noitemsep,topsep=2pt]
        \item Compute $\textit{dist}_s(u) + 1 + \textit{dist}_t(v)$ and $\textit{dist}_s(v) + 1 + \textit{dist}_t(u)$.
        \item \textbf{Reject} if either value $\leq d$ (would create a path $\leq d$ from source to target).
        \item Otherwise, add edge $(u, v)$ and propagate updated distances via incremental BFS.
    \end{itemize}
\end{enumerate}
\medskip
\textbf{Step 4: Verification.}
\begin{enumerate}[noitemsep,topsep=2pt]
    \item Run BFS from source to target. Verify shortest path length $= d$.
    \item Count shortest paths. Verify exactly 1 unique shortest path exists.
\end{enumerate}
\end{promptbox}
When scaling complexity levels we keep the parameters $d, n_c, c_e$ constant, and vary the noise density $\rho$ to meet the desired number of edges. Once the maximum number of edges are achieved with the set parameters, we increment $d$ by 1 and continue from appropriate $p$ based on required total edges.

\subsection{Translation and Annotation procedure}\label{sec:annot_pipe}
For translating and auditing our puzzles, we employed native speakers through a third-party aggregator to verify and check the templates. Each translation was validated and audited by two separate annotators. No issues were found for $70\%$ of the translations, while minor stylistic changes were made for $30\%$ of the prompts. The annotators were appropriately compensated. Below we present the guidelines provided to the annotators.
\begin{promptbox}
\textbf{Task}: This is a Translation Verification task. 

\medskip
\textbf{Metrics}: Check for semantic equivalence with English ground truth and naturalness of the words used in translated versions.

\medskip
\textbf{Rater Process}:
\begin{enumerate}[noitemsep,topsep=2pt]
    \item For each locale, we provide a separate tab within the same google sheet labeled with the locale name.
    \item Column A: ``ENGLISH\_GROUNDTRUTH'' to check against.
    \item Column B: ``\{lang\_id\}\_TRANSLATION'' (e.g., AR\_TRANSLATION) to verify.
    \item Column C: ``Translation OK'' (Yes/No).
    \item Column D: Edited Translation to add the fixed translations if required.
    \item Column E: Comments.
\end{enumerate}

\medskip
\textbf{Instructions}:
\begin{itemize}[noitemsep,topsep=2pt]
    \item Compare the English ground truth with translation based on metrics defined above.
    \item Before you go ahead with translation please look at the rows marked as full example to see the full context of the task which will help with any ambiguities.
    \item If translation has semantic equivalence with English ground truth and naturalness of the words, Column C: ``Translation OK'' = ``Yes''.
    \item If any translation edit is required:
    \begin{itemize}[noitemsep]
        \item Column C: ``Translation OK'' = ``No''.
        \item Fill out Column D: Edited Translation.
        \item Fill out Column E: Comments with comments on what the issue was and what changes were made.
    \end{itemize}
\end{itemize}

\medskip
\textbf{Audit Process}: To be done by a separate set of annotators.
\begin{itemize}[noitemsep,topsep=2pt]
    \item \textbf{Column F}: ``Audit Done?'' [Data Validation: Yes, No].
    \item \textbf{Column G}: ``Issues Found?'' [Data Validation: True, False].
    \item \textbf{Column H}: ``Comments'' [If issue found? = True, provide a justification].
\end{itemize}
\end{promptbox}

\subsection{Prompts \label{subsecprompts}}
We now provide example prompts from all five puzzle types in our benchmark, shown across all seven evaluation languages. Each prompt is generated from the actual puzzle templates with small configuration parameters for readability. The underlying puzzle structure (states, edges, players, etc.) is identical across all languages---only the surface-level rendering changes.

\subsubsection{Sequential Linear Transforms (SLT)}\label{sec:slt_example}

\paragraph{English.\\}
\begin{promptbox}
A merchant departed from R4257 with an initial capital of 7. He traveled far and wide, met many people, and carefully recorded every transaction in his ledger. However, a sudden fierce wind scattered the ledger's pages, completely disrupting his itinerary.
Your task is to help this merchant piece the ledger back together. Starting from R4257, infer whom he met in order, reconstruct the timeline, and calculate his exact wealth after each encounter.

\medskip
\begin{itemize}[noitemsep,topsep=2pt]
    \item After ending business with Z1106, the merchant set off to visit Y3615. Y3615 reviewed the accounts and multiplied the merchant's balance by a factor of $-2$. As a token of goodwill, Y3615 also gifted the merchant 9.
    \item After ending business with N4611, the merchant set off to visit O5557. O5557 reviewed the accounts and multiplied the merchant's balance by a factor of $-4$. Before parting ways, O5557 demanded a fee of 2.
    \item After ending business with W9928, the merchant set off to visit N4611. N4611 reviewed the accounts and multiplied the merchant's balance by a multiple of $-1$. Before parting ways, N4611 demanded a fee of 1.
    \item After ending business with K5552, the merchant set off to visit E4527. E4527 reviewed the accounts and multiplied the merchant's balance by a factor of $-1$. Before parting ways, E4527 demanded a fee of 2.
    \item After ending business with O5557, the merchant set off to visit Z1106. Z1106 reviewed the accounts and multiplied the merchant's balance by a factor of 2. Before parting ways, Z1106 demanded a fee of 6.
    \item After ending business with W7924, the merchant set off to visit K5552. K5552 reviewed the accounts and multiplied the merchant's balance by a factor of 2. Before parting ways, K5552 demanded a fee of 7.
    \item After ending business with R4257, the merchant set off to visit W9928. W9928 reviewed the accounts and multiplied the merchant's balance by a factor of 1. Before parting ways, W9928 demanded a fee of 6.
    \item After ending business with Y3615, the merchant set off to visit W7924. W7924 reviewed the accounts and multiplied the merchant's balance by a factor of $-5$. Before parting ways, W7924 demanded a fee of 8.
\end{itemize}
\medskip
IMPORTANT: Output ONLY the answer in this exact format with no other text: \{``chain'': [v0, v1, v2, ...]\} where each value is the merchant's wealth after each transaction, in chronological order. Do not show any intermediate steps or explanations. Your response must be a single valid JSON object and nothing else.
\end{promptbox}

\paragraph{Arabic.\\}
\begin{promptbox}
{\arabicfont
انطلق تاجر من R4257 برأس مال أولي قدره 7. جاب الآفاق، والتقى بالعديد من الأشخاص، وسجل كل صفقة بعناية في الدفتر. ومع ذلك، بعثرت رياح عاتية ومفاجئة صفحات الدفتر، مما أربك مسار رحلته تماماً.
مهمتك هي مساعدة هذا التاجر في تجميع الدفتر من جديد. بدءاً من R4257، استنتج من التقى بهم بالترتيب، وأعد بناء التسلسل الزمني، واحسب ثروته الدقيقة بعد كل لقاء.

\medskip
\begin{itemize}[noitemsep,topsep=2pt]
    \item بعد إنهاء أعماله مع Z1106، انطلق التاجر لزيارة Y3615.راجع Y3615 الحسابات وضرب رصيد التاجر في $-2$.كعربون مودة، أهدى Y3615 التاجر أيضاً 9.
    \item بعد إنهاء أعماله مع N4611، انطلق التاجر لزيارة O5557.راجع O5557 الحسابات وضرب رصيد التاجر في $-4$.قبل الافتراق، طالب O5557 برسوم قدرها 2.
    \item بعد إنهاء أعماله مع W9928، انطلق التاجر لزيارة N4611.راجع N4611 الحسابات وضرب رصيد التاجر في $-1$.قبل الافتراق، طالب N4611 برسوم قدرها 1.
    \item بعد إنهاء أعماله مع K5552، انطلق التاجر لزيارة E4527.راجع E4527 الحسابات وضرب رصيد التاجر في $-1$.قبل الافتراق، طالب E4527 برسوم قدرها 2.
    \item بعد إنهاء أعماله مع O5557، انطلق التاجر لزيارة Z1106.راجع Z1106 الحسابات وضرب رصيد التاجر في 2.قبل الافتراق، طالب Z1106 برسوم قدرها 6.
    \item بعد إنهاء أعماله مع W7924، انطلق التاجر لزيارة K5552.راجع K5552 الحسابات وضرب رصيد التاجر في 2.قبل الافتراق، طالب K5552 برسوم قدرها 7.
    \item بعد إنهاء أعماله مع R4257، انطلق التاجر لزيارة W9928.راجع W9928 الحسابات وضرب رصيد التاجر في 1.قبل الافتراق، طالب W9928 برسوم قدرها 6.
    \item بعد إنهاء أعماله مع Y3615، انطلق التاجر لزيارة W7924.راجع W7924 الحسابات وضرب رصيد التاجر في $-5$.قبل الافتراق، طالب W7924 برسوم قدرها 8.
\end{itemize}
\medskip
\textnormal{IMPORTANT: Output ONLY the answer in this exact format with no other text: \{``chain'': [v0, v1, v2, ...]\} where each value is the merchant's wealth after each transaction, in chronological order. Do not show any intermediate steps or explanations. Your response must be a single valid JSON object and nothing else.}
}
\end{promptbox}

\paragraph{Hindi.\\}
\begin{promptbox}
{\hindifont
एक व्यापारी ने \textnormal{R4257} से 7 की पूंजी के साथ अपनी व्यापारिक यात्रा शुरू की। उसने दूर-दूर तक यात्रा की, कई लोगों से व्यापार किया और अपनी हर लेन-देन को सावधानीपूर्वक अपने बही-खाते में दर्ज किया। लेकिन अचानक आई एक भयानक आंधी ने उसके बही-खाते के पन्ने बिखेर दिए, जिससे उसकी यात्रा का पूरा क्रम बिगड़ गया।
अब आपको इस व्यापारी की मदद करनी है। \textnormal{R4257} से शुरू करते हुए, यह पता लगाएं कि वह किसके बाद किससे मिला, और हर मुलाकात के बाद उसकी सटीक संपत्ति की गणना करें।

\medskip
\begin{itemize}[noitemsep,topsep=2pt]
    \item \textnormal{Z1106} के साथ अपना काम खत्म करने के बाद, व्यापारी \textnormal{Y3615} से मिलने निकल पड़ा। \textnormal{Y3615} ने बही-खाते की जांच की, और व्यापारी की शेष राशि को $-2$ से गुणा किया। सद्भावना के प्रतीक के रूप में, \textnormal{Y3615} ने व्यापारी को 9 का उपहार भी दिया।
    \item \textnormal{N4611} के साथ अपना काम खत्म करने के बाद, व्यापारी \textnormal{O5557} से मिलने निकल पड़ा। \textnormal{O5557} ने बही-खाते की जांच की, और व्यापारी की शेष राशि को $-4$ से गुणा किया। वहां से विदा लेने से पहले, \textnormal{O5557} ने 2 का शुल्क मांगा।
    \item \textnormal{W9928} के साथ अपना काम खत्म करने के बाद, व्यापारी \textnormal{N4611} से मिलने निकल पड़ा। \textnormal{N4611} ने बही-खाते की जांच की, और व्यापारी की शेष राशि को $-1$ से गुणा किया। वहां से विदा लेने से पहले, \textnormal{N4611} ने 1 का शुल्क मांगा।
    \item \textnormal{K5552} के साथ अपना काम खत्म करने के बाद, व्यापारी \textnormal{E4527} से मिलने निकल पड़ा। \textnormal{E4527} ने बही-खाते की जांच की, और व्यापारी की शेष राशि को $-1$ से गुणा किया। वहां से विदा लेने से पहले, \textnormal{E4527} ने 2 का शुल्क मांगा।
    \item \textnormal{O5557} के साथ अपना काम खत्म करने के बाद, व्यापारी \textnormal{Z1106} से मिलने निकल पड़ा। \textnormal{Z1106} ने बही-खाते की जांच की, और व्यापारी की शेष राशि को 2 से गुणा किया। वहां से विदा लेने से पहले, \textnormal{Z1106} ने 6 का शुल्क मांगा।
    \item \textnormal{W7924} के साथ अपना काम खत्म करने के बाद, व्यापारी \textnormal{K5552} से मिलने निकल पड़ा। \textnormal{K5552} ने बही-खाते की जांच की, और व्यापारी की शेष राशि को 2 से गुणा किया। वहां से विदा लेने से पहले, \textnormal{K5552} ने 7 का शुल्क मांगा।
    \item \textnormal{R4257} के साथ अपना काम खत्म करने के बाद, व्यापारी \textnormal{W9928} से मिलने निकल पड़ा। \textnormal{W9928} ने बही-खाते की जांच की, और व्यापारी की शेष राशि को 1 से गुणा किया। वहां से विदा लेने से पहले, \textnormal{W9928} ने 6 का शुल्क मांगा।
    \item \textnormal{Y3615} के साथ अपना काम खत्म करने के बाद, व्यापारी \textnormal{W7924} से मिलने निकल पड़ा। \textnormal{W7924} ने बही-खाते की जांच की, और व्यापारी की शेष राशि को $-5$ से गुणा किया। वहां से विदा लेने से पहले, \textnormal{W7924} ने 8 का शुल्क मांगा।
\end{itemize}
\medskip
\textnormal{IMPORTANT: Output ONLY the answer in this exact format with no other text: \{``chain'': [v0, v1, v2, ...]\} where each value is the merchant's wealth after each transaction, in chronological order. Do not show any intermediate steps or explanations. Your response must be a single valid \textnormal{JSON} object and nothing else.}
}
\end{promptbox}

\paragraph{Japanese.\\}
\begin{promptbox}
{\japanesefont
ある商人が初期資本7を持ってR4257を出発しました。彼はあちこちを旅し、多くの人と出会い、 すべての取引を帳簿に注意深く記録しました。しかし、突然の強風で帳簿のページが散乱し、 彼の旅程は完全に分からなくなってしまいました。
あなたの任務は、この商人が帳簿を元通りに繋ぎ合わせるのを手伝うことです。R4257から出発し、 彼が順番に誰と出会ったかを推測してタイムラインを再構築し、それぞれの出会いの後の彼の正確な所持金を計算してください。

\medskip
\begin{itemize}[noitemsep,topsep=2pt]
    \item Z1106との取引を終えた後、商人はY3615を訪ねるために出発しました。Y3615は帳簿を確認し、商人の残高に$-2$を掛けました。親善の印として、Y3615は商人に9を贈りました。
    \item N4611との取引を終えた後、商人はO5557を訪ねるために出発しました。O5557は帳簿を確認し、商人の残高に$-4$を掛けました。別れる前に、O5557は2の料金を要求しました。
    \item W9928との取引を終えた後、商人はN4611を訪ねるために出発しました。N4611は帳簿を確認し、商人の残高に$-1$を掛けました。別れる前に、N4611は1の料金を要求しました。
    \item K5552との取引を終えた後、商人はE4527を訪ねるために出発しました。E4527は帳簿を確認し、商人の残高に$-1$を掛けました。別れる前に、E4527は2の料金を要求しました。
    \item O5557との取引を終えた後、商人はZ1106を訪ねるために出発しました。Z1106は帳簿を確認し、商人の残高に2を掛けました。別れる前に、Z1106は6の料金を要求しました。
    \item W7924との取引を終えた後、商人はK5552を訪ねるために出発しました。K5552は帳簿を確認し、商人の残高に2を掛けました。別れる前に、K5552は7の料金を要求しました。
    \item R4257との取引を終えた後、商人はW9928を訪ねるために出発しました。W9928は帳簿を確認し、商人の残高に1を掛けました。別れる前に、W9928は6の料金を要求しました。
    \item Y3615との取引を終えた後、商人はW7924を訪ねるために出発しました。W7924は帳簿を確認し、商人の残高に$-5$を掛けました。別れる前に、W7924は8の料金を要求しました。
\end{itemize}
\medskip
IMPORTANT: Output ONLY the answer in this exact format with no other text: \{``chain'': [v0, v1, v2, ...]\} where each value is the merchant's wealth after each transaction, in chronological order. Do not show any intermediate steps or explanations. Your response must be a single valid JSON object and nothing else.
}
\end{promptbox}

\paragraph{Tamil.\\}
\begin{promptbox}
{\tamilfont
ஒரு வியாபாரி \textnormal{R4257} -லிருந்து 7 என்ற ஆரம்ப மூலதனத்துடன் புறப்பட்டார். அவர் பல இடங்களுக்குப் பயணம் செய்து, பலரைச் சந்தித்து, ஒவ்வொரு பரிவர்த்தனையையும் தனது கணக்குப் புத்தகத்தில் கவனமாகப் பதிவு செய்தார். இருப்பினும், திடீரென வீசிய பலத்த காற்று கணக்குப் புத்தகத்தின் பக்கங்களைச் சிதறடித்து, அவரது பயணத் திட்டத்தை முற்றிலுமாக சீர்குலைத்தது.
இந்த வியாபாரிக்கு தனது கணக்குப் புத்தகத்தை மீண்டும் ஒன்றிணைக்க உதவுவதே உங்கள் பணியாகும். \textnormal{R4257} -லிருந்து தொடங்கி, அவர் யாரை, எந்த வரிசையில் சந்தித்தார் என்பதை ஊகித்து, காலவரிசையை மறுகட்டமைப்பு செய்து, ஒவ்வொரு சந்திப்பிற்கும் பிறகு அவரது சரியான செல்வத்தைக் கணக்கிடுங்கள்.

\medskip
\begin{itemize}[noitemsep,topsep=2pt]
    \item \textnormal{Z1106} உடனான வியாபாரத்தை முடித்த பிறகு, வியாபாரி \textnormal{Y3615} ஐச் சந்திக்கப் புறப்பட்டார். \textnormal{Y3615} கணக்குகளைச் சரிபார்த்து, வியாபாரியின் இருப்பை $-2$ ஆல் பெருக்கினார். நல்லெண்ணத்தின் அடையாளமாக, \textnormal{Y3615} வியாபாரிக்கு 9 -ஐப் பரிசளித்தார்.
    \item \textnormal{N4611} உடனான வியாபாரத்தை முடித்த பிறகு, வியாபாரி \textnormal{O5557} ஐச் சந்திக்கப் புறப்பட்டார். \textnormal{O5557} கணக்குகளைச் சரிபார்த்து, வியாபாரியின் இருப்பை $-4$ ஆல் பெருக்கினார். பிரியும் முன், \textnormal{O5557} 2 -ஐக் கட்டணமாகக் கோரினார்.
    \item \textnormal{W9928} உடனான வியாபாரத்தை முடித்த பிறகு, வியாபாரி \textnormal{N4611} ஐச் சந்திக்கப் புறப்பட்டார். \textnormal{N4611} கணக்குகளைச் சரிபார்த்து, வியாபாரியின் இருப்பை $-1$ ஆல் பெருக்கினார். பிரியும் முன், \textnormal{N4611} 1 -ஐக் கட்டணமாகக் கோரினார்.
    \item \textnormal{K5552} உடனான வியாபாரத்தை முடித்த பிறகு, வியாபாரி \textnormal{E4527} ஐச் சந்திக்கப் புறப்பட்டார். \textnormal{E4527} கணக்குகளைச் சரிபார்த்து, வியாபாரியின் இருப்பை $-1$ ஆல் பெருக்கினார். பிரியும் முன், \textnormal{E4527} 2 -ஐக் கட்டணமாகக் கோரினார்.
    \item \textnormal{O5557} உடனான வியாபாரத்தை முடித்த பிறகு, வியாபாரி \textnormal{Z1106} ஐச் சந்திக்கப் புறப்பட்டார். \textnormal{Z1106} கணக்குகளைச் சரிபார்த்து, வியாபாரியின் இருப்பை 2 ஆல் பெருக்கினார். பிரியும் முன், \textnormal{Z1106} 6 -ஐக் கட்டணமாகக் கோரினார்.
    \item \textnormal{W7924} உடனான வியாபாரத்தை முடித்த பிறகு, வியாபாரி \textnormal{K5552} ஐச் சந்திக்கப் புறப்பட்டார். \textnormal{K5552} கணக்குகளைச் சரிபார்த்து, வியாபாரியின் இருப்பை 2 ஆல் பெருக்கினார். பிரியும் முன், \textnormal{K5552} 7 -ஐக் கட்டணமாகக் கோரினார்.
    \item \textnormal{R4257} உடனான வியாபாரத்தை முடித்த பிறகு, வியாபாரி \textnormal{W9928} ஐச் சந்திக்கப் புறப்பட்டார். \textnormal{W9928} கணக்குகளைச் சரிபார்த்து, வியாபாரியின் இருப்பை 1 ஆல் பெருக்கினார். பிரியும் முன், \textnormal{W9928} 6 -ஐக் கட்டணமாகக் கோரினார்.
    \item \textnormal{Y3615} உடனான வியாபாரத்தை முடித்த பிறகு, வியாபாரி \textnormal{W7924} ஐச் சந்திக்கப் புறப்பட்டார். \textnormal{W7924} கணக்குகளைச் சரிபார்த்து, வியாபாரியின் இருப்பை $-5$ ஆல் பெருக்கினார். பிரியும் முன், \textnormal{W7924} 8 -ஐக் கட்டணமாகக் கோரினார்.
\end{itemize}
\medskip
\textnormal{IMPORTANT: Output ONLY the answer in this exact format with no other text: \{``chain'': [v0, v1, v2, ...]\} where each value is the merchant's wealth after each transaction, in chronological order. Do not show any intermediate steps or explanations. Your response must be a single valid \textnormal{JSON} object and nothing else.}
}
\end{promptbox}

\paragraph{Telugu.\\}
\begin{promptbox}
{\telugufont
ఒక వ్యాపారి 7 ప్రారంభ పెట్టుబడితో \textnormal{R4257} నుండి బయలుదేరాడు. అతను దేశదేశాలు తిరిగి, ఎంతో మందిని కలిసి, ప్రతి లావాదేవీని తన ఖాతా పుస్తకంలో జాగ్రత్తగా నమోదు చేసుకున్నాడు. అయితే, అకస్మాత్తుగా వీచిన బలమైన గాలి ఖాతా పుస్తకం పేజీలను చెల్లాచెదురు చేసింది, దీని వల్ల అతని ప్రయాణ వివరాలు పూర్తిగా తారుమారయ్యాయి.
ఈ వ్యాపారి తన ఖాతా పుస్తకాన్ని మళ్లీ ఒక క్రమంలో పెట్టడానికి సహాయం చేయడమే మీ పని. \textnormal{R4257} నుండి ప్రారంభించి, అతను వరుసగా ఎవరిని కలిశాడో ఊహించండి, కాలక్రమాన్ని పునర్నిర్మించండి మరియు ప్రతి ఒక్కరినీ కలిసిన తర్వాత అతని వద్ద ఉన్న ఖచ్చితమైన సంపదను లెక్కించండి.

\medskip
\begin{itemize}[noitemsep,topsep=2pt]
    \item \textnormal{Z1106} తో వ్యాపారం ముగిసిన తర్వాత, వ్యాపారి \textnormal{Y3615} ని కలవడానికి బయలుదేరాడు.\textnormal{Y3615} ఖాతాలను పరిశీలించి, వ్యాపారి బ్యాలెన్స్‌ను $-2$ తో గుణించారు. సద్భావనకు చిహ్నంగా, \textnormal{Y3615} వ్యాపారికి 9 కూడా బహుమతిగా ఇచ్చారు.
    \item \textnormal{N4611} తో వ్యాపారం ముగిసిన తర్వాత, వ్యాపారి \textnormal{O5557} ని కలవడానికి బయలుదేరాడు.\textnormal{O5557} ఖాతాలను పరిశీలించి, వ్యాపారి బ్యాలెన్స్‌ను $-4$ తో గుణించారు. విడిపోయే ముందు, \textnormal{O5557} 2 రుసుముగా అడిగారు.
    \item \textnormal{W9928} తో వ్యాపారం ముగిసిన తర్వాత, వ్యాపారి \textnormal{N4611} ని కలవడానికి బయలుదేరాడు.\textnormal{N4611} ఖాతాలను పరిశీలించి, వ్యాపారి బ్యాలెన్స్‌ను $-1$ తో గుణించారు. విడిపోయే ముందు, \textnormal{N4611} 1 రుసుముగా అడిగారు.
    \item \textnormal{K5552} తో వ్యాపారం ముగిసిన తర్వాత, వ్యాపారి \textnormal{E4527} ని కలవడానికి బయలుదేరాడు.\textnormal{E4527} ఖాతాలను పరిశీలించి, వ్యాపారి బ్యాలెన్స్‌ను $-1$ తో గుణించారు. విడిపోయే ముందు, \textnormal{E4527} 2 రుసుముగా అడిగారు.
    \item \textnormal{O5557} తో వ్యాపారం ముగిసిన తర్వాత, వ్యాపారి \textnormal{Z1106} ని కలవడానికి బయలుదేరాడు.\textnormal{Z1106} ఖాతాలను పరిశీలించి, వ్యాపారి బ్యాలెన్స్‌ను 2 తో గుణించారు. విడిపోయే ముందు, \textnormal{Z1106} 6 రుసుముగా అడిగారు.
    \item \textnormal{W7924} తో వ్యాపారం ముగిసిన తర్వాత, వ్యాపారి \textnormal{K5552} ని కలవడానికి బయలుదేరాడు.\textnormal{K5552} ఖాతాలను పరిశీలించి, వ్యాపారి బ్యాలెన్స్‌ను 2 తో గుణించారు. విడిపోయే ముందు, \textnormal{K5552} 7 రుసుముగా అడిగారు.
    \item \textnormal{R4257} తో వ్యాపారం ముగిసిన తర్వాత, వ్యాపారి \textnormal{W9928} ని కలవడానికి బయలుదేరాడు.\textnormal{W9928} ఖాతాలను పరిశీలించి, వ్యాపారి బ్యాలెన్స్‌ను 1 తో గుణించారు. విడిపోయే ముందు, \textnormal{W9928} 6 రుసుముగా అడిగారు.
    \item \textnormal{Y3615} తో వ్యాపారం ముగిసిన తర్వాత, వ్యాపారి \textnormal{W7924} ని కలవడానికి బయలుదేరాడు.\textnormal{W7924} ఖాతాలను పరిశీలించి, వ్యాపారి బ్యాలెన్స్‌ను $-5$ తో గుణించారు. విడిపోయే ముందు, \textnormal{W7924} 8 రుసుముగా అడిగారు.
\end{itemize}
\medskip
\textnormal{IMPORTANT: Output ONLY the answer in this exact format with no other text: \{``chain'': [v0, v1, v2, ...]\} where each value is the merchant's wealth after each transaction, in chronological order. Do not show any intermediate steps or explanations. Your response must be a single valid \textnormal{JSON} object and nothing else.}
}
\end{promptbox}

\paragraph{Chinese.\\}
\begin{promptbox}
一位商人带着7的初始本金从R4257启程。他游历四方，结识了许多人，并将每一笔交易都仔细记录在账本中。然而，一阵突如其来的狂风吹散了账本的书页，将他的行程彻底打乱。
你的任务是帮助这位商人重新拼凑账本。从R4257出发，推断出他先后遇见了谁，重建时间线，并计算出他每次遭遇后的确切财富。

\medskip
\begin{itemize}[noitemsep,topsep=2pt]
    \item 在与Z1106结束生意后，商人启程去拜访Y3615。Y3615审核了账目，将商人的余额乘以了$-2$。作为善意的象征，Y3615还赠予了商人9。
    \item 在与N4611结束生意后，商人启程去拜访O5557。O5557审核了账目，将商人的余额乘以了$-4$。在分道扬镳之前，O5557索要了2的费用。
    \item 在与W9928结束生意后，商人启程去拜访N4611。N4611审核了账目，将商人的余额乘以了$-1$。在分道扬镳之前，N4611索要了1的费用。
    \item 在与K5552结束生意后，商人启程去拜访E4527。E4527审核了账目，将商人的余额乘以了$-1$。在分道扬镳之前，E4527索要了2的费用。
    \item 在与O5557结束生意后，商人启程去拜访Z1106。Z1106审核了账目，将商人的余额乘以了2。在分道扬镳之前，Z1106索要了6的费用。
    \item 在与W7924结束生意后，商人启程去拜访K5552。K5552审核了账目，将商人的余额乘以了2。在分道扬镳之前，K5552索要了7的费用。
    \item 在与R4257结束生意后，商人启程去拜访W9928。W9928审核了账目，将商人的余额乘以了1。在分道扬镳之前，W9928索要了6的费用。
    \item 在与Y3615结束生意后，商人启程去拜访W7924。W7924审核了账目，将商人的余额乘以了$-5$。在分道扬镳之前，W7924索要了8的费用。
\end{itemize}
\medskip
IMPORTANT: Output ONLY the answer in this exact format with no other text: \{``chain'': [v0, v1, v2, ...]\} where each value is the merchant's wealth after each transaction, in chronological order. Do not show any intermediate steps or explanations. Your response must be a single valid JSON object and nothing else.
\end{promptbox}

\subsubsection{Discrete State Automata (DSA)}\label{sec:det_fsm_example}

\paragraph{English.\\}
\begin{promptbox}
You are navigating a building with different rooms. Each room has keys that take you to other rooms.

\medskip
You start in Room E2679.

\begin{itemize}[noitemsep,topsep=2pt]
    \item In Room V9935, if you use the Gold Key, you reach Room H4657.
    \item In Room V9935, if you use the Silver Key, you reach Room A5506.
    \item In Room U2824, if you use the Silver Key, you reach Room V9935.
    \item In Room A5506, if you use the Gold Key, you reach Room V9935.
    \item In Room A5506, if you use the Silver Key, you reach Room V9935.
    \item In Room E2679, if you use the Silver Key, you reach Room A5506.
    \item In Room H4657, if you use the Gold Key, you reach Room A5506.
    \item In Room E2679, if you use the Gold Key, you reach Room A5506.
    \item In Room U2824, if you use the Gold Key, you reach Room E2679.
    \item In Room H4657, if you use the Silver Key, you reach Room V9935.
\end{itemize}

\medskip
If you use the following sequence of keys: 
\begin{enumerate}[noitemsep,topsep=2pt]
    \item Gold Key
    \item Silver Key
    \item Gold Key
    \item Gold Key
    \item Silver Key, which room do you end up in?
\end{enumerate}
\medskip
Example response: \{``state'': ``Room D6881"\}

\medskip
IMPORTANT: The sequence has exactly 5 numbered triggers. Process them step by step: for each numbered trigger, find the matching transition rule for your current room and that trigger, then move to the next room. After applying all 5 triggers, report the final room.
\medskip
For example, if the sequence is:
\begin{enumerate}[noitemsep,topsep=2pt]
    \item Gold Key
    \item Silver Key
    \item Gold Key
\end{enumerate}
you apply trigger 1 (Gold Key), then trigger 2 (Silver Key), then trigger 3 (Gold Key), and report the room after exactly 3 steps.
\medskip
IMPORTANT: Your response must be a single valid JSON object and nothing else.
\end{promptbox}

\paragraph{Arabic.\\}
\begin{promptbox}
{\arabicfont
أنت تتنقل في مبنى به غرف مختلفة. كل غرفة بها مفاتيح تأخذك إلى غرف أخرى.

\medskip
تبدأ في غرفة E2679.

\begin{itemize}[noitemsep,topsep=2pt]
    \item في غرفة V9935، إذا استخدمت المفتاح الذهبي، فسوف ينتهي بك الأمر في غرفة H4657.
    \item في غرفة V9935، إذا استخدمت المفتاح الفضي، فسوف ينتهي بك الأمر في غرفة A5506.
    \item في غرفة U2824، إذا استخدمت المفتاح الفضي، فسوف ينتهي بك الأمر في غرفة V9935.
    \item في غرفة A5506، إذا استخدمت المفتاح الذهبي، فسوف ينتهي بك الأمر في غرفة V9935.
    \item في غرفة A5506، إذا استخدمت المفتاح الفضي، فسوف ينتهي بك الأمر في غرفة V9935.
    \item في غرفة E2679، إذا استخدمت المفتاح الفضي، فسوف ينتهي بك الأمر في غرفة A5506.
    \item في غرفة H4657، إذا استخدمت المفتاح الذهبي، فسوف ينتهي بك الأمر في غرفة A5506.
    \item في غرفة E2679، إذا استخدمت المفتاح الذهبي، فسوف ينتهي بك الأمر في غرفة A5506.
    \item في غرفة U2824، إذا استخدمت المفتاح الذهبي، فسوف ينتهي بك الأمر في غرفة E2679.
    \item في غرفة H4657، إذا استخدمت المفتاح الفضي، فسوف ينتهي بك الأمر في غرفة V9935.
\end{itemize}

\medskip
إذا استخدمت تسلسل المفاتيح التالي: 
\begin{enumerate}[noitemsep,topsep=2pt]
    \item المفتاح الذهبي
    \item المفتاح الفضي
    \item المفتاح الذهبي
    \item المفتاح الذهبي
    \item المفتاح الفضي، فما هي الغرفة التي ستنتهي بها؟
\end{enumerate}
\medskip
مثال على الاستجابة: \{``state'': ``غرفة D6881``\}

\medskip
\textnormal{IMPORTANT: The sequence has exactly 5 numbered triggers. Process them step by step: for each numbered trigger, find the matching transition rule for your current room and that trigger, then move to the next room. After applying all 5 triggers, report the final room.
\medskip
For example, if the sequence is:}
\begin{enumerate}[noitemsep,topsep=2pt]
    \item المفتاح الذهبي
    \item المفتاح الفضي
    \item المفتاح الذهبي
\end{enumerate}
\textnormal{you apply trigger 1 (}المفتاح الذهبي \textnormal{), then trigger 2 (}المفتاح الفضي \textnormal{), then trigger 3 (}المفتاح الذهبي \textnormal{) and report the room after exactly 3 steps.
\medskip
IMPORTANT: Your response must be a single valid JSON object and nothing else.}
}
\end{promptbox}

\paragraph{Hindi.\\}
\begin{promptbox}
{\hindifont
आप एक इमारत में अलग-अलग कमरों से गुज़र रहे हैं। हर कमरे में चाबियाँ हैं जो आपको दूसरे कमरों में ले जाती हैं।

\medskip
आप कमरा \textnormal{E2679} में शुरू करते हैं।

\begin{itemize}[noitemsep,topsep=2pt]
    \item कमरा \textnormal{V9935} में, यदि आप सोने की चाबी का उपयोग करते हैं, तो आप कमरा \textnormal{H4657} में पहुँचते हैं।
    \item कमरा \textnormal{V9935} में, यदि आप चांदी की चाबी का उपयोग करते हैं, तो आप कमरा \textnormal{A5506} में पहुँचते हैं।
    \item कमरा \textnormal{U2824} में, यदि आप चांदी की चाबी का उपयोग करते हैं, तो आप कमरा \textnormal{V9935} में पहुँचते हैं।
    \item कमरा \textnormal{A5506} में, यदि आप सोने की चाबी का उपयोग करते हैं, तो आप कमरा \textnormal{V9935} में पहुँचते हैं।
    \item कमरा \textnormal{A5506} में, यदि आप चांदी की चाबी का उपयोग करते हैं, तो आप कमरा \textnormal{V9935} में पहुँचते हैं।
    \item कमरा \textnormal{E2679} में, यदि आप चांदी की चाबी का उपयोग करते हैं, तो आप कमरा \textnormal{A5506} में पहुँचते हैं।
    \item कमरा \textnormal{H4657} में, यदि आप सोने की चाबी का उपयोग करते हैं, तो आप कमरा \textnormal{A5506} में पहुँचते हैं।
    \item कमरा \textnormal{E2679} में, यदि आप सोने की चाबी का उपयोग करते हैं, तो आप कमरा \textnormal{A5506} में पहुँचते हैं।
    \item कमरा \textnormal{U2824} में, यदि आप सोने की चाबी का उपयोग करते हैं, तो आप कमरा \textnormal{E2679} में पहुँचते हैं।
    \item कमरा \textnormal{H4657} में, यदि आप चांदी की चाबी का उपयोग करते हैं, तो आप कमरा \textnormal{V9935} में पहुँचते हैं।
\end{itemize}

\medskip
यदि आप चाबियों के निम्नलिखित क्रम का उपयोग करते हैं: 
\begin{enumerate}[noitemsep,topsep=2pt]
    \item सोने की चाबी
    \item चांदी की चाबी
    \item सोने की चाबी
    \item सोने की चाबी
    \item चांदी की चाबी, तो आप अंत में किस कमरे में पहुँचेंगे?
\end{enumerate}
\medskip
उदाहरण प्रतिक्रिया: \{\textnormal{``state'': ``}कमरा \textnormal{D6881"}\}

\medskip
\textnormal{IMPORTANT: The sequence has exactly 5 numbered triggers. Process them step by step: for each numbered trigger, find the matching transition rule for your current room and that trigger, then move to the next room. After applying all 5 triggers, report the final room.
\medskip
For example, if the sequence is:}
\begin{enumerate}[noitemsep,topsep=2pt]
    \item सोने की चाबी
    \item चांदी की चाबी
    \item सोने की चाबी
\end{enumerate}
\textnormal{you apply trigger 1 }(सोने की चाबी), \textnormal{then trigger 2 }(चांदी की चाबी), \textnormal{then trigger 3 }(सोने की चाबी), \textnormal{and report the room after exactly 3 steps.}
\medskip
\textnormal{IMPORTANT: Your response must be a single valid }\textnormal{JSON object and nothing else.}
}
\end{promptbox}

\paragraph{Japanese.\\}
\begin{promptbox}
{\japanesefont
\medskip
あなたは異なる部屋がある建物を移動しています。各部屋には他の部屋に移動できる鍵があります。

\medskip
あなたは 部屋 E2679 から始めます。

\begin{itemize}[noitemsep,topsep=2pt]
    \item 部屋 V9935 で 金の鍵 を使うと、部屋 H4657 にたどり着きます。
    \item 部屋 V9935 で 銀の鍵 を使うと、部屋 A5506 にたどり着きます。
    \item 部屋 U2824 で 銀の鍵 を使うと、部屋 V9935 にたどり着きます。
    \item 部屋 A5506 で 金の鍵 を使うと、部屋 V9935 にたどり着きます。
    \item 部屋 A5506 で 銀の鍵 を使うと、部屋 V9935 にたどり着きます。
    \item 部屋 E2679 で 銀の鍵 を使うと、部屋 A5506 にたどり着きます。
    \item 部屋 H4657 で 金の鍵 を使うと、部屋 A5506 にたどり着きます。
    \item 部屋 E2679 で 金の鍵 を使うと、部屋 A5506 にたどり着きます。
    \item 部屋 U2824 で 金の鍵 を使うと、部屋 E2679 にたどり着きます。
    \item 部屋 H4657 で 銀の鍵 を使うと、部屋 V9935 にたどり着きます。
\end{itemize}

\medskip
次の鍵のシーケンスを使用した場合: 
\begin{enumerate}[noitemsep,topsep=2pt]
    \item 金の鍵
    \item 銀の鍵
    \item 金の鍵
    \item 金の鍵
    \item 銀の鍵、どの部屋にたどり着きますか？
\end{enumerate}
\medskip
回答の例: \{``state'': ``部屋 D6881"\}

\medskip
IMPORTANT: The sequence has exactly 5 numbered triggers. Process them step by step: for each numbered trigger, find the matching transition rule for your current room and that trigger, then move to the next room. After applying all 5 triggers, report the final room.
\medskip
For example, if the sequence is:
\begin{enumerate}[noitemsep,topsep=2pt]
    \item 金の鍵
    \item 銀の鍵
    \item 金の鍵
\end{enumerate}
you apply trigger 1 (金の鍵), then trigger 2 (銀の鍵), then trigger 3 (金の鍵), and report the room after exactly 3 steps.
\medskip
IMPORTANT: Your response must be a single valid JSON object and nothing else.
}
\end{promptbox}

\paragraph{Tamil.\\}
\begin{promptbox}
{\tamilfont
\medskip
நீங்கள் வெவ்வேறு அறைகள் கொண்ட ஒரு கட்டிடத்திற்குள் செல்கிறீர்கள். ஒவ்வொரு அறையிலும் மற்ற அறைகளுக்கு அழைத்துச் செல்லும் சாவிகள் உள்ளன.

\medskip
நீங்கள் அறை \textnormal{E2679} இல் தொடங்குகிறீர்கள்.

\begin{itemize}[noitemsep,topsep=2pt]
    \item அறை \textnormal{V9935} இல், நீங்கள் தங்கச் சாவி ஐப் பயன்படுத்தினால், நீங்கள் அறை \textnormal{H4657}-ஐ அடைவீர்கள்.
    \item அறை \textnormal{V9935} இல், நீங்கள் வெள்ளிச் சாவி ஐப் பயன்படுத்தினால், நீங்கள் அறை \textnormal{A5506}-ஐ அடைவீர்கள்.
    \item அறை \textnormal{U2824} இல், நீங்கள் வெள்ளிச் சாவி ஐப் பயன்படுத்தினால், நீங்கள் அறை \textnormal{V9935}-ஐ அடைவீர்கள்.
    \item அறை \textnormal{A5506} இல், நீங்கள் தங்கச் சாவி ஐப் பயன்படுத்தினால், நீங்கள் அறை \textnormal{V9935}-ஐ அடைவீர்கள்.
    \item அறை \textnormal{A5506} இல், நீங்கள் வெள்ளிச் சாவி ஐப் பயன்படுத்தினால், நீங்கள் அறை \textnormal{V9935}-ஐ அடைவீர்கள்.
    \item அறை \textnormal{E2679} இல், நீங்கள் வெள்ளிச் சாவி ஐப் பயன்படுத்தினால், நீங்கள் அறை \textnormal{A5506}-ஐ அடைவீர்கள்.
    \item அறை \textnormal{H4657} இல், நீங்கள் தங்கச் சாவி ஐப் பயன்படுத்தினால், நீங்கள் அறை \textnormal{A5506}-ஐ அடைவீர்கள்.
    \item அறை \textnormal{E2679} இல், நீங்கள் தங்கச் சாவி ஐப் பயன்படுத்தினால், நீங்கள் அறை \textnormal{A5506}-ஐ அடைவீர்கள்.
    \item அறை \textnormal{U2824} இல், நீங்கள் தங்கச் சாவி ஐப் பயன்படுத்தினால், நீங்கள் அறை \textnormal{E2679}-ஐ அடைவீர்கள்.
    \item அறை \textnormal{H4657} இல், நீங்கள் வெள்ளிச் சாவி ஐப் பயன்படுத்தினால், நீங்கள் அறை \textnormal{V9935}-ஐ அடைவீர்கள்.
\end{itemize}

\medskip
பின்வரும் சாவிகளின் வரிசையை நீங்கள் பயன்படுத்தினால்: 
\begin{enumerate}[noitemsep,topsep=2pt]
    \item தங்கச் சாவி
    \item வெள்ளிச் சாவி
    \item தங்கச் சாவி
    \item தங்கச் சாவி
    \item வெள்ளிச் சாவி, இறுதியில் நீங்கள் எந்த அறையை அடைவீர்கள்?
\end{enumerate}
\medskip
எடுத்துக்காட்டு பதில்: \{\textnormal{``state'': ``}அறை \textnormal{D6881"}\}

\medskip
\textnormal{IMPORTANT: The sequence has exactly 5 numbered triggers. Process them step by step: for each numbered trigger, find the matching transition rule for your current room and that trigger, then move to the next room. After applying all 5 triggers, report the final room.
\medskip
For example, if the sequence is:}
\begin{enumerate}[noitemsep,topsep=2pt]
    \item தங்கச் சாவி
    \item வெள்ளிச் சாவி
    \item தங்கச் சாவி
\end{enumerate}
\textnormal{you apply trigger 1 }(தங்கச் சாவி), \textnormal{then trigger 2 }(வெள்ளிச் சாவி), \textnormal{then trigger 3 }(தங்கச் சாவி), \textnormal{and report the room after exactly 3 steps.}
\medskip
\textnormal{IMPORTANT: Your response must be a single valid} \textnormal{JSON object and nothing else.}
}
\end{promptbox}

\paragraph{Telugu.\\}
\begin{promptbox}
{\telugufont
మీరు వివిధ గదులు ఉన్న భవనంలో నడుస్తున్నారు. ప్రతి గదిలో మిమ్మల్ని ఇతర గదులకు తీసుకెళ్ళే తాళంచెవులు ఉన్నాయి.

\medskip
మీరు గది \textnormal{E2679} లో ప్రారంభించారు

\begin{itemize}[noitemsep,topsep=2pt]
    \item గది \textnormal{V9935}లో, మీరు బంగారు తాళంచెవి ని ఉపయోగిస్తే, మీరు గది \textnormal{H4657} కి చేరుకుంటారు.
    \item గది \textnormal{V9935}లో, మీరు వెండి తాళంచెవి ని ఉపయోగిస్తే, మీరు గది \textnormal{A5506} కి చేరుకుంటారు.
    \item గది \textnormal{U2824}లో, మీరు వెండి తాళంచెవి ని ఉపయోగిస్తే, మీరు గది \textnormal{V9935} కి చేరుకుంటారు.
    \item గది \textnormal{A5506}లో, మీరు బంగారు తాళంచెవి ని ఉపయోగిస్తే, మీరు గది \textnormal{V9935} కి చేరుకుంటారు.
    \item గది \textnormal{A5506}లో, మీరు వెండి తాళంచెవి ని ఉపయోగిస్తే, మీరు గది \textnormal{V9935} కి చేరుకుంటారు.
    \item గది \textnormal{E2679}లో, మీరు వెండి తాళంచెవి ని ఉపయోగిస్తే, మీరు గది \textnormal{A5506} కి చేరుకుంటారు.
    \item గది \textnormal{H4657}లో, మీరు బంగారు తాళంచెవి ని ఉపయోగిస్తే, మీరు గది \textnormal{A5506} కి చేరుకుంటారు.
    \item గది \textnormal{E2679}లో, మీరు బంగారు తాళంచెవి ని ఉపయోగిస్తే, మీరు గది \textnormal{A5506} కి చేరుకుంటారు.
    \item గది \textnormal{U2824}లో, మీరు బంగారు తాళంచెవి ని ఉపయోగిస్తే, మీరు గది \textnormal{E2679} కి చేరుకుంటారు.
    \item గది \textnormal{H4657}లో, మీరు వెండి తాళంచెవి ని ఉపయోగిస్తే, మీరు గది \textnormal{V9935} కి చేరుకుంటారు.
\end{itemize}

\medskip
మీరు క్రింది తాళంచెవుల క్రమాన్ని ఉపయోగిస్తే: 
\begin{enumerate}[noitemsep,topsep=2pt]
    \item బంగారు తాళంచెవి
    \item వెండి తాళంచెవి
    \item బంగారు తాళంచెవి
    \item బంగారు తాళంచెవి
    \item వెండి తాళంచెవి, చివరకు మీరు ఏ గదికి చేరుకుంటారు?
\end{enumerate}
\medskip
ఉదాహరణ ప్రతిస్పందన: \{\textnormal{``state'': ``}గది \textnormal{D6881"}\}

\medskip
\textnormal{IMPORTANT: The sequence has exactly 5 numbered triggers. Process them step by step: for each numbered trigger, find the matching transition rule for your current room and that trigger, then move to the next room. After applying all 5 triggers, report the final room.
\medskip
For example, if the sequence is:}
1. బంగారు తాళంచెవి
2. వెండి తాళంచెవి
3. బంగారు తాళంచెవి
\textnormal{you apply trigger 1 }(బంగారు తాళంచెవి), \textnormal{then trigger 2 }(వెండి తాళంచెవి), \textnormal{then trigger 3 }(బంగారు తాళంచెవి), \textnormal{and report the room after exactly 3 steps.}
\medskip
\textnormal{IMPORTANT: Your response must be a single valid} \textnormal{JSON object and nothing else.}
}
\end{promptbox}

\paragraph{Chinese.\\}
\begin{promptbox}
你正在一栋建筑物中穿行，里面有不同的房间，每个房间里都有通往其他房间的钥匙。

\medskip
你从房间 E2679 开始。

\begin{itemize}[noitemsep,topsep=2pt]
    \item 在房间 V9935，如果使用金钥匙，你会到达房间 H4657。
    \item 在房间 V9935，如果使用银钥匙，你会到达房间 A5506。
    \item 在房间 U2824，如果使用银钥匙，你会到达房间 V9935。
    \item 在房间 A5506，如果使用金钥匙，你会到达房间 V9935。
    \item 在房间 A5506，如果使用银钥匙，你会到达房间 V9935。
    \item 在房间 E2679，如果使用银钥匙，你会到达房间 A5506。
    \item 在房间 H4657，如果使用金钥匙，你会到达房间 A5506。
    \item 在房间 E2679，如果使用金钥匙，你会到达房间 A5506。
    \item 在房间 U2824，如果使用金钥匙，你会到达房间 E2679。
    \item 在房间 H4657，如果使用银钥匙，你会到达房间 V9935。
\end{itemize}

如果你按以下顺序使用钥匙：1. 金钥匙
\begin{enumerate}[noitemsep,topsep=2pt]
    \item 银钥匙
    \item 金钥匙
    \item 金钥匙
    \item 银钥匙，你最终会在哪个房间？
\end{enumerate}
\medskip
示例回答：\{``state''：``房间 D6881"\}

\medskip
IMPORTANT: The sequence has exactly 5 numbered triggers. Process them step by step: for each numbered trigger, find the matching transition rule for your current room and that trigger, then move to the next room. After applying all 5 triggers, report the final room.
\medskip
For example, if the sequence is:
\begin{enumerate}[noitemsep,topsep=2pt]
    \item 金钥匙
    \item 银钥匙
    \item 金钥匙
\end{enumerate}
you apply trigger 1 (金钥匙), then trigger 2 (银钥匙), then trigger 3 (金钥匙), and report the room after exactly 3 steps.
\medskip
IMPORTANT: Your response must be a single valid JSON object and nothing else.
\end{promptbox}

\subsubsection{Probabilistic Discrete State Automata (Pr-DSA)}\label{sec:prob_fsm_example}

\paragraph{English.\\}
\begin{promptbox}
You navigate through a building with different rooms. Each room has keys, which can lead you into one of two possible rooms, each with a specific probability.

\medskip
You begin in Room U5506.

\begin{itemize}[noitemsep,topsep=2pt]
    \item In Room D5012, if you use the Silver Key, you reach Room D5506 with probability 0.3 or Room A5012 with probability 0.7.
    \item In Room U5506, if you use the Silver Key, you reach Room D5012 with probability 0.1 or Room U5506 with probability 0.9.
    \item In Room A5012, if you use the Gold Key, you reach Room D5012 with probability 0.2 or Room D5506 with probability 0.8.
    \item In Room D5506, if you use the Gold Key, you reach Room U5506 with probability 0.2 or Room U5012 with probability 0.8.
    \item In Room D5012, if you use the Gold Key, you reach Room A5012 with probability 0.1 or Room D5012 with probability 0.9.
    \item In Room U5012, if you use the Gold Key, you reach Room A5012 with probability 0.1 or Room U5012 with probability 0.9.
    \item In Room D5506, if you use the Silver Key, you reach Room U5506 with probability 0.2 or Room A5012 with probability 0.8.
    \item In Room U5012, if you use the Silver Key, you reach Room U5012 with probability 0.2 or Room D5506 with probability 0.8.
    \item In Room U5506, if you use the Gold Key, you reach Room A5012 with probability 0.3 or Room U5012 with probability 0.7.
    \item In Room A5012, if you use the Silver Key, you reach Room D5012 with probability 0.1 or Room U5012 with probability 0.9.
\end{itemize}

If you use the following sequence of keys: Gold Key, Gold Key, Gold Key, Gold Key, Silver Key, Gold Key, Gold Key, Gold Key, Silver Key, Silver Key, Silver Key, Silver Key, Gold Key, Silver Key, Silver Key, which are the 3 most probable rooms you could reach? List these in the order from the highest to the lowest probability.
\medskip
Example answer: \{``states'': [``Room G5374", ``Room W2169", ``Room T3803", ``Room R5010", ``Room F8573", ``Room M5422", ``Room U4598", ``Room V6313"]\}

\medskip
IMPORTANT: Solve the puzzle step by step. Apply each trigger in the sequence exactly once, in order. The sequence has exactly 15 triggers. After applying all 15 triggers, report the top 8 most probable rooms at that point.
\medskip
For example, if the sequence is ``Gold Key, Silver Key, Gold Key`` (3 triggers), you apply trigger 1 (Gold Key), then trigger 2 (Silver Key), then trigger 3 (Gold Key), and report the top 3 rooms after exactly 3 steps.
\medskip
IMPORTANT: Your response must be a single valid JSON object and nothing else.
\end{promptbox}

\paragraph{Arabic.\\}
\begin{promptbox}
{\arabicfont
أنت تتنقل في مبنى به غرف مختلفة. كل غرفة بها مفاتيح قد تأخذك إلى واحدة من غرفتين محتملتين، كل منهما باحتمالية محددة.

\medskip
تبدأ في غرفة U5506.

\begin{itemize}[noitemsep,topsep=2pt]
    \item في غرفة D5012، إذا استخدمت المفتاح الفضي، فستنتقل إلى غرفة D5506 باحتمالية 0.3 أو إلى غرفة A5012 باحتمالية 0.7.
    \item في غرفة U5506، إذا استخدمت المفتاح الفضي، فستنتقل إلى غرفة D5012 باحتمالية 0.1 أو إلى غرفة U5506 باحتمالية 0.9.
    \item في غرفة A5012، إذا استخدمت المفتاح الذهبي، فستنتقل إلى غرفة D5012 باحتمالية 0.2 أو إلى غرفة D5506 باحتمالية 0.8.
    \item في غرفة D5506، إذا استخدمت المفتاح الذهبي، فستنتقل إلى غرفة U5506 باحتمالية 0.2 أو إلى غرفة U5012 باحتمالية 0.8.
    \item في غرفة D5012، إذا استخدمت المفتاح الذهبي، فستنتقل إلى غرفة A5012 باحتمالية 0.1 أو إلى غرفة D5012 باحتمالية 0.9.
    \item في غرفة U5012، إذا استخدمت المفتاح الذهبي، فستنتقل إلى غرفة A5012 باحتمالية 0.1 أو إلى غرفة U5012 باحتمالية 0.9.
    \item في غرفة D5506، إذا استخدمت المفتاح الفضي، فستنتقل إلى غرفة U5506 باحتمالية 0.2 أو إلى غرفة A5012 باحتمالية 0.8.
    \item في غرفة U5012، إذا استخدمت المفتاح الفضي، فستنتقل إلى غرفة U5012 باحتمالية 0.2 أو إلى غرفة D5506 باحتمالية 0.8.
    \item في غرفة U5506، إذا استخدمت المفتاح الذهبي، فستنتقل إلى غرفة A5012 باحتمالية 0.3 أو إلى غرفة U5012 باحتمالية 0.7.
    \item في غرفة A5012، إذا استخدمت المفتاح الفضي، فستنتقل إلى غرفة D5012 باحتمالية 0.1 أو إلى غرفة U5012 باحتمالية 0.9.
\end{itemize}

إذا استخدمت تسلسل المفاتيح التالي: المفتاح الذهبي, المفتاح الذهبي, المفتاح الذهبي, المفتاح الذهبي, المفتاح الفضي, المفتاح الذهبي, المفتاح الذهبي, المفتاح الذهبي, المفتاح الفضي, المفتاح الفضي, المفتاح الفضي, المفتاح الفضي, المفتاح الذهبي, المفتاح الفضي, المفتاح الفضي، فما هي أفضل 8 غرف محتملة يمكن أن ينتهي بك الأمر فيها؟ اذكرها بترتيب من الأكثر إلى الأقل احتمالية.

\medskip
مثال على الاستجابة: \{``states'': [``غرفة G5374``, ``غرفة W2169``, ``غرفة T3803``, ``غرفة R5010``, ``غرفة F8573``, ``غرفة M5422``, ``غرفة U4598``, ``غرفة V6313``]\}

\medskip
\textnormal{IMPORTANT: Solve the puzzle step by step. Apply each trigger in the sequence exactly once, in order. The sequence has exactly 15 triggers. After applying all 15 triggers, report the top 3 most probable rooms at that point.
\medskip
For example, if the sequence is ``Gold Key, Silver Key, Gold Key`` (3 triggers), you apply trigger 1 (Gold Key), then trigger 2 (Silver Key), then trigger 3 (Gold Key), and report the top 3 rooms after exactly 3 steps.
\medskip
IMPORTANT: Your response must be a single valid JSON object and nothing else.}
}
\end{promptbox}

\paragraph{Hindi.\\}
\begin{promptbox}
{\hindifont
आप अलग-अलग कमरों वाली एक इमारत में घूम रहे हैं। प्रत्येक कमरे में ऐसी चाबियां हैं जो आपको दो संभावित कमरों में से किसी एक में ले जा सकती हैं, और हर एक की अपनी एक निश्चित प्रायिकता है।

\medskip
आप कमरा \textnormal{U5506} में शुरू करते हैं।

\begin{itemize}[noitemsep,topsep=2pt]
    \item कमरा \textnormal{D5012} में, यदि आप चाँदी की चाबी का उपयोग करते हैं, तो आप 0.3 प्रायिकता से कमरा \textnormal{D5506} में या 0.7 प्रायिकता से कमरा \textnormal{A5012} में पहुँचते हैं।
    \item कमरा \textnormal{U5506} में, यदि आप चाँदी की चाबी का उपयोग करते हैं, तो आप 0.1 प्रायिकता से कमरा \textnormal{D5012} में या 0.9 प्रायिकता से कमरा \textnormal{U5506} में पहुँचते हैं।
    \item कमरा \textnormal{A5012} में, यदि आप सोने की चाबी का उपयोग करते हैं, तो आप 0.2 प्रायिकता से कमरा \textnormal{D5012} में या 0.8 प्रायिकता से कमरा \textnormal{D5506} में पहुँचते हैं।
    \item कमरा \textnormal{D5506} में, यदि आप सोने की चाबी का उपयोग करते हैं, तो आप 0.2 प्रायिकता से कमरा \textnormal{U5506} में या 0.8 प्रायिकता से कमरा \textnormal{U5012} में पहुँचते हैं।
    \item कमरा \textnormal{D5012} में, यदि आप सोने की चाबी का उपयोग करते हैं, तो आप 0.1 प्रायिकता से कमरा \textnormal{A5012} में या 0.9 प्रायिकता से कमरा \textnormal{D5012} में पहुँचते हैं।
    \item कमरा \textnormal{U5012} में, यदि आप सोने की चाबी का उपयोग करते हैं, तो आप 0.1 प्रायिकता से कमरा \textnormal{A5012} में या 0.9 प्रायिकता से कमरा \textnormal{U5012} में पहुँचते हैं।
    \item कमरा \textnormal{D5506} में, यदि आप चाँदी की चाबी का उपयोग करते हैं, तो आप 0.2 प्रायिकता से कमरा \textnormal{U5506} में या 0.8 प्रायिकता से कमरा \textnormal{A5012} में पहुँचते हैं।
    \item कमरा \textnormal{U5012} में, यदि आप चाँदी की चाबी का उपयोग करते हैं, तो आप 0.2 प्रायिकता से कमरा \textnormal{U5012} में या 0.8 प्रायिकता से कमरा \textnormal{D5506} में पहुँचते हैं।
    \item कमरा \textnormal{U5506} में, यदि आप सोने की चाबी का उपयोग करते हैं, तो आप 0.3 प्रायिकता से कमरा \textnormal{A5012} में या 0.7 प्रायिकता से कमरा \textnormal{U5012} में पहुँचते हैं।
    \item कमरा \textnormal{A5012} में, यदि आप चाँदी की चाबी का उपयोग करते हैं, तो आप 0.1 प्रायिकता से कमरा \textnormal{D5012} में या 0.9 प्रायिकता से कमरा \textnormal{U5012} में पहुँचते हैं।
\end{itemize}

यदि आप चाबियों के निम्नलिखित क्रम का उपयोग करते हैं: सोने की चाबी, सोने की चाबी, सोने की चाबी, सोने की चाबी, चाँदी की चाबी, सोने की चाबी, सोने की चाबी, सोने की चाबी, चाँदी की चाबी, चाँदी की चाबी, चाँदी की चाबी, चाँदी की चाबी, सोने की चाबी, चाँदी की चाबी, चाँदी की चाबी, तो वे टॉप 8 कमरे कौनसे हैं जिनमें आपके पहुँचने की प्रायिकता सबसे ज़्यादा है? प्रायिकता के घटते क्रम में सूचीबद्ध करें।

\medskip
उदाहरण प्रतिक्रिया: \{\textnormal{``states'': [``}कमरा \textnormal{G5374", ``}कमरा \textnormal{W2169", ``}कमरा \textnormal{T3803", ``}कमरा \textnormal{R5010", ``}कमरा \textnormal{F8573", ``}कमरा \textnormal{M5422", ``}कमरा \textnormal{U4598", ``}कमरा \textnormal{V6313"]}\}

\medskip
\textnormal{IMPORTANT: Solve the puzzle step by step. Apply each trigger in the sequence exactly once, in order. The sequence has exactly 15 triggers. After applying all 15 triggers, report the top 3 most probable rooms at that point.
\medskip
For example, if the sequence is ``\textnormal{Gold Key}, \textnormal{Silver Key}, \textnormal{Gold Key}`` (3 triggers), you apply trigger 1 (\textnormal{Gold Key}), then trigger 2 (\textnormal{Silver Key}), then trigger 3 (\textnormal{Gold Key}), and report the top 3 rooms after exactly 3 steps.
\medskip
\textnormal{IMPORTANT}: Your response must be a single valid \textnormal{JSON} object and nothing else.
}}
\end{promptbox}

\paragraph{Japanese.\\}
\begin{promptbox}
{\japanesefont
\medskip
あなたは、さまざまな部屋がある建物の中を移動しています。各部屋には鍵があり、鍵を使うと、候補となる2つの部屋のうちの1つに一定の確率で入れます。

\medskip
あなたは 部屋 U5506 から始めます。

\begin{itemize}[noitemsep,topsep=2pt]
    \item 部屋 D5012 で 銀の鍵 を使うと、0.3 の確率で 部屋 D5506、または 0.7 の確率で 部屋 A5012 に入ります。
    \item 部屋 U5506 で 銀の鍵 を使うと、0.1 の確率で 部屋 D5012、または 0.9 の確率で 部屋 U5506 に入ります。
    \item 部屋 A5012 で 金の鍵 を使うと、0.2 の確率で 部屋 D5012、または 0.8 の確率で 部屋 D5506 に入ります。
    \item 部屋 D5506 で 金の鍵 を使うと、0.2 の確率で 部屋 U5506、または 0.8 の確率で 部屋 U5012 に入ります。
    \item 部屋 D5012 で 金の鍵 を使うと、0.1 の確率で 部屋 A5012、または 0.9 の確率で 部屋 D5012 に入ります。
    \item 部屋 U5012 で 金の鍵 を使うと、0.1 の確率で 部屋 A5012、または 0.9 の確率で 部屋 U5012 に入ります。
    \item 部屋 D5506 で 銀の鍵 を使うと、0.2 の確率で 部屋 U5506、または 0.8 の確率で 部屋 A5012 に入ります。
    \item 部屋 U5012 で 銀の鍵 を使うと、0.2 の確率で 部屋 U5012、または 0.8 の確率で 部屋 D5506 に入ります。
    \item 部屋 U5506 で 金の鍵 を使うと、0.3 の確率で 部屋 A5012、または 0.7 の確率で 部屋 U5012 に入ります。
    \item 部屋 A5012 で 銀の鍵 を使うと、0.1 の確率で 部屋 D5012、または 0.9 の確率で 部屋 U5012 に入ります。
\end{itemize}

次の鍵のシーケンスを使用した場合: 金の鍵, 金の鍵, 金の鍵, 金の鍵, 銀の鍵, 金の鍵, 金の鍵, 金の鍵, 銀の鍵, 銀の鍵, 銀の鍵, 銀の鍵, 金の鍵, 銀の鍵, 銀の鍵、最終的に到達する可能性が最も高い上位8つの部屋は何ですか？最も可能性の高いものから低いものの順にリストしてください。

\medskip
回答の例: \{``states'': [``部屋 G5374", ``部屋 W2169", ``部屋 T3803", ``部屋 R5010", ``部屋 F8573", ``部屋 M5422", ``部屋 U4598", ``部屋 V6313"]\}

\medskip
IMPORTANT: Solve the puzzle step by step. Apply each trigger in the sequence exactly once, in order. The sequence has exactly 15 triggers. After applying all 15 triggers, report the top 3 most probable rooms at that point.
\medskip
For example, if the sequence is ``Gold Key, Silver Key, Gold Key`` (3 triggers), you apply trigger 1 (Gold Key), then trigger 2 (Silver Key), then trigger 3 (Gold Key), and report the top 3 rooms after exactly 3 steps.
\medskip
IMPORTANT: Your response must be a single valid JSON object and nothing else.
}
\end{promptbox}

\paragraph{Tamil.\\}
\begin{promptbox}
{\tamilfont
\medskip
நீங்கள் வெவ்வேறு அறைகள் கொண்ட ஒரு கட்டிடத்திற்குள் செல்கிறீர்கள். ஒவ்வொரு அறையிலும் சாவிகள் உள்ளன. அவை உங்களை இரண்டு சாத்தியமான அறைகளில் ஒன்றுக்கு அழைத்துச் செல்லக்கூடும், ஒவ்வொன்றும் ஒரு குறிப்பிட்ட சாத்தியக்கூறைக் கொண்டுள்ளது.

\medskip
நீங்கள் அறை \textnormal{U5506} இல் தொடங்குகிறீர்கள்.

\begin{itemize}[noitemsep,topsep=2pt]
    \item அறை \textnormal{D5012} இல், நீங்கள் வெள்ளிச் சாவி ஐப் பயன்படுத்தினால், நீங்கள் 0.3 சாத்தியக்கூறுடன் அறை \textnormal{D5506} க்கு அல்லது 0.7 சாத்தியக்கூறுடன் அறை \textnormal{A5012} க்குச் செல்வீர்கள்.
    \item அறை \textnormal{U5506} இல், நீங்கள் வெள்ளிச் சாவி ஐப் பயன்படுத்தினால், நீங்கள் 0.1 சாத்தியக்கூறுடன் அறை \textnormal{D5012} க்கு அல்லது 0.9 சாத்தியக்கூறுடன் அறை \textnormal{U5506} க்குச் செல்வீர்கள்.
    \item அறை \textnormal{A5012} இல், நீங்கள் தங்க சாவி ஐப் பயன்படுத்தினால், நீங்கள் 0.2 சாத்தியக்கூறுடன் அறை \textnormal{D5012} க்கு அல்லது 0.8 சாத்தியக்கூறுடன் அறை \textnormal{D5506} க்குச் செல்வீர்கள்.
    \item அறை \textnormal{D5506} இல், நீங்கள் தங்க சாவி ஐப் பயன்படுத்தினால், நீங்கள் 0.2 சாத்தியக்கூறுடன் அறை \textnormal{U5506} க்கு அல்லது 0.8 சாத்தியக்கூறுடன் அறை \textnormal{U5012} க்குச் செல்வீர்கள்.
    \item அறை \textnormal{D5012} இல், நீங்கள் தங்க சாவி ஐப் பயன்படுத்தினால், நீங்கள் 0.1 சாத்தியக்கூறுடன் அறை \textnormal{A5012} க்கு அல்லது 0.9 சாத்தியக்கூறுடன் அறை \textnormal{D5012} க்குச் செல்வீர்கள்.
    \item அறை \textnormal{U5012} இல், நீங்கள் தங்க சாவி ஐப் பயன்படுத்தினால், நீங்கள் 0.1 சாத்தியக்கூறுடன் அறை \textnormal{A5012} க்கு அல்லது 0.9 சாத்தியக்கூறுடன் அறை \textnormal{U5012} க்குச் செல்வீர்கள்.
    \item அறை \textnormal{D5506} இல், நீங்கள் வெள்ளிச் சாவி ஐப் பயன்படுத்தினால், நீங்கள் 0.2 சாத்தியக்கூறுடன் அறை \textnormal{U5506} க்கு அல்லது 0.8 சாத்தியக்கூறுடன் அறை \textnormal{A5012} க்குச் செல்வீர்கள்.
    \item அறை \textnormal{U5012} இல், நீங்கள் வெள்ளிச் சாவி ஐப் பயன்படுத்தினால், நீங்கள் 0.2 சாத்தியக்கூறுடன் அறை \textnormal{U5012} க்கு அல்லது 0.8 சாத்தியக்கூறுடன் அறை \textnormal{D5506} க்குச் செல்வீர்கள்.
    \item அறை \textnormal{U5506} இல், நீங்கள் தங்க சாவி ஐப் பயன்படுத்தினால், நீங்கள் 0.3 சாத்தியக்கூறுடன் அறை \textnormal{A5012} க்கு அல்லது 0.7 சாத்தியக்கூறுடன் அறை \textnormal{U5012} க்குச் செல்வீர்கள்.
    \item அறை \textnormal{A5012} இல், நீங்கள் வெள்ளிச் சாவி ஐப் பயன்படுத்தினால், நீங்கள் 0.1 சாத்தியக்கூறுடன் அறை \textnormal{D5012} க்கு அல்லது 0.9 சாத்தியக்கூறுடன் அறை \textnormal{U5012} க்குச் செல்வீர்கள்.
\end{itemize}

\medskip
பின்வரும் சாவிகளின் வரிசையை நீங்கள் பயன்படுத்தினால்: தங்க சாவி, தங்க சாவி, தங்க சாவி, தங்க சாவி, வெள்ளிச் சாவி, தங்க சாவி, தங்க சாவி, தங்க சாவி, வெள்ளிச் சாவி, வெள்ளிச் சாவி, வெள்ளிச் சாவி, வெள்ளிச் சாவி, தங்க சாவி, வெள்ளிச் சாவி, வெள்ளிச் சாவி, நீங்கள் முடிவடையக்கூடிய அதிக நிகழ்தகவு 8 அறைகள் எவை? அதிக நிகழ்தகவிலிருந்து குறைவான நிகழ்தகவு வரை வரிசையாக பட்டியலிடுங்கள்.
\medskip
எடுத்துக்காட்டு பதில்: \{\textnormal{``states'': [``}அறை \textnormal{G5374", ``}அறை \textnormal{W2169", ``}அறை \textnormal{T3803", ``}அறை \textnormal{R5010", ``}அறை \textnormal{F8573", ``}அறை \textnormal{M5422", ``}அறை \textnormal{U4598", ``}அறை \textnormal{V6313"]}\}

\medskip
\textnormal{IMPORTANT: Solve the puzzle step by step. Apply each trigger in the sequence exactly once, in order. The sequence has exactly 15 triggers. After applying all 15 triggers, report the top 3 most probable rooms at that point.
\medskip
For example, if the sequence is ``Gold Key, Silver Key, Gold Key" (3 triggers), you apply trigger 1 (Gold Key), then trigger 2 (Silver Key), then trigger 3 (Gold Key), and report the top 3 rooms after exactly 3 steps.
\medskip
\textnormal{IMPORTANT}: Your response must be a single valid JSON object and nothing else.
}}
\end{promptbox}

\paragraph{Telugu.\\}
\begin{promptbox}
{\telugufont
మీరు వివిధ గదులు ఉన్న భవనంలో నడుస్తున్నారు. ప్రతి గదిలో తాళంచెవులు ఉన్నాయి, అవి మిమ్మల్ని రెండు సాధ్యమైన గదులలో ఒకదానికి తీసుకెళ్ళవచ్చు, ప్రతిదానికీ ఒక నిర్దిష్ట సంభావ్యత ఉంటుంది.

\medskip
మీరు గది \textnormal{U5506} లో ప్రారంభిస్తారు.

\begin{itemize}[noitemsep,topsep=2pt]
    \item గది \textnormal{D5012} లో, మీరు వెండి తాళంచెవి ని ఉపయోగిస్తే, మీరు 0.3 సంభావ్యతతో గది \textnormal{D5506} కి లేదా 0.7 సంభావ్యతతో గది \textnormal{A5012} కి చేరుకుంటారు.
    \item గది \textnormal{U5506} లో, మీరు వెండి తాళంచెవి ని ఉపయోగిస్తే, మీరు 0.1 సంభావ్యతతో గది \textnormal{D5012} కి లేదా 0.9 సంభావ్యతతో గది \textnormal{U5506} కి చేరుకుంటారు.
    \item గది \textnormal{A5012} లో, మీరు బంగారు తాళంచెవి ని ఉపయోగిస్తే, మీరు 0.2 సంభావ్యతతో గది \textnormal{D5012} కి లేదా 0.8 సంభావ్యతతో గది \textnormal{D5506} కి చేరుకుంటారు.
    \item గది \textnormal{D5506} లో, మీరు బంగారు తాళంచెవి ని ఉపయోగిస్తే, మీరు 0.2 సంభావ్యతతో గది \textnormal{U5506} కి లేదా 0.8 సంభావ్యతతో గది \textnormal{U5012} కి చేరుకుంటారు.
    \item గది \textnormal{D5012} లో, మీరు బంగారు తాళంచెవి ని ఉపయోగిస్తే, మీరు 0.1 సంభావ్యతతో గది \textnormal{A5012} కి లేదా 0.9 సంభావ్యతతో గది \textnormal{D5012} కి చేరుకుంటారు.
    \item గది \textnormal{U5012} లో, మీరు బంగారు తాళంచెవి ని ఉపయోగిస్తే, మీరు 0.1 సంభావ్యతతో గది \textnormal{A5012} కి లేదా 0.9 సంభావ్యతతో గది \textnormal{U5012} కి చేరుకుంటారు.
    \item గది \textnormal{D5506} లో, మీరు వెండి తాళంచెవి ని ఉపయోగిస్తే, మీరు 0.2 సంభావ్యతతో గది \textnormal{U5506} కి లేదా 0.8 సంభావ్యతతో గది \textnormal{A5012} కి చేరుకుంటారు.
    \item గది \textnormal{U5012} లో, మీరు వెండి తాళంచెవి ని ఉపయోగిస్తే, మీరు 0.2 సంభావ్యతతో గది \textnormal{U5012} కి లేదా 0.8 సంభావ్యతతో గది \textnormal{D5506} కి చేరుకుంటారు.
    \item గది \textnormal{U5506} లో, మీరు బంగారు తాళంచెవి ని ఉపయోగిస్తే, మీరు 0.3 సంభావ్యతతో గది \textnormal{A5012} కి లేదా 0.7 సంభావ్యతతో గది \textnormal{U5012} కి చేరుకుంటారు.
    \item గది \textnormal{A5012} లో, మీరు వెండి తాళంచెవి ని ఉపయోగిస్తే, మీరు 0.1 సంభావ్యతతో గది \textnormal{D5012} కి లేదా 0.9 సంభావ్యతతో గది \textnormal{U5012} కి చేరుకుంటారు.
\end{itemize}

మీరు క్రింది తాళంచెవుల క్రమాన్ని ఉపయోగిస్తే: బంగారు తాళంచెవి, బంగారు తాళంచెవి, బంగారు తాళంచెవి, బంగారు తాళంచెవి, వెండి తాళంచెవి, బంగారు తాళంచెవి, బంగారు తాళంచెవి, బంగారు తాళంచెవి, వెండి తాళంచెవి, వెండి తాళంచెవి, వెండి తాళంచెవి, వెండి తాళంచెవి, బంగారు తాళంచెవి, వెండి తాళంచెవి, వెండి తాళంచెవి, మీరు చేరుకోగల అత్యంత సంభావ్య 8 గదులు ఏవి? అత్యంత సంభావ్యం నుండి తక్కువ సంభావ్యం వరకు క్రమంలో జాబితా చేయండి.

\medskip
ఉదాహరణ ప్రతిస్పందన: \{\textnormal{``states'': [``}గది \textnormal{G5374", ``}గది \textnormal{W2169", ``}గది \textnormal{T3803", ``}గది \textnormal{R5010", ``}గది \textnormal{F8573", ``}గది \textnormal{M5422", ``}గది \textnormal{U4598", ``}గది \textnormal{V6313"]}\}

\medskip
\textnormal{IMPORTANT: Solve the puzzle step by step. Apply each trigger in the sequence exactly once, in order. The sequence has exactly 15 triggers. After applying all 15 triggers, report the top 3 most probable rooms at that point.
\medskip
For example, if the sequence is ``\textnormal{Gold Key}, \textnormal{Silver Key}, \textnormal{Gold Key}`` (3 triggers), you apply trigger 1 (\textnormal{Gold Key}), then trigger 2 (\textnormal{Silver Key}), then trigger 3 (\textnormal{Gold Key}), and report the top 3 rooms after exactly 3 steps.
\medskip
\textnormal{IMPORTANT}: Your response must be a single valid \textnormal{JSON} object and nothing else.
}}
\end{promptbox}

\paragraph{Chinese.\\}
\begin{promptbox}
你正在一栋建筑物中穿行，里面有不同的房间，每个房间里都有一些钥匙。使用钥匙后，你可以前往两个特定房间中的一个，且前往每个房间的概率是预设好的。

\medskip
你从房间 U5506 开始。

\begin{itemize}[noitemsep,topsep=2pt]
    \item 在房间 D5012，如果使用银钥匙，你有 0.3 的概率会到达 房间 D5506，有 0.7 的概率会到达 房间 A5012。
    \item 在房间 U5506，如果使用银钥匙，你有 0.1 的概率会到达 房间 D5012，有 0.9 的概率会到达 房间 U5506。
    \item 在房间 A5012，如果使用金钥匙，你有 0.2 的概率会到达 房间 D5012，有 0.8 的概率会到达 房间 D5506。
    \item 在房间 D5506，如果使用金钥匙，你有 0.2 的概率会到达 房间 U5506，有 0.8 的概率会到达 房间 U5012。
    \item 在房间 D5012，如果使用金钥匙，你有 0.1 的概率会到达 房间 A5012，有 0.9 的概率会到达 房间 D5012。
    \item 在房间 U5012，如果使用金钥匙，你有 0.1 的概率会到达 房间 A5012，有 0.9 的概率会到达 房间 U5012。
    \item 在房间 D5506，如果使用银钥匙，你有 0.2 的概率会到达 房间 U5506，有 0.8 的概率会到达 房间 A5012。
    \item 在房间 U5012，如果使用银钥匙，你有 0.2 的概率会到达 房间 U5012，有 0.8 的概率会到达 房间 D5506。
    \item 在房间 U5506，如果使用金钥匙，你有 0.3 的概率会到达 房间 A5012，有 0.7 的概率会到达 房间 U5012。
    \item 在房间 A5012，如果使用银钥匙，你有 0.1 的概率会到达 房间 D5012，有 0.9 的概率会到达 房间 U5012。
\end{itemize}

假设按以下顺序使用钥匙：金钥匙, 金钥匙, 金钥匙, 金钥匙, 银钥匙, 金钥匙, 金钥匙, 金钥匙, 银钥匙, 银钥匙, 银钥匙, 银钥匙, 金钥匙, 银钥匙, 银钥匙。请问你最终所处位置概率最高的 8 个房间分别是？请按概率从高到低依次列出。

\medskip
响应示例：\{``states'': [``房间 G5374", ``房间 W2169", ``房间 T3803", ``房间 R5010", ``房间 F8573", ``房间 M5422", ``房间 U4598", ``房间 V6313"]\}

\medskip
IMPORTANT: Solve the puzzle step by step. Apply each trigger in the sequence exactly once, in order. The sequence has exactly 15 triggers. After applying all 15 triggers, report the top 3 most probable rooms at that point.
\medskip
For example, if the sequence is ``Gold Key, Silver Key, Gold Key`` (3 triggers), you apply trigger 1 (Gold Key), then trigger 2 (Silver Key), then trigger 3 (Gold Key), and report the top 3 rooms after exactly 3 steps.
\medskip
IMPORTANT: Your response must be a single valid JSON object and nothing else.
\end{promptbox}

\subsubsection{Graph Shortest Path}\label{sec:graph_sp_example}

\paragraph{English.\\}
\begin{promptbox}
Social network connections:
\begin{enumerate}[noitemsep,topsep=2pt]
    \item W9928 and C4582 are neighbors.
    \item C7912 and R4257 are classmates.
    \item H4657 and C7912 are neighbors.
    \item H9279 and U2824 are classmates.
    \item U2824 and H4657 are coworkers.
    \item A5506 and B1488 are coworkers.
    \item W9928 and V9935 are relatives.
    \item T1434 and V9935 are relatives.
    \item B1488 and R4257 are relatives.
    \item E2679 and T1434 are relatives.
    \item E2679 and A5506 are classmates.
    \item H9279 and C4582 are friends.
\end{enumerate}
\medskip
Question: What is the shortest chain of connections from H4657 to V9935? List the people in order from H4657 to V9935, including both H4657 and V9935.

\medskip
Solve the puzzle step by step using BFS algorithm. At the very end, only output your final answer as a single \textnormal{JSON} object on its own line in this exact format:
\{``path'': [``\textnormal{A1234}'', ``\textnormal{B5678}'', ``\textnormal{C9012}'']\}
The path should list all nodes from the source to the target in order, including both the source and the target. Do not include any other text after the \textnormal{JSON}.
\end{promptbox}

\paragraph{Arabic.\\}
\begin{promptbox}
{\arabicfont
روابط الشبكة الاجتماعية:
\begin{enumerate}[noitemsep,topsep=2pt]
    \item W9928 وC4582 جيران.
    \item C7912 وR4257 زملاء دراسة.
    \item H4657 وC7912 جيران.
    \item H9279 وU2824 زملاء دراسة.
    \item U2824 وH4657 زملاء عمل.
    \item A5506 وB1488 زملاء عمل.
    \item W9928 وV9935 أقارب.
    \item T1434 وV9935 أقارب.
    \item B1488 وR4257 أقارب.
    \item E2679 وT1434 أقارب.
    \item E2679 وA5506 زملاء دراسة.
    \item H9279 وC4582 أصدقاء.
\end{enumerate}
\medskip
السؤال: ما هي أقصر سلسلة اتصالات من H4657 إلى V9935؟ اذكر الأشخاص بالترتيب من H4657 إلى V9935، بما في ذلك H4657 وV9935.
}
\end{promptbox}

\paragraph{Hindi.\\}
\begin{promptbox}
{\hindifont
सामाजिक नेटवर्क के संबंध:
\begin{enumerate}[noitemsep,topsep=2pt]
    \item \textnormal{W9928} और \textnormal{C4582} पड़ोसी हैं।
    \item \textnormal{C7912} और \textnormal{R4257} सहपाठी हैं।
    \item \textnormal{H4657} और \textnormal{C7912} पड़ोसी हैं।
    \item \textnormal{H9279} और \textnormal{U2824} सहपाठी हैं।
    \item \textnormal{U2824} और \textnormal{H4657} सहकर्मी हैं।
    \item \textnormal{A5506} और \textnormal{B1488} सहकर्मी हैं।
    \item \textnormal{W9928} और \textnormal{V9935} रिश्तेदार हैं।
    \item \textnormal{T1434} और \textnormal{V9935} रिश्तेदार हैं।
    \item \textnormal{B1488} और \textnormal{R4257} रिश्तेदार हैं।
    \item \textnormal{E2679} और \textnormal{T1434} रिश्तेदार हैं।
    \item \textnormal{E2679} और \textnormal{A5506} सहपाठी हैं।
    \item \textnormal{H9279} और \textnormal{C4582} मित्र हैं।
\end{enumerate}
\medskip
प्रश्न: \textnormal{H4657} से \textnormal{V9935} तक पहुँचने का सबसे छोटा संपर्क मार्ग क्या है? \textnormal{H4657} से \textnormal{V9935} तक के सभी लोगों को क्रम में बताइए, \textnormal{H4657} और \textnormal{V9935} दोनों को शामिल करते हुए।
}

\medskip
Solve the puzzle step by step using BFS algorithm. At the very end, only output your final answer as a single \textnormal{JSON} object on its own line in this exact format:
\{``path'': [``\textnormal{A1234}'', ``\textnormal{B5678}'', ``\textnormal{C9012}'']\}
The path should list all nodes from the source to the target in order, including both the source and the target. Do not include any other text after the \textnormal{JSON}.
\end{promptbox}

\paragraph{Japanese.\\}
\begin{promptbox}
{\japanesefont
ソーシャルネットワークのつながり：
\begin{enumerate}[noitemsep,topsep=2pt]
    \item W9928とC4582は隣人です。
    \item C7912とR4257は同級生です。
    \item H4657とC7912は隣人です。
    \item H9279とU2824は同級生です。
    \item U2824とH4657は同僚です。
    \item A5506とB1488は同僚です。
    \item W9928とV9935は親戚です。
    \item T1434とV9935は親戚です。
    \item B1488とR4257は親戚です。
    \item E2679とT1434は親戚です。
    \item E2679とA5506は同級生です。
    \item H9279とC4582は友達です。
\end{enumerate}
質問： H4657からV9935までの最も短いつながりは何ですか？H4657から V9935 までの人物を、H4657 および V9935を含めて順番に挙げてください。
}

\medskip
Solve the puzzle step by step using BFS algorithm. At the very end, only output your final answer as a single \textnormal{JSON} object on its own line in this exact format:
\{``path'': [``\textnormal{A1234}'', ``\textnormal{B5678}'', ``\textnormal{C9012}'']\}
The path should list all nodes from the source to the target in order, including both the source and the target. Do not include any other text after the \textnormal{JSON}.
\end{promptbox}

\paragraph{Tamil.\\}
\begin{promptbox}
{\tamilfont
சமூக வலைப்பின்னல் இணைப்புகள்:
\begin{enumerate}[noitemsep,topsep=2pt]
    \item \textnormal{W9928} மற்றும் \textnormal{C4582} அண்டை வீட்டார்.
    \item \textnormal{C7912} மற்றும் \textnormal{R4257} வகுப்புத் தோழர்கள்.
    \item \textnormal{H4657} மற்றும் \textnormal{C7912} அண்டை வீட்டார்.
    \item \textnormal{H9279} மற்றும் \textnormal{U2824} வகுப்புத் தோழர்கள்.
    \item \textnormal{U2824} மற்றும் \textnormal{H4657} சக ஊழியர்கள்.
    \item \textnormal{A5506} மற்றும் \textnormal{B1488} சக ஊழியர்கள்.
    \item \textnormal{W9928} மற்றும் \textnormal{V9935} உறவினர்கள்.
    \item \textnormal{T1434} மற்றும் \textnormal{V9935} உறவினர்கள்.
    \item \textnormal{B1488} மற்றும் \textnormal{R4257} உறவினர்கள்.
    \item \textnormal{E2679} மற்றும் \textnormal{T1434} உறவினர்கள்.
    \item \textnormal{E2679} மற்றும் \textnormal{A5506} வகுப்புத் தோழர்கள்.
    \item \textnormal{H9279} மற்றும் \textnormal{C4582} நண்பர்கள்.
\end{enumerate}
\medskip
கேள்வி: \textnormal{H4657} இலிருந்து \textnormal{V9935} வரையிலான மிகக் குறுகிய தொடர்பு சங்கிலி என்ன? \textnormal{H4657} இலிருந்து \textnormal{V9935} வரை வரிசையாக நபர்களைப் பட்டியலிடுங்கள், \textnormal{H4657} மற்றும் \textnormal{V9935} இருவரையும் சேர்த்து.
}

\medskip
Solve the puzzle step by step using BFS algorithm. At the very end, only output your final answer as a single \textnormal{JSON} object on its own line in this exact format:
\{``path'': [``\textnormal{A1234}'', ``\textnormal{B5678}'', ``\textnormal{C9012}'']\}
The path should list all nodes from the source to the target in order, including both the source and the target. Do not include any other text after the \textnormal{JSON}.
\end{promptbox}

\paragraph{Telugu.\\}
\begin{promptbox}
{\telugufont
సామాజిక నెట్‌వర్క్ సంబంధాలు:
\begin{enumerate}[noitemsep,topsep=2pt]
    \item \textnormal{W9928} మరియు \textnormal{C4582} ఇరుగుపొరుగువారు.
    \item \textnormal{C7912} మరియు \textnormal{R4257} సహవిద్యార్థులు.
    \item \textnormal{H4657} మరియు \textnormal{C7912} ఇరుగుపొరుగువారు.
    \item \textnormal{H9279} మరియు \textnormal{U2824} సహవిద్యార్థులు.
    \item \textnormal{U2824} మరియు \textnormal{H4657} సహోద్యోగులు.
    \item \textnormal{A5506} మరియు \textnormal{B1488} సహోద్యోగులు.
    \item \textnormal{W9928} మరియు \textnormal{V9935} బంధువులు.
    \item \textnormal{T1434} మరియు \textnormal{V9935} బంధువులు.
    \item \textnormal{B1488} మరియు \textnormal{R4257} బంధువులు.
    \item \textnormal{E2679} మరియు \textnormal{T1434} బంధువులు.
    \item \textnormal{E2679} మరియు \textnormal{A5506} సహవిద్యార్థులు.
    \item \textnormal{H9279} మరియు \textnormal{C4582} స్నేహితులు.
\end{enumerate}
\medskip
ప్రశ్న: \textnormal{H4657} మరియు \textnormal{V9935} మధ్య ఉన్న అతి చిన్న పరిచయ సంబంధం ఏమిటి?\textnormal{H4657} నుండి \textnormal{V9935} వరకు వ్యక్తులను వరుసగా తెలపండి,\textnormal{H4657} మరియు \textnormal{V9935} ఇద్దరినీ చేర్చండి.
}

\medskip
Solve the puzzle step by step using BFS algorithm. At the very end, only output your final answer as a single \textnormal{JSON} object on its own line in this exact format:
\{``path'': [``\textnormal{A1234}'', ``\textnormal{B5678}'', ``\textnormal{C9012}'']\}
The path should list all nodes from the source to the target in order, including both the source and the target. Do not include any other text after the \textnormal{JSON}.
\end{promptbox}

\paragraph{Chinese.\\}
\begin{promptbox}
社交网络关系：
\begin{enumerate}[noitemsep,topsep=2pt]
    \item W9928和C4582是邻居。
    \item C7912和R4257是同学。
    \item H4657和C7912是邻居。
    \item H9279和U2824是同学。
    \item U2824和H4657是同事。
    \item A5506和B1488是同事。
    \item W9928和V9935是亲戚。
    \item T1434和V9935是亲戚。
    \item B1488和R4257是亲戚。
    \item E2679和T1434是亲戚。
    \item E2679和A5506是同学。
    \item H9279和C4582是朋友。
\end{enumerate}
\medskip
问题： 从H4657到V9935，最短的人脉关系链是什么？请按顺序列出所有将两人联系起来的人，包括H4657和V9935本人。

\medskip
Solve the puzzle step by step using BFS algorithm. At the very end, only output your final answer as a single \textnormal{JSON} object on its own line in this exact format:
\{``path'': [``\textnormal{A1234}'', ``\textnormal{B5678}'', ``\textnormal{C9012}'']\}
The path should list all nodes from the source to the target in order, including both the source and the target. Do not include any other text after the \textnormal{JSON}.
\end{promptbox}

\subsubsection{Tournament Ranking}\label{sec:tsort_example}

\paragraph{English.\\}
\begin{promptbox}
Match results:
\begin{enumerate}[noitemsep,topsep=2pt]
    \item Player V9935 defeated Player A5506.
    \item Player U2824 defeated Player A5506.
    \item Player E2679 defeated Player H4657.
    \item Player E2679 defeated Player A5506.
    \item Player A5506 defeated Player H4657.
    \item Player U2824 defeated Player E2679.
    \item Player V9935 defeated Player E2679.
    \item Player V9935 defeated Player U2824.
    \item Player V9935 defeated Player H4657.
    \item Player U2824 defeated Player H4657.
\end{enumerate}
\medskip
Question: Based on the results given above, rank all players from strongest to weakest.

\medskip
This is a topological ranking puzzle. The match results define a directed acyclic graph where each edge represents one player defeating another. Your task is to find the unique total ordering consistent with all results.

\medskip
There are exactly 5 players in total. Your final ranking must include all of them.

\medskip
Solve the puzzle step by step.
At the very end, only output your final answer as a single JSON object on its own line. Use only the player IDs (the alphanumeric codes like ``D5506'', not the full names) in this exact format:
\{``ranking'': [``<best\_id>``, ``<2nd\_id>``, ..., ``<worst\_id>``]\}
For example if Player A1234 is best and Player B5678 is worst: \{``ranking'': [``A1234'', ``B5678'']\}
Do not include any other text after the JSON.
\end{promptbox}
\paragraph{Arabic.\\}
\begin{promptbox}
{\arabicfont
نتائج البطولة:
\begin{enumerate}[noitemsep,topsep=2pt]
    \item لاعب V9935 هزم لاعب A5506.
    \item لاعب U2824 هزم لاعب A5506.
    \item لاعب E2679 هزم لاعب H4657.
    \item لاعب E2679 هزم لاعب A5506.
    \item لاعب A5506 هزم لاعب H4657.
    \item لاعب U2824 هزم لاعب E2679.
    \item لاعب V9935 هزم لاعب E2679.
    \item لاعب V9935 هزم لاعب U2824.
    \item لاعب V9935 هزم لاعب H4657.
    \item لاعب U2824 هزم لاعب H4657.
\end{enumerate}
\medskip
السؤال: بناءً على النتائج المذكورة أعلاه، رتب جميع اللاعبين من الأقوى إلى الأضعف

\medskip
\textnormal{This is a topological ranking puzzle. The match results define a directed acyclic graph where each edge represents one player defeating another. Your task is to find the unique total ordering consistent with all results.}

\medskip
هناك بالضبط 5 لاعبين في المجموع. يجب أن يتضمن تصنيفك النهائي جميعهم.

\medskip
\textnormal{Solve the puzzle step by step.
At the very end, only output your final answer as a single JSON object on its own line. Use only the player IDs (the alphanumeric codes like ``D5506'', not the full names) in this exact format:
\{``ranking'': [``<best\_id>``, ``<2nd\_id>``, ..., ``<worst\_id>``]\}
For example if Player A1234 is best and Player B5678 is worst: \{``ranking'': [``A1234'', ``B5678'']\}
Do not include any other text after the JSON.}
}
\end{promptbox}
\paragraph{Hindi.\\}
\begin{promptbox}
{\hindifont
प्रतियोगिता के परिणाम:
\begin{enumerate}[noitemsep,topsep=2pt]
    \item खिलाड़ी \textnormal{V9935} ने खिलाड़ी \textnormal{A5506} को हराया।
    \item खिलाड़ी \textnormal{U2824} ने खिलाड़ी \textnormal{A5506} को हराया।
    \item खिलाड़ी \textnormal{E2679} ने खिलाड़ी \textnormal{H4657} को हराया।
    \item खिलाड़ी \textnormal{E2679} ने खिलाड़ी \textnormal{A5506} को हराया।
    \item खिलाड़ी \textnormal{A5506} ने खिलाड़ी \textnormal{H4657} को हराया।
    \item खिलाड़ी \textnormal{U2824} ने खिलाड़ी \textnormal{E2679} को हराया।
    \item खिलाड़ी \textnormal{V9935} ने खिलाड़ी \textnormal{E2679} को हराया।
    \item खिलाड़ी \textnormal{V9935} ने खिलाड़ी \textnormal{U2824} को हराया।
    \item खिलाड़ी \textnormal{V9935} ने खिलाड़ी \textnormal{H4657} को हराया।
    \item खिलाड़ी \textnormal{U2824} ने खिलाड़ी \textnormal{H4657} को हराया।
\end{enumerate}

\medskip
प्रश्न: ऊपर दिए गए परिणामों के आधार पर, सभी खिलाड़ियों को सर्वश्रेष्ठ से सबसे कमज़ोर तक क्रमित करें।

\medskip
\textnormal{This is a topological ranking puzzle. The match results define a directed acyclic graph where each edge represents one player defeating another. Your task is to find the unique total ordering consistent with all results.}

\medskip
इसमें कुल मिलाकर 5 खिलाड़ी हैं। आपकी रैंकिंग में सभी खिलाड़ी शामिल होने चाहिए।

\medskip
\textnormal{Solve the puzzle step by step.
At the very end, only output your final answer as a single \textnormal{JSON} object on its own line. Use only the player IDs (the alphanumeric codes like ``\textnormal{D5506}'', not the full names) in this exact format:
\{``ranking'': [``<best\_id>``, ``<2nd\_id>``, ..., ``<worst\_id>``]\}
For example if \textnormal{Player} \textnormal{A1234} is best and \textnormal{Player} \textnormal{B5678} is worst: \{``ranking'': [``\textnormal{A1234}'', ``\textnormal{B5678}'']\}
Do not include any other text after the \textnormal{JSON}.}
}
\end{promptbox}
\paragraph{Japanese.\\}
\begin{promptbox}
{\japanesefont
大会結果:
\begin{enumerate}[noitemsep,topsep=2pt]
    \item プレイヤー V9935 は プレイヤー A5506 に勝利した。
    \item プレイヤー U2824 は プレイヤー A5506 に勝利した。
    \item プレイヤー E2679 は プレイヤー H4657 に勝利した。
    \item プレイヤー E2679 は プレイヤー A5506 に勝利した。
    \item プレイヤー A5506 は プレイヤー H4657 に勝利した。
    \item プレイヤー U2824 は プレイヤー E2679 に勝利した。
    \item プレイヤー V9935 は プレイヤー E2679 に勝利した。
    \item プレイヤー V9935 は プレイヤー U2824 に勝利した。
    \item プレイヤー V9935 は プレイヤー H4657 に勝利した。
    \item プレイヤー U2824 は プレイヤー H4657 に勝利した。
\end{enumerate}

\medskip
質問: 上記の結果に基づいて、すべてのプレイヤーを強い人から弱い人へ順に並べてください。

\medskip
\textnormal{This is a topological ranking puzzle. The match results define a directed acyclic graph where each edge represents one player defeating another. Your task is to find the unique total ordering consistent with all results.}

\medskip
合計でちょうど 5 人のプレイヤーがいます。最終的なランキングには全員を含める必要があります。

\medskip
\textnormal{Solve the puzzle step by step.
At the very end, only output your final answer as a single JSON object on its own line. Use only the player IDs (the alphanumeric codes like ``D5506'', not the full names) in this exact format:
\{``ranking'': [``<best\_id>``, ``<2nd\_id>``, ..., ``<worst\_id>``]\}
For example if Player A1234 is best and Player B5678 is worst: \{``ranking'': [``A1234'', ``B5678'']\}
Do not include any other text after the JSON.}
}
\end{promptbox}
\paragraph{Tamil.\\}
\begin{promptbox}
{\tamilfont
டூர்னமென்ட் முடிவுகள்
\begin{enumerate}[noitemsep,topsep=2pt]
    \item வீரர் \textnormal{V9935} வீரர் \textnormal{A5506}ஐ தோற்கடித்தார்.
    \item வீரர் \textnormal{U2824} வீரர் \textnormal{A5506}ஐ தோற்கடித்தார்.
    \item வீரர் \textnormal{E2679} வீரர் \textnormal{H4657}ஐ தோற்கடித்தார்.
    \item வீரர் \textnormal{E2679} வீரர் \textnormal{A5506}ஐ தோற்கடித்தார்.
    \item வீரர் \textnormal{A5506} வீரர் \textnormal{H4657}ஐ தோற்கடித்தார்.
    \item வீரர் \textnormal{U2824} வீரர் \textnormal{E2679}ஐ தோற்கடித்தார்.
    \item வீரர் \textnormal{V9935} வீரர் \textnormal{E2679}ஐ தோற்கடித்தார்.
    \item வீரர் \textnormal{V9935} வீரர் \textnormal{U2824}ஐ தோற்கடித்தார்.
    \item வீரர் \textnormal{V9935} வீரர் \textnormal{H4657}ஐ தோற்கடித்தார்.
    \item வீரர் \textnormal{U2824} வீரர் \textnormal{H4657}ஐ தோற்கடித்தார்.
\end{enumerate}

\medskip
கேள்வி: மேலே உள்ள முடிவுகளின் அடிப்படையில், அனைத்து வீரர்களையும் வலிமையானவரிலிருந்து பலவீனமானவர் வரை வரிசைப்படுத்துங்கள்.

\medskip
\textnormal{This is a topological ranking puzzle. The match results define a directed acyclic graph where each edge represents one player defeating another. Your task is to find the unique total ordering consistent with all results.}

\medskip
மொத்தம் சரியாக 5 வீரர்கள் உள்ளனர். உங்கள் தரவரிசையில் அந்த அனைத்து வீரர்களும் இடம்பெற வேண்டும்.

\medskip
\textnormal{Solve the puzzle step by step.
At the very end, only output your final answer as a single \textnormal{JSON} object on its own line. Use only the player IDs (the alphanumeric codes like ``\textnormal{D5506}'', not the full names) in this exact format:
\{``ranking'': [``<best\_id>``, ``<2nd\_id>``, ..., ``<worst\_id>``]\}
For example if \textnormal{Player} \textnormal{A1234} is best and \textnormal{Player} \textnormal{B5678} is worst: \{``ranking'': [``\textnormal{A1234}'', ``\textnormal{B5678}'']\}
Do not include any other text after the \textnormal{JSON}.}
}
\end{promptbox}
\paragraph{Telugu.\\}
\begin{promptbox}
{\telugufont
టోర్నమెంట్ ఫలితాలు:
\begin{enumerate}[noitemsep,topsep=2pt]
    \item ప్లేయర్ \textnormal{V9935} ప్లేయర్ \textnormal{A5506}ను ఓడించారు.
    \item ప్లేయర్ \textnormal{U2824} ప్లేయర్ \textnormal{A5506}ను ఓడించారు.
    \item ప్లేయర్ \textnormal{E2679} ప్లేయర్ \textnormal{H4657}ను ఓడించారు.
    \item ప్లేయర్ \textnormal{E2679} ప్లేయర్ \textnormal{A5506}ను ఓడించారు.
    \item ప్లేయర్ \textnormal{A5506} ప్లేయర్ \textnormal{H4657}ను ఓడించారు.
    \item ప్లేయర్ \textnormal{U2824} ప్లేయర్ \textnormal{E2679}ను ఓడించారు.
    \item ప్లేయర్ \textnormal{V9935} ప్లేయర్ \textnormal{E2679}ను ఓడించారు.
    \item ప్లేయర్ \textnormal{V9935} ప్లేయర్ \textnormal{U2824}ను ఓడించారు.
    \item ప్లేయర్ \textnormal{V9935} ప్లేయర్ \textnormal{H4657}ను ఓడించారు.
    \item ప్లేయర్ \textnormal{U2824} ప్లేయర్ \textnormal{H4657}ను ఓడించారు.
\end{enumerate}

\medskip
ప్రశ్న: పైన ఇచ్చిన ఫలితాల ఆధారంగా, అందరు ప్లేయర్ బలమైన వారి నుండి బలహీనమైన వారి వరకు ర్యాంక్ చేయండి.

\medskip
\textnormal{This is a topological ranking puzzle. The match results define a directed acyclic graph where each edge represents one player defeating another. Your task is to find the unique total ordering consistent with all results.}

\medskip
మొత్తం 5 మంది ప్లేయర్స్ ఉన్నారు. మీ ఫైనల్ ర్యాంకింగ్‌లో అందరినీ చేర్చాలి.

\medskip
\textnormal{Solve the puzzle step by step.
At the very end, only output your final answer as a single \textnormal{JSON} object on its own line. Use only the player IDs (the alphanumeric codes like ``\textnormal{D5506}'', not the full names) in this exact format:
\{``ranking'': [``<best\_id>``, ``<2nd\_id>``, ..., ``<worst\_id>``]\}
For example if \textnormal{Player} \textnormal{A1234} is best and \textnormal{Player} \textnormal{B5678} is worst: \{``ranking'': [``\textnormal{A1234}'', ``\textnormal{B5678}'']\}
Do not include any other text after the \textnormal{JSON}.}
}
\end{promptbox}
\paragraph{Chinese.\\}
\begin{promptbox}
比赛结果：
\begin{enumerate}[noitemsep,topsep=2pt]
    \item 选手 V9935 击败了 选手 A5506。
    \item 选手 U2824 击败了 选手 A5506。
    \item 选手 E2679 击败了 选手 H4657。
    \item 选手 E2679 击败了 选手 A5506。
    \item 选手 A5506 击败了 选手 H4657。
    \item 选手 U2824 击败了 选手 E2679。
    \item 选手 V9935 击败了 选手 E2679。
    \item 选手 V9935 击败了 选手 U2824。
    \item 选手 V9935 击败了 选手 H4657。
    \item 选手 U2824 击败了 选手 H4657。
\end{enumerate}

\medskip
问题： 根据上述结果，将所有选手按实力从强到弱进行排序。

\medskip
This is a topological ranking puzzle. The match results define a directed acyclic graph where each edge represents one player defeating another. Your task is to find the unique total ordering consistent with all results.

\medskip
共有 5 名选手。您的最终排名必须包括所有选手。

\medskip
Solve the puzzle step by step.
At the very end, only output your final answer as a single JSON object on its own line. Use only the player IDs (the alphanumeric codes like ``D5506'', not the full names) in this exact format:
\{``ranking'': [``<best\_id>``, ``<2nd\_id>``, ..., ``<worst\_id>``]\}
For example if Player A1234 is best and Player B5678 is worst: \{``ranking'': [``A1234'', ``B5678'']\}
Do not include any other text after the JSON.
\end{promptbox}

\end{document}